\documentclass{article}

 \usepackage[final]{neurips_2025}

\makeatletter
\renewcommand{\@noticestring}{}%
\makeatother

\usepackage[utf8]{inputenc}
\usepackage[T1]{fontenc}
\usepackage{hyperref}
\usepackage{url}
\usepackage{booktabs}
\usepackage{amsfonts}
\usepackage{amsmath,amssymb}
\usepackage{nicefrac}
\usepackage{microtype}
\usepackage{xcolor}
\usepackage{graphicx}
\usepackage{enumitem}
\usepackage{wrapfig}

\usepackage{tabularx}

\usepackage{subcaption}
\usepackage[section]{placeins} 

\newcolumntype{Y}{>{\raggedright\arraybackslash}X} 

\usepackage{subcaption} 
\usepackage{array,longtable,booktabs,makecell}
\newcolumntype{L}[1]{>{\raggedright\arraybackslash}p{#1}}
\newcommand{\rowsep}{\rule{0pt}{1.2em}} 

\usepackage{natbib} 
\defcitealias{Zelik2016}{Z16}
\defcitealias{Umberger2003}{U03}
\defcitealias{Sarabon2021Knee}{S21}
\defcitealias{Winter2009}{W09}
\defcitealias{Farris2012}{F12}
\defcitealias{Kram1990}{K90}
\defcitealias{Gottschall2003}{G03}
\defcitealias{Halsey2012StairClimbing}{H12}
\defcitealias{Baltrusch2019}{B19}
\defcitealias{Shimoyama1990}{Sh90}
\defcitealias{Seroyer2010}{Se10}

\title{Human-Level Actuation for Humanoids}

\author{%
  M. Sunbeam \\
  Teragon Research \\
  \texttt{sunbeam@teragon.ai}
}

\begin{document}
\maketitle

\begin{abstract}
Claims that humanoid robots achieve ``human-level'' actuation are common but rarely quantified. Peak torque or speed specifications tell us little about whether a joint can deliver the right combination of torque, power, and endurance at task-relevant postures and rates. We introduce a comprehensive framework that makes ``human-level'' measurable and comparable across systems. Our approach has three components. First, a kinematic \emph{DoF atlas} standardizes joint coordinate systems and ranges of motion using ISB-based conventions, ensuring that human and robot joints are compared in the same reference frames. Second, \emph{Human-Equivalence Envelopes (HEE)} define per-joint requirements by measuring whether a robot meets human torque \emph{and} power simultaneously at the same joint angle and rate $(q,\omega)$, weighted by positive mechanical work in task-specific bands (walking, stairs, lifting, reaching, and hand actions). Third, the \emph{Human-Level Actuation Score (HLAS)} aggregates six physically grounded factors: workspace coverage (ROM and DoF), HEE coverage, torque-mode bandwidth, efficiency, and thermal sustainability. We provide detailed measurement protocols using dynamometry, electrical power monitoring, and thermal testing that yield every HLAS input from reproducible experiments. A worked example demonstrates HLAS computation for a multi-joint humanoid, showing how the score exposes actuator trade-offs (gearing ratio versus bandwidth and efficiency) that peak-torque specifications obscure. The framework serves as both a design specification for humanoid development and a benchmarking standard for comparing actuation systems, with all components grounded in published human biomechanics data.
\end{abstract}

\section{Introduction}

\subsection{The problem: ``human-level'' claims without quantitative grounding}
Humanoid robotics has reached a point where multiple platforms claim ``human-level'' or ``human-like'' actuation, yet these claims lack standardized definitions or reproducible tests. Reported specifications (peak torque at unspecified postures, no-load speeds, or burst-duration power) tell us little about whether a joint can deliver the right combination of torque, power, and endurance at task-relevant operating points. Without knowing \emph{where} in the joint workspace these numbers apply, \emph{how long} they can be sustained, and \emph{how} they are rendered under contact, systems that appear similar on paper can behave very differently in practice.

This paper addresses the gap by proposing a comprehensive framework that turns human biomechanics data into explicit, measurable requirements for humanoid actuation. Our approach defines what ``human-level'' means at each joint for specific tasks, provides measurement protocols to test those definitions, and aggregates the results into a single interpretable score with full diagnostic decomposition.

\subsection{Biomechanics as the reference: what humans actually do}
Human joint work is highly structured and task-dependent. During steady walking, positive mechanical work concentrates at the ankle near push-off. As speed increases, power redistributes across ankle, knee, and hip \citep{Farris2012,Zelik2016}. Dynamic-walking models and energetics studies explain why mid-band operation matters for locomotor economy and why well-timed ankle push-off reduces center-of-mass collision losses \citep{Kuo2001,Kuo2002}. Other activities emphasize different regimes: stair ascent and lifting require sustained low-rate torque plateaus with thermal headroom rather than brief peak bursts \citep{Kram1990,Halsey2012StairClimbing}, while manipulation tasks demand high bandwidth and low backdrive impedance for safe, responsive interaction.

Decades of biomechanics research quantify these demands. Published datasets report joint moments, powers, and ranges of motion across tasks, speeds, and populations \citep{Winter2009,Farris2012,Zelik2016}. Values are typically normalized by body mass, enabling direct scaling to robots if segment anthropometry and coordinate systems are handled consistently \citep{DeLeva1996}. Clinical and functional ROM studies bound the postures humans use in everyday tasks \citep{Morrey1981,Palmer1985,AAOSROM2020}. Together, these sources define for each joint: (i) the postures humans actually use, (ii) the torque and power they produce there, and (iii) how long they can sustain it.

Our goal is to operationalize these human envelopes as \emph{robot acceptance tests}: explicit, reproducible criteria that say when a humanoid joint is genuinely operating at a human-like level for a given task.

\subsection{Reference body and scaling conventions}
To convert normalized human data into absolute joint requirements, we adopt a 75\,kg, 1.75\,m adult male as our reference subject. This choice aligns with common humanoid target anthropometries and allows straightforward comparison across platforms. Segment masses and inertias are computed using de~Leva's adjustments to the Zatsiorsky--Seluyanov regression model \citep{DeLeva1996}, which provides validated parameters for inverse dynamics and scaling. Absolute targets follow from normalized values via
\[
T_{\mathrm{abs}} = m\,(T/m), \qquad P_{\mathrm{abs}} = m\,(P/m),
\]
where $m{=}75$\,kg, and $T/m$ and $P/m$ are the mass-normalized torques and powers from the literature.

While a single reference body simplifies exposition, the framework generalizes: Section~\ref{sec:biomech} discusses population variability, and the measurement protocols (Section~\ref{sec:hlas_protocols}) apply regardless of target mass or morphology. Users can substitute different reference subjects by re-scaling the human demand fields $T^{\mathrm{hum}}_{j,t}(q,\omega)$ and $P^{\mathrm{hum}}_{j,t}(q,\omega)$ accordingly.

\subsection{Kinematic conventions: a shared coordinate system}
Comparing human and robot joints requires a common reference frame. We adopt the International Society of Biomechanics (ISB) joint coordinate system recommendations for the lower limb, spine, and upper limb \citep{Wu2002ISB,Wu2005ISB}, along with established conventions for knee and ankle complexes \citep{Grood1983,Cappozzo1995}. This ensures that terms like ``ankle dorsiflexion,'' ``shoulder internal rotation,'' or ``wrist radial deviation'' refer to the same anatomical axis, origin, and positive direction in both human and robot.

Section~\ref{sec:kinematics} formalizes these conventions into a \emph{DoF atlas}: a visual and tabular reference specifying axis directions, signs, and functional ranges of motion (ROM) for every major joint. The atlas serves two purposes: (i) it standardizes how robot designers map their actuators to human anatomy, and (ii) it defines the workspace over which performance is evaluated, ensuring that robots are not credited for torque capability in postures humans never use.

\subsection{Task-derived operating bands}
Not all regions of a joint's workspace matter equally. We define \emph{task-derived operating bands} $\mathcal{R}_{j,t} \subseteq \{(q,\omega)\}$ that restrict evaluation to posture-rate combinations where humans actually produce positive mechanical work for task $t$ at joint $j$. These bands are extracted from canonical gait, activity, and manipulation studies that report joint angle, moment, power, and angular velocity trajectories over task phases (e.g., gait cycle, reach trajectory, stair ascent cycle).

For example, during level walking the ankle push-off band $\mathcal{R}_{\mathrm{ankle,walk}}$ covers roughly $0$--$25^{\circ}$ plantarflexion at 8--12\,rad/s, where the ankle produces the majority of its positive work. In contrast, the knee support band during stair ascent emphasizes low angular rates ($<2$\,rad/s) at mid-flexion angles. By tying requirements to these empirically grounded bands, we prevent robots from gaming scores through high performance in irrelevant operating points (e.g., claiming ``human-level ankle torque'' at 0\,rad/s and $50^{\circ}$ dorsiflexion, a posture rarely used in locomotion).

\subsection{Actuator trade-offs: why peak specs are insufficient}
Actuation design involves fundamental trade-offs among torque density, bandwidth, efficiency, transparency, and thermal capacity. Series elastic actuators (SEA) introduce compliance that improves force control and shock tolerance but reduces closed-loop bandwidth and can introduce transmission losses \citep{Pratt1995,Vanderborght2013,Paine2015}. Quasi-direct-drive and direct-drive designs minimize gear ratio to maximize backdrivability and torque-mode bandwidth, at the cost of motor size and thermal stress \citep{Hutter2016ANYmal,Grimminger,Katz2019}. Hydraulic actuation offers exceptional power density and continuous-torque capability but brings integration complexity and maintenance overhead. 

Transmission choice further shapes performance. Comparative studies show that strain-wave, cycloidal, and planetary gearboxes differ significantly in bidirectional efficiency, friction, backlash, and reflected inertia \citep{LopezGarcia2020,Oberneder2024}. Increasing gear ratio $N$ raises static torque but also increases reflected inertia (${\propto}N^2$) and friction, which degrades torque-mode bandwidth, efficiency at mid-speeds, and stable impedance range. Static torque gains can thus come at the expense of interaction quality, usable ROM, or sustainable duty cycle.

Any metric for ``human-level'' actuation must expose these coupled trade-offs. A score that rewards peak torque without considering bandwidth, efficiency, or thermal endurance invites designs optimized for specifications rather than real-world task performance.

\subsection{Interaction quality and the case for continuous-safe maps}
How torque is rendered during contact matters as much as nominal capability. Impedance control and haptics theory establish that delays, sampling rates, and transmission friction determine the range of passive mechanical behaviors a robot can stably display (the Z-width) \citep{Hogan1985,ColgateBrown1994}. High gear ratios that boost static torque can reduce torque-mode bandwidth to the point where safe, responsive interaction becomes impossible, even if the actuator can theoretically produce the required forces.

We therefore argue for \emph{continuous-safe performance maps} rather than isolated peak specifications. These maps report torque as a function of joint angle and rate $(q,\omega)$ under realistic thermal and electrical conditions: after a warm-up soak at task duty, with no current saturation, and with temperature rise constrained to safe levels ($<0.5^{\circ}$C/s). Complementary measurements include task-weighted efficiency (from DC-bus power accounting), closed-loop torque-mode bandwidth with representative reflected inertia, and thermal duty profiles showing time-to-derate. These quantities are standardizable using established dynamometry and control-system identification practices \citep{Dvir2025,Hogan1985,ColgateBrown1994}.

Together, such maps provide the inputs for a holistic actuation score that reflects what a joint can actually deliver in sustained, interactive use.

\subsection{Contributions of this work}
This paper makes five contributions toward quantifying and standardizing ``human-level'' actuation:
\begin{enumerate}[leftmargin=12pt,itemsep=2pt]
\item \textbf{Kinematic foundation and DoF atlas.} We provide a comprehensive reference (Section~\ref{sec:kinematics}) that standardizes joint coordinate systems, axis directions, signs, and functional ROM across all major human joints using ISB conventions \citep{Wu2002ISB,Wu2005ISB,Grood1983,Cappozzo1995}. This atlas ensures that human and robot joint specifications are directly comparable, eliminating ambiguity about which axis or direction a torque requirement refers to.

\item \textbf{Biomechanics synthesis and human reference targets.} We systematically review and synthesize human joint moment, power, and angular velocity data from the biomechanics literature (Section~\ref{sec:biomech}), converting normalized values (Nm/kg, W/kg) to absolute targets via de~Leva anthropometry for a 75\,kg, 1.75\,m reference body. For each task, we extract joint-level torque and power requirements, angular rate bounds, and positive-work distributions, providing a coherent, robot-readable specification of human capability.

\item \textbf{Human-Equivalence Envelopes (HEE).} We define task-specific operating bands $\mathcal{R}_{j,t}$ and per-joint torque-speed-posture requirements (Section~\ref{sec:hlas}). The HEE metric asks whether a robot can meet human torque \emph{and} power requirements \emph{simultaneously} at the same $(q,\omega)$, weighted by the positive mechanical work humans produce at each point. This simultaneity condition prevents gaming via high torque at low speed combined with high speed at low torque.

\item \textbf{Human-Level Actuation Score (HLAS).} We introduce a scalar score (Section~\ref{sec:hlas}) that aggregates six physically motivated factors: workspace coverage (ROM and DoF availability), HEE coverage, torque-mode bandwidth, task-weighted efficiency, and thermal sustainability. The score is normalized so that a human scores 1.0 on their own operating bands, and it decomposes into task- and joint-level contributions for diagnostic purposes. We show through worked examples how HLAS exposes actuator trade-offs that peak-torque specifications obscure.

\item \textbf{Reproducible measurement and benchmark protocols.} We provide detailed experimental procedures (Sections~\ref{sec:hlas_protocols} and \ref{sec:benchmarks}) for collecting every input to HLAS: continuous-safe torque-speed-posture maps via dynamometry, closed-loop bandwidth via torque-mode sine sweeps, task-weighted efficiency from DC-bus logging, thermal duty tests, and ROM/DoF verification. Joint-level benchmarks establish actuator limits and task-level trials validate that these limits translate to functional performance. These protocols resist gaming through continuous-safe measurement, task-representative loading, and pre-registered weights.
\end{enumerate}

Together, these components constitute a practical specification for making ``human-level actuation'' claims verifiable and comparable. The framework serves both as a design target (guiding actuator and transmission selection during development) and as a benchmarking standard (enabling objective comparison across humanoid platforms and actuator modules).

\subsection{Paper organization}
The remainder of the paper is organized as follows. Section~\ref{sec:kinematics} presents the DoF atlas with ISB-aligned coordinate systems and ROM norms. Section~\ref{sec:biomech} compiles human biomechanics data into absolute joint-level targets for the 75\,kg reference subject. Section~\ref{sec:hlas} defines the HLAS framework, including HEE, the six aggregated factors, and the aggregation scheme. Sections~\ref{sec:hlas_protocols} and \ref{sec:benchmarks} detail measurement protocols at the joint and task levels. Section~\ref{sec:hlas_example} walks through a complete HLAS computation for a synthetic multi-joint robot. Section~\ref{sec:related} surveys related work in biomechanics, actuation, and benchmarking. Section~\ref{sec:limitations} discusses limitations and extensions, and Section~\ref{sec:conclusion} concludes with recommendations for adoption.

\section{Kinematic Foundations: DoF Atlas and Joint Coordinate Systems}
\label{sec:kinematics}

\subsection{Purpose and scope}

This section establishes the kinematic reference frame for all human-to-robot comparisons in HLAS. We define a \emph{DoF atlas} that specifies joint coordinate systems, axis directions, signs, and functional ranges of motion (ROM) for every major joint in the human body. This atlas serves three purposes:

\begin{enumerate}[leftmargin=12pt,itemsep=2pt]
\item \textbf{Common reference frames.} Ensures that terms like ``ankle dorsiflexion'' or ``shoulder internal rotation'' refer to the same anatomical axis, origin, and positive direction in both human biomechanics data and robot specifications.
\item \textbf{Workspace definition.} Defines the joint-angle ranges $I^{\mathrm{func}}_{j,t}(a)$ used to compute ROM coverage $\rho^{\mathrm{ROM}}_{j,t}$ in HLAS (Eq.~\ref{eq:rom}), ensuring robots are evaluated over postures humans actually use.
\item \textbf{DoF accounting.} Specifies which degrees of freedom exist at each joint, enabling the DoF sufficiency check $d^{\mathrm{DoF}}_{j,t}$ (Eq.~\ref{eq:dof}) that verifies a robot implements the axes required for each task.
\end{enumerate}

All human torque and power requirements in Section~\ref{sec:biomech} and all workspace-related HLAS terms in Section~\ref{sec:hlas} are expressed relative to this atlas. Figures~\ref{fig:upper_I} through \ref{fig:digits} provide visual definitions of axes and motions while Tables~\ref{tab:dof_inventory} and \ref{tab:rom_compact} tabulate DoF counts and ROM norms.

\textbf{Note on atlas scope.} The atlas includes all major joints and digits for completeness, but the worked examples and measurement protocols in this paper focus on proximal joints critical for locomotion and manipulation (hip, knee, ankle, shoulder, elbow, wrist). The digit definitions support future extensions to dexterous tasks but are not required for basic HLAS computation.

\captionsetup[subfigure]{font=small,justification=centering}
\newcommand{\panelw}{0.43\linewidth}   
\newcommand{\panelsep}{0.04\linewidth} 
\newcommand{\panelh}{4.5cm}            

\newcommand{\atlasimg}[3]{%
  \begin{subfigure}[t]{\panelw}
    \centering
    #1%
    \caption{#2}%
    \label{#3}%
  \end{subfigure}%
}

\newcommand{\atlasimgwide}[3]{%
  \begin{subfigure}[t]{0.90\linewidth} 
    \centering
    #1%
    \caption{#2}%
    \label{#3}%
  \end{subfigure}%
}

\subsection{Coordinate system conventions}

\paragraph{Frame definition and orientation.}
We adopt segment-attached, right-handed Cartesian frames with consistent orientation across all joints:
\begin{itemize}[leftmargin=12pt,itemsep=2pt]
\item $\mathbf{x}$-axis: anterior (forward)
\item $\mathbf{y}$-axis: left (medial-lateral)
\item $\mathbf{z}$-axis: superior (up)
\end{itemize}
These axes satisfy $\mathbf{x} \times \mathbf{y} = \mathbf{z}$ and follow ISB recommendations for lower limb, spine, and upper limb joints \citep{Wu2002ISB,Wu2005ISB}, with supplementary conventions for the knee (Grood-Suntay) and ankle complex \citep{Grood1983,Cappozzo1995}.

\paragraph{Sign conventions.}
Positive rotations follow the right-hand rule about each axis. Axis directions are chosen to align with anatomical convention, so positive angles correspond to:
\begin{itemize}[leftmargin=12pt,itemsep=2pt]
\item Flexion (e.g., elbow flexion, knee flexion)
\item Abduction (moving away from body midline)
\item Internal rotation (e.g., shoulder, hip)
\item Supination (forearm rotation, palm up)
\item Dorsiflexion (ankle, toes up)
\item Radial deviation (wrist, thumb side)
\item Inversion (ankle, sole inward)
\item Left yaw and left lateral bend (spine, neck)
\end{itemize}

\paragraph{Bilateral symmetry.}
Left and right limbs mirror in geometry but share the same semantic interpretation: ``positive hip flexion'' means the same motion on both sides, even though the physical rotation direction reverses. This ensures that human data reported for one side applies symmetrically.

\paragraph{Visual grammar in figures.}
To aid interpretation, we use consistent notation across all atlas figures:
\begin{itemize}[leftmargin=12pt,itemsep=2pt]
\item \textbf{In-plane rotations}: Curved arcs with arrowheads
\item \textbf{Out-of-plane rotations}: Small circular arrows about the rotation axis, annotated with $\odot$ (out of page) or $\otimes$ (into page) where needed
\item \textbf{Translational motions}: Straight arrows (scapular elevation/depression, protraction/retraction)
\end{itemize}
This grammar allows direct mapping from visual diagrams to the symbolic DoF notation in Table~\ref{tab:dof_inventory}.

\paragraph{Oblique anatomical axes.}
Some anatomical axes are intrinsically oblique relative to the Cartesian frame (e.g., subtalar inversion/eversion, thumb carpometacarpal motion). In these cases we define the primary Cartesian frame as above but note the anatomical axis tilt. This distinction matters for actuator placement but does not affect the sign conventions or ROM intervals used in HLAS scoring.


\subsection{Upper body}
\label{subsec:upper_body}

\begin{figure}[!htbp]
  \centering
  \atlasimgwide{%
    \includegraphics[height=\panelh,keepaspectratio]{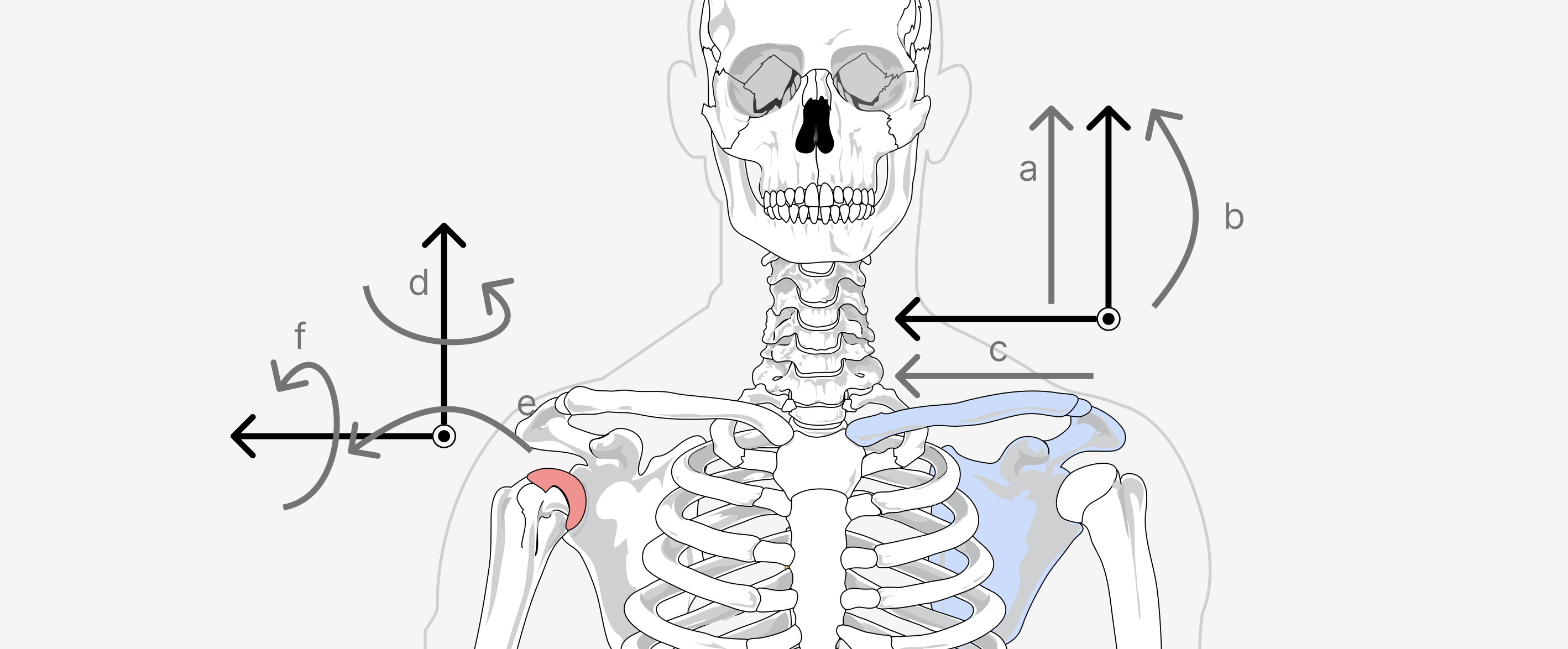}%
  }{%
    Shoulder complex.\\
    (a) elevation/depression\\
    (b) upward/downward rotation\\
    (c) retraction/protraction\\
    (d) external/internal rotation\\
    (e) abduction/adduction\\
    (f) flexion/extension.%
  }{fig:shoulder_complex}
  \caption{Upper body atlas I: Shoulder complex including scapulothoracic contributions. Origins are schematic and do not correspond to exact anatomical joint centers. Motions (a)--(c) are scapular translations/rotations, and (d)--(f) are glenohumeral rotations. This figure and all following ones of the skeleton image was adapted from \cite{pixabay-skeleton-41550}.}
  \label{fig:upper_I}
\end{figure}

\begin{figure}[!htbp]
  \centering
  \atlasimg{%
    \includegraphics[height=\panelh,keepaspectratio]{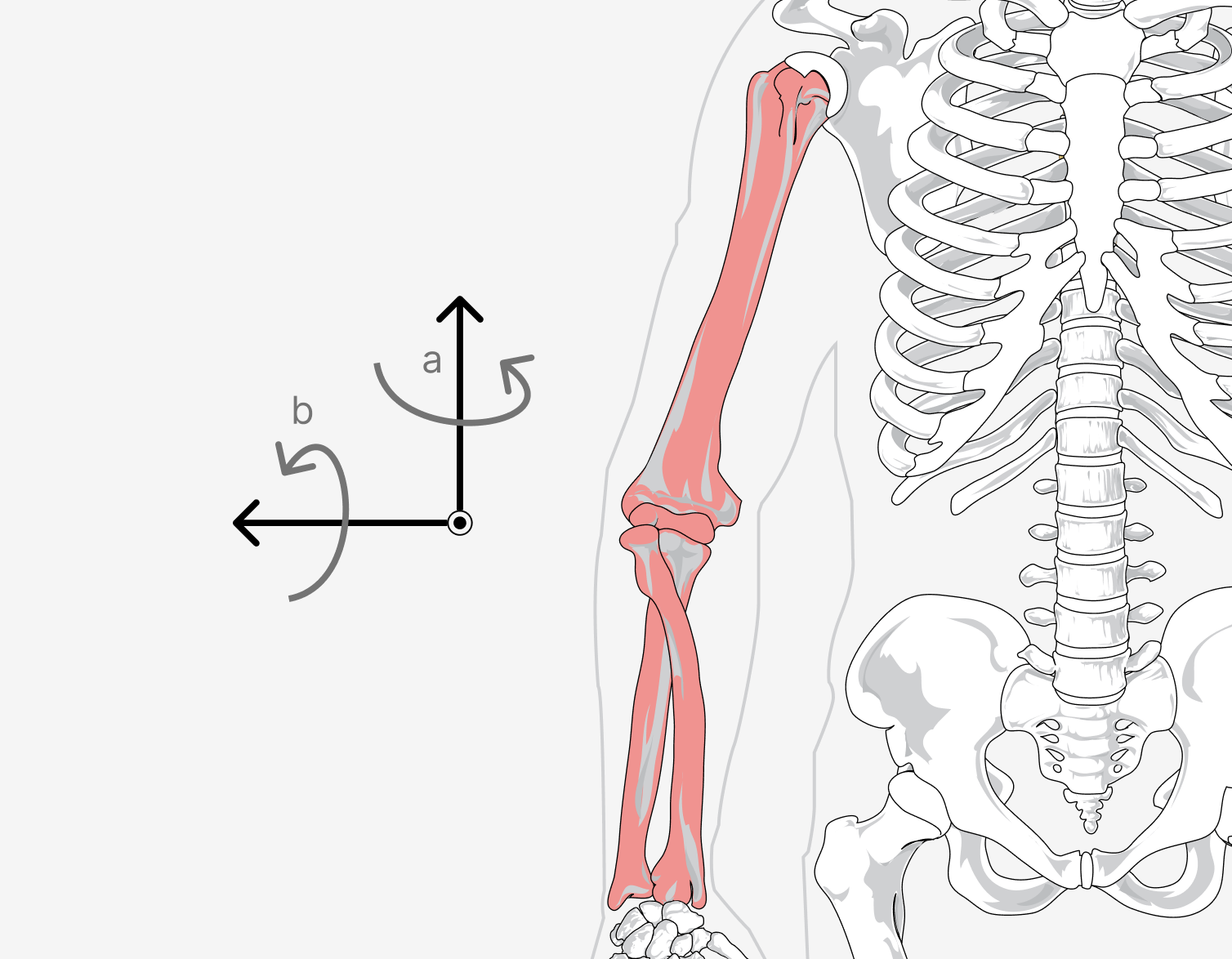}%
  }{%
    Arm.\\
    (a) pronation/supination\\
    (b) flexion/extension.%
  }{fig:arm}
  \hspace{\panelsep}
  \atlasimg{%
    \includegraphics[height=\panelh,keepaspectratio]{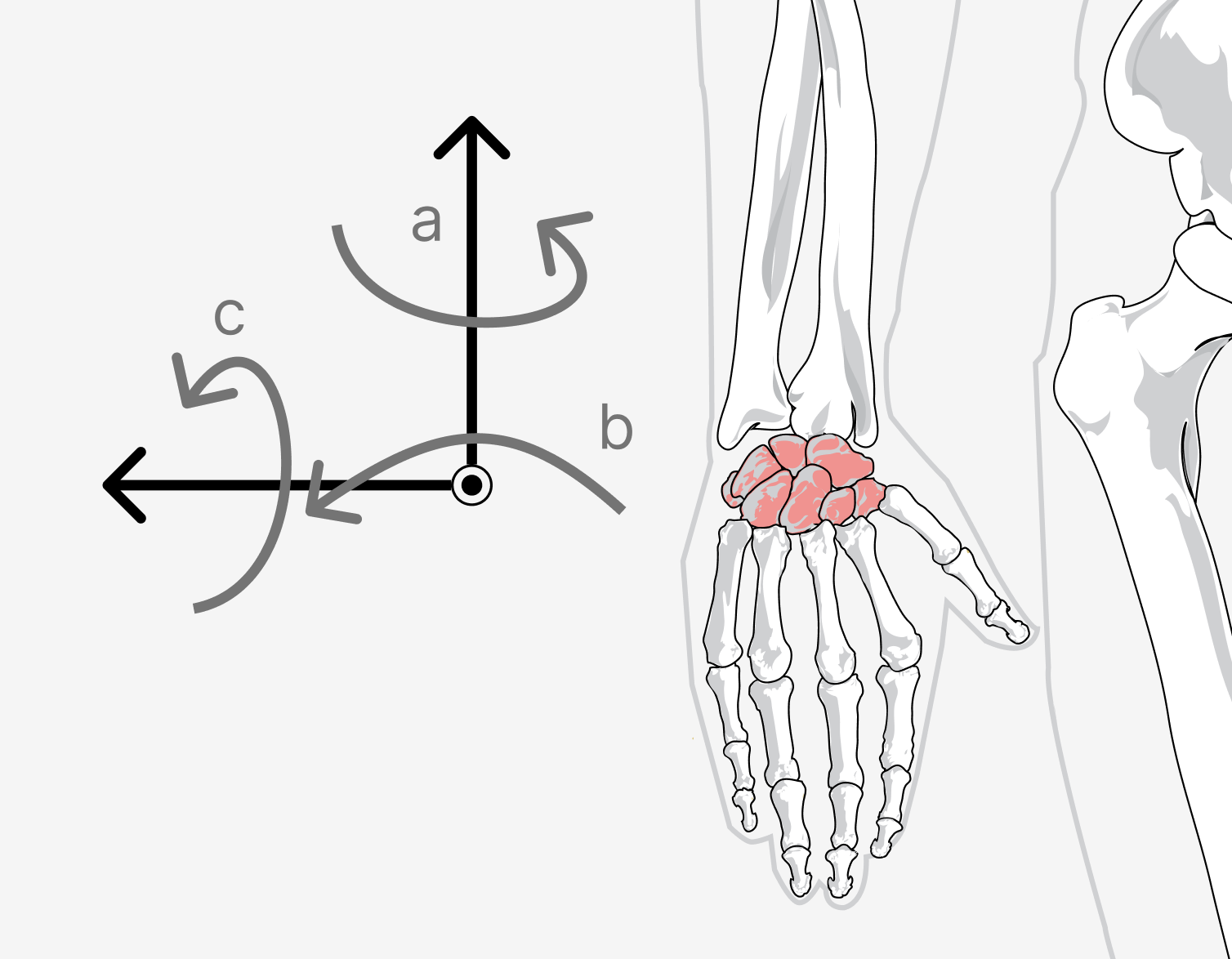}%
  }{%
    Wrist.\\
    (a) slight axial rotation\\
    (b) radial/ulnar deviation\\
    (c) flexion/extension.%
  }{fig:wrist}
  \caption{Upper body atlas II: Elbow and wrist degrees of freedom. Forearm pronation/supination (a, left) occurs at the proximal and distal radioulnar joints and is independent of wrist motion.}
  \label{fig:upper_II}
\end{figure}

\begin{figure}[!htbp]
  \centering
  \atlasimg{%
    \includegraphics[height=\panelh,keepaspectratio]{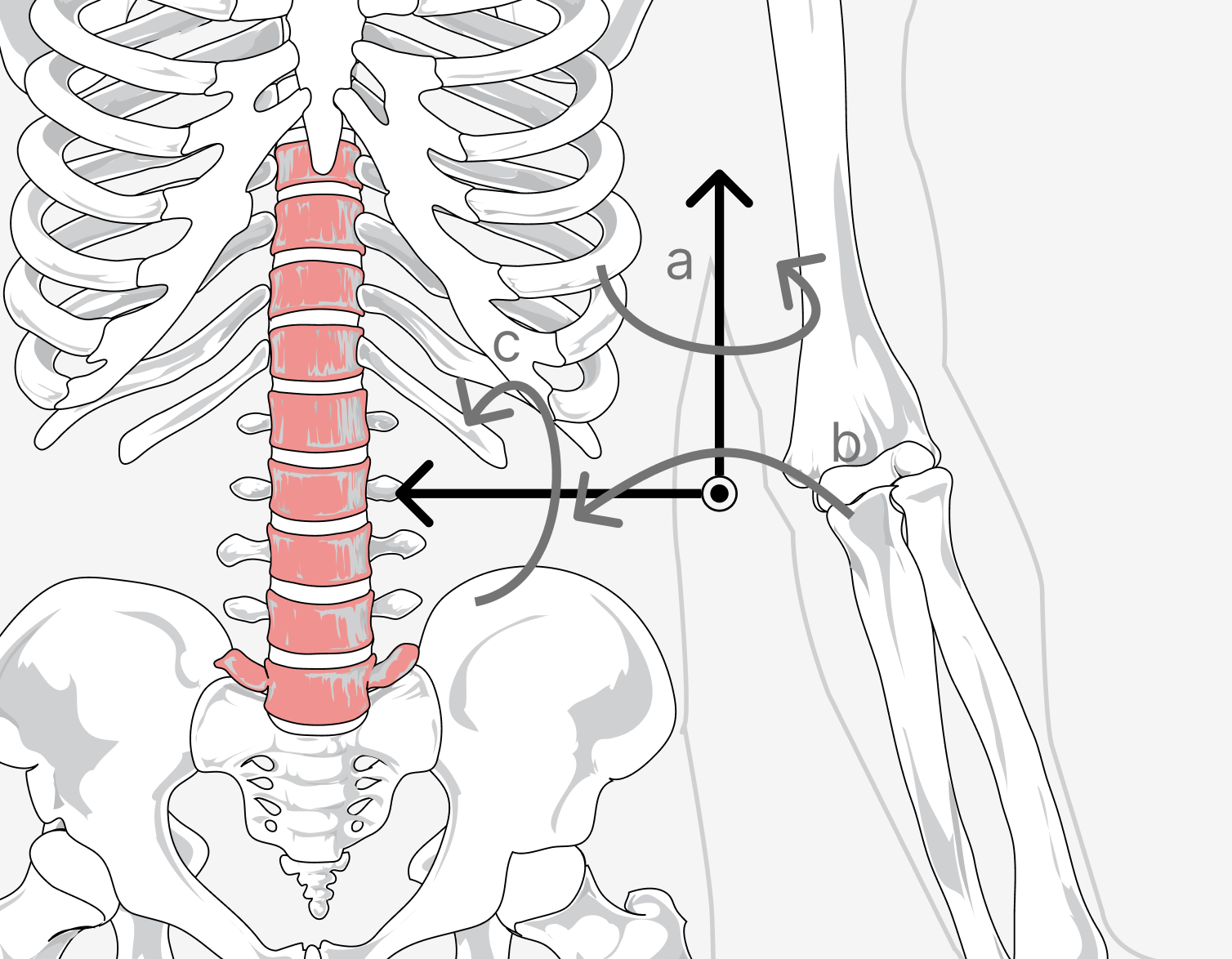}%
  }{%
    Spine.\\
    (a) axial rotation\\
    (b) lateral flexion\\
    (c) flexion/extension.%
  }{fig:spine}
  \hspace{\panelsep}
  \atlasimg{%
    \includegraphics[height=\panelh,keepaspectratio]{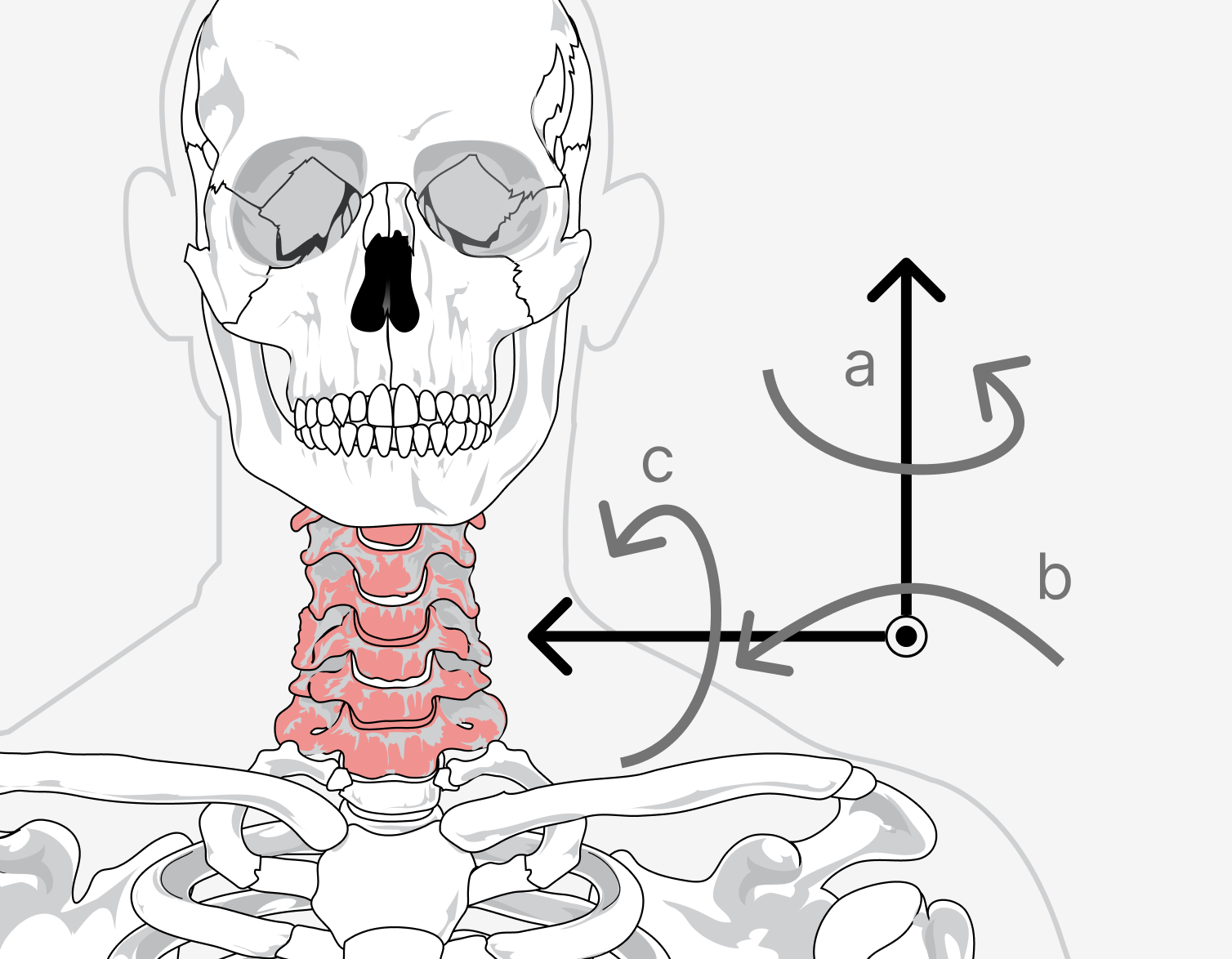}%
  }{%
    Neck.\\
    (a) axial rotation\\
    (b) lateral flexion\\
    (c) flexion/extension.%
  }{fig:neck}
  \caption{Upper body atlas III: Thoracolumbar spine and cervical spine (neck) degrees of freedom. Both regions support three primary rotations about their respective frames.}
  \label{fig:upper_III}
\end{figure}

The upper-body atlas (Figures~\ref{fig:upper_I}--\ref{fig:upper_III}) covers the shoulder complex, elbow and forearm, wrist, neck, and thoracolumbar spine. Key design choices:

\begin{itemize}[leftmargin=12pt,itemsep=2pt]
\item \textbf{Shoulder complex}: We separate glenohumeral (GH) rotations from scapulothoracic (ST) contributions to clarify which motions are actuated at the ball-and-socket joint versus achieved through scapular motion on the rib cage.
\item \textbf{Forearm}: Pronation/supination is treated as a dedicated forearm DoF rather than being folded into ``wrist rotation,'' consistent with anatomical function and clinical measurement.
\item \textbf{Wrist}: Three rotational DoFs (flexion/extension, radial/ulnar deviation, slight axial rotation) are specified, though the axial component is small and task-dependent.
\end{itemize}

These definitions establish the axes used when assigning joint weights $u_{j,t}$ for reaching and manipulation tasks in HLAS.

\FloatBarrier

\subsection{Lower body}
\label{subsec:lower_body}

\begin{figure}[!htbp]
  \centering
  \atlasimg{%
    \includegraphics[height=\panelh,keepaspectratio]{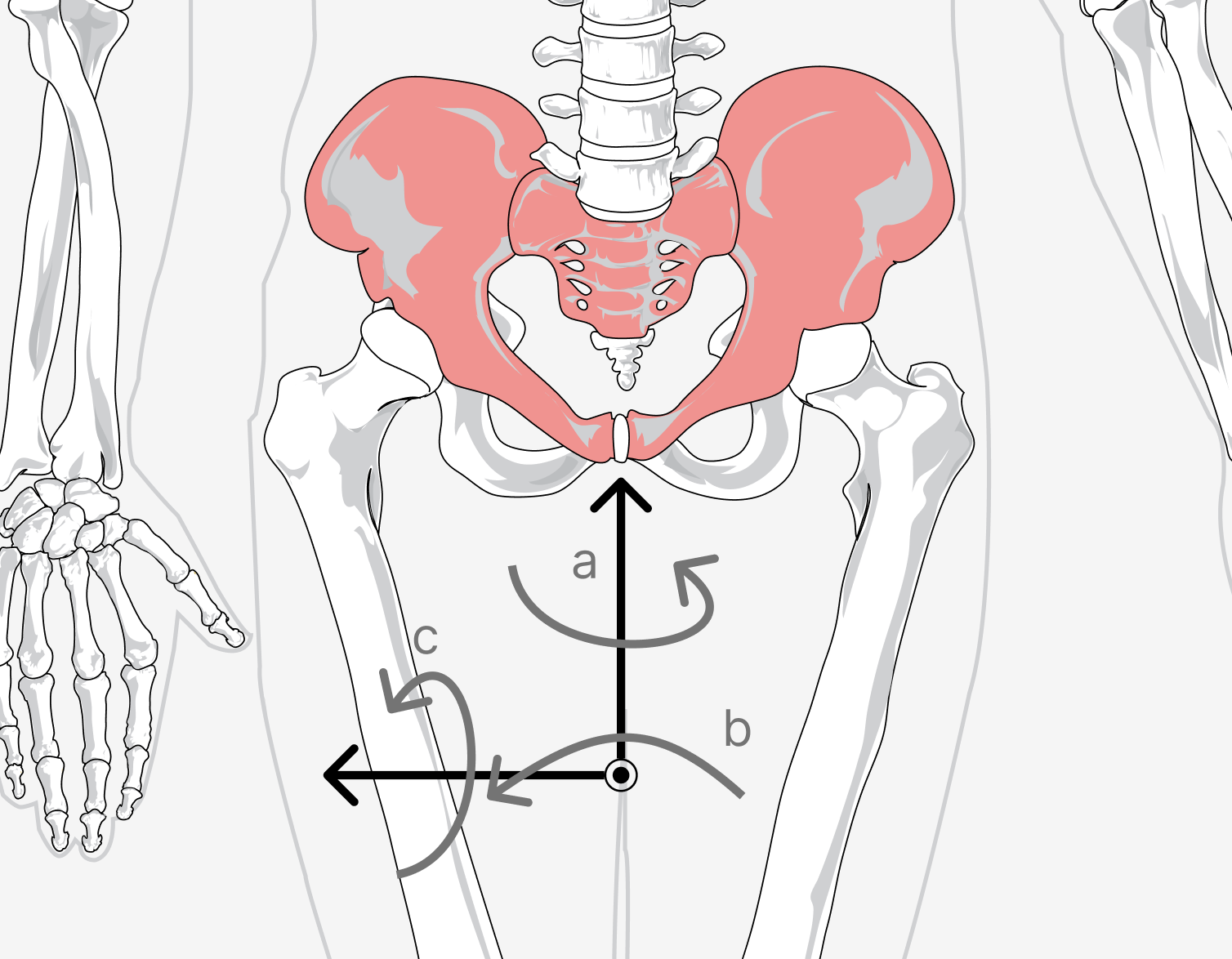}%
  }{%
    Pelvis.\\
    (a) axial rotation\\
    (b) lateral tilt\\
    (c) anterior/posterior tilt.%
  }{fig:pelvis}
  \hspace{\panelsep}
  \atlasimg{%
    \includegraphics[height=\panelh,keepaspectratio]{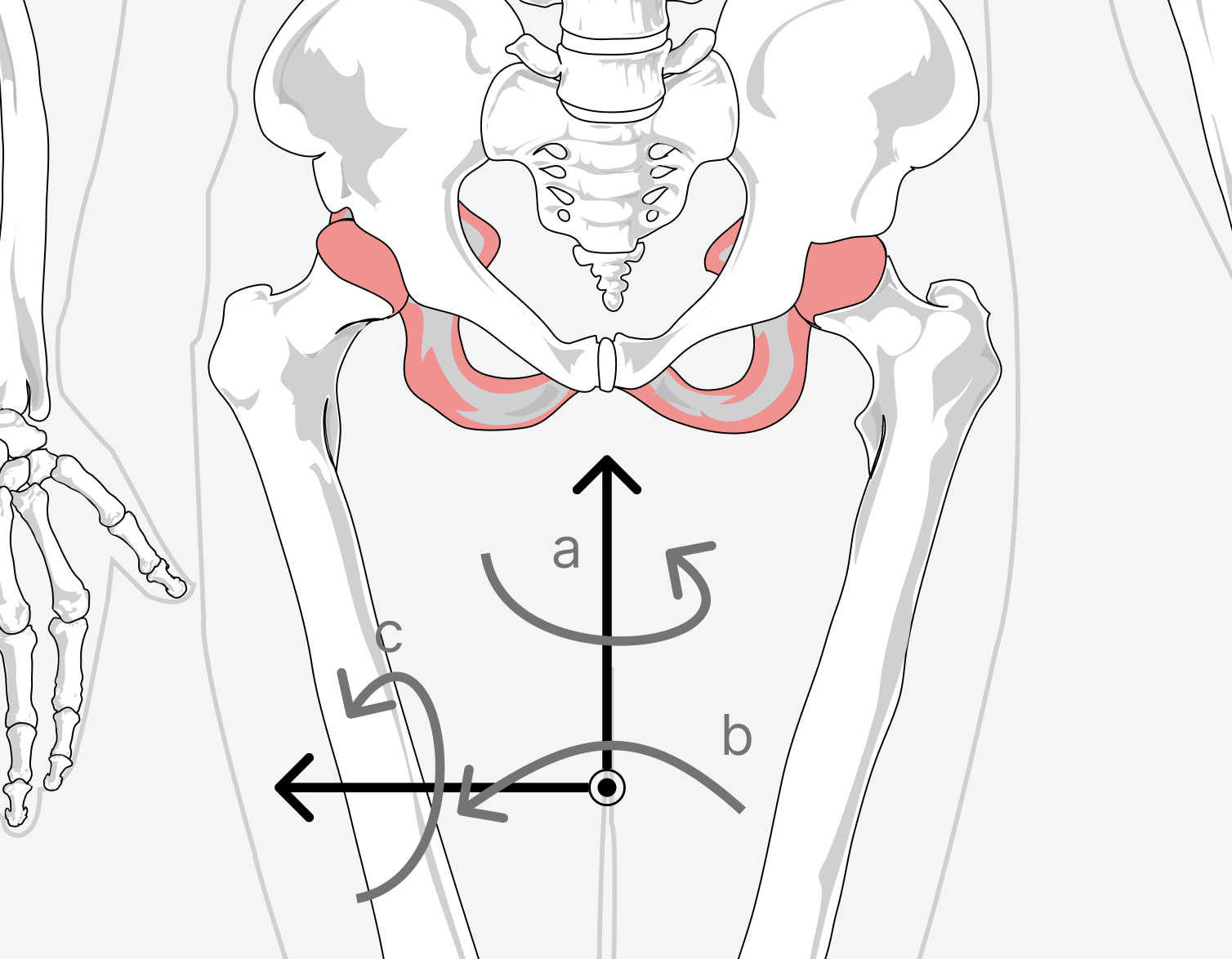}%
  }{%
    Hip.\\
    (a) external/internal rotation\\
    (b) abduction/adduction\\
    (c) extension/flexion.%
  }{fig:hip}
  \caption{Lower body atlas I: Pelvis and hip degrees of freedom. Pelvic motion is relative to a global or trunk frame while hip motion is relative to the pelvis.}
  \label{fig:lower_I}
\end{figure}

\begin{figure}[!htbp]
  \centering
  \atlasimg{%
    \includegraphics[height=\panelh,keepaspectratio]{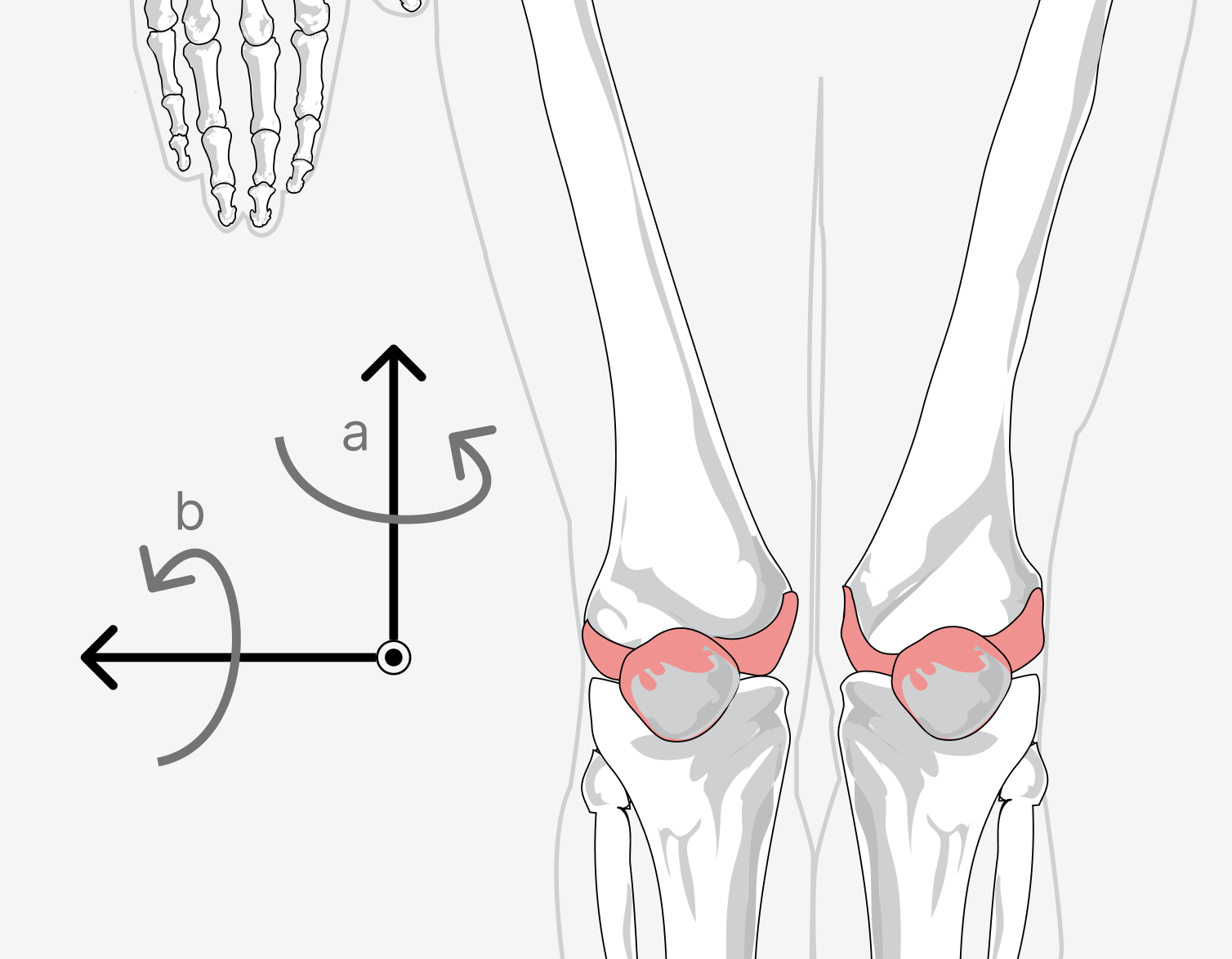}%
  }{%
    Knee.\\
    (a) slight axial rotation\\
    (b) flexion/extension.%
  }{fig:knee}
  \hspace{\panelsep}
  \atlasimg{%
    \includegraphics[height=\panelh,keepaspectratio]{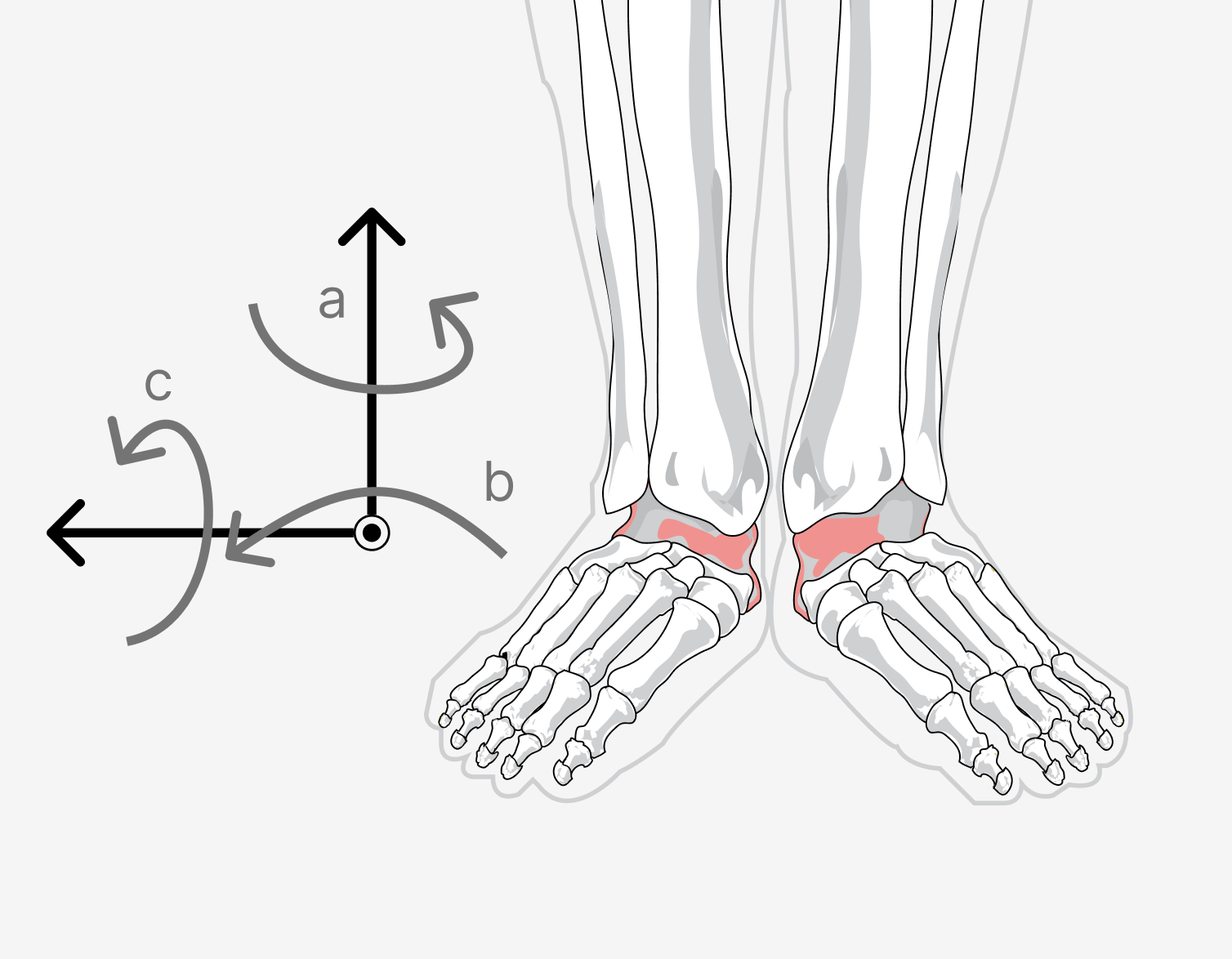}%
  }{%
    Ankle complex.\\
    (a) slight axial rotation \\
    (b) inversion/eversion (subtalar)\\
    (c) dorsiflexion/plantarflexion (talocrural).%
  }{fig:ankle}
  \caption{Lower body atlas II: Knee and ankle complex. The knee is primarily 1 DoF (flexion/extension) with small passive rotation. The ankle combines talocrural and subtalar contributions for three rotational DoFs.}
  \label{fig:lower_II}
\end{figure}

The lower-body atlas (Figures~\ref{fig:lower_I}--\ref{fig:lower_II}) specifies pelvic motion, hip rotations, knee flexion/extension, and the ankle complex. Key features:

\begin{itemize}[leftmargin=12pt,itemsep=2pt]
\item \textbf{Hip}: Three rotational DoFs (flexion/extension, abduction/adduction, internal/external rotation) modeled as a ball-and-socket joint.
\item \textbf{Knee}: Primarily a hinge joint (1 DoF: flexion/extension) with small passive axial rotation that couples to flexion angle. For HLAS purposes we treat the knee as 1 actuated DoF unless the robot explicitly controls knee rotation independently.
\item \textbf{Ankle complex}: Combines talocrural (dorsiflexion/plantarflexion) and subtalar (inversion/eversion) contributions, with small coupled axial rotation. All three components are included in the atlas. Task-specific operating bands $\mathcal{R}_{j,t}$ will emphasize the talocrural component for gait.
\end{itemize}

These definitions are the basis for placing gait and stair-related human torque/power curves onto joint-angle and angular-rate coordinates in Section~\ref{sec:biomech}.

\FloatBarrier

\subsection{Digits: fingers and toes}
\label{subsec:digits}

Digits account for a large fraction of the total DoF count in the atlas (68R out of 110 total DoFs, Table~\ref{tab:dof_inventory}). We specify:

\begin{itemize}[leftmargin=12pt,itemsep=2pt]
\item \textbf{Fingers}: Metacarpophalangeal (MCP), proximal interphalangeal (PIP), and distal interphalangeal (DIP) joints for digits 2--5, plus carpometacarpal (CMC), MCP, and interphalangeal (IP) joints for the thumb.
\item \textbf{Toes}: Metatarsophalangeal (MTP) and interphalangeal (IP) joints for the hallux and toes 2--5.
\end{itemize}

\textbf{Scope note.} While HLAS examples in this paper emphasize proximal joints (shoulder through ankle), the digit definitions are included for completeness and to support future work on dexterous manipulation and foot-grasping tasks. Users may omit digits from their HLAS computation if these tasks are not mission-relevant.

\begin{figure}[!htbp]
  \centering
  \atlasimg{%
    \includegraphics[height=\panelh,keepaspectratio]{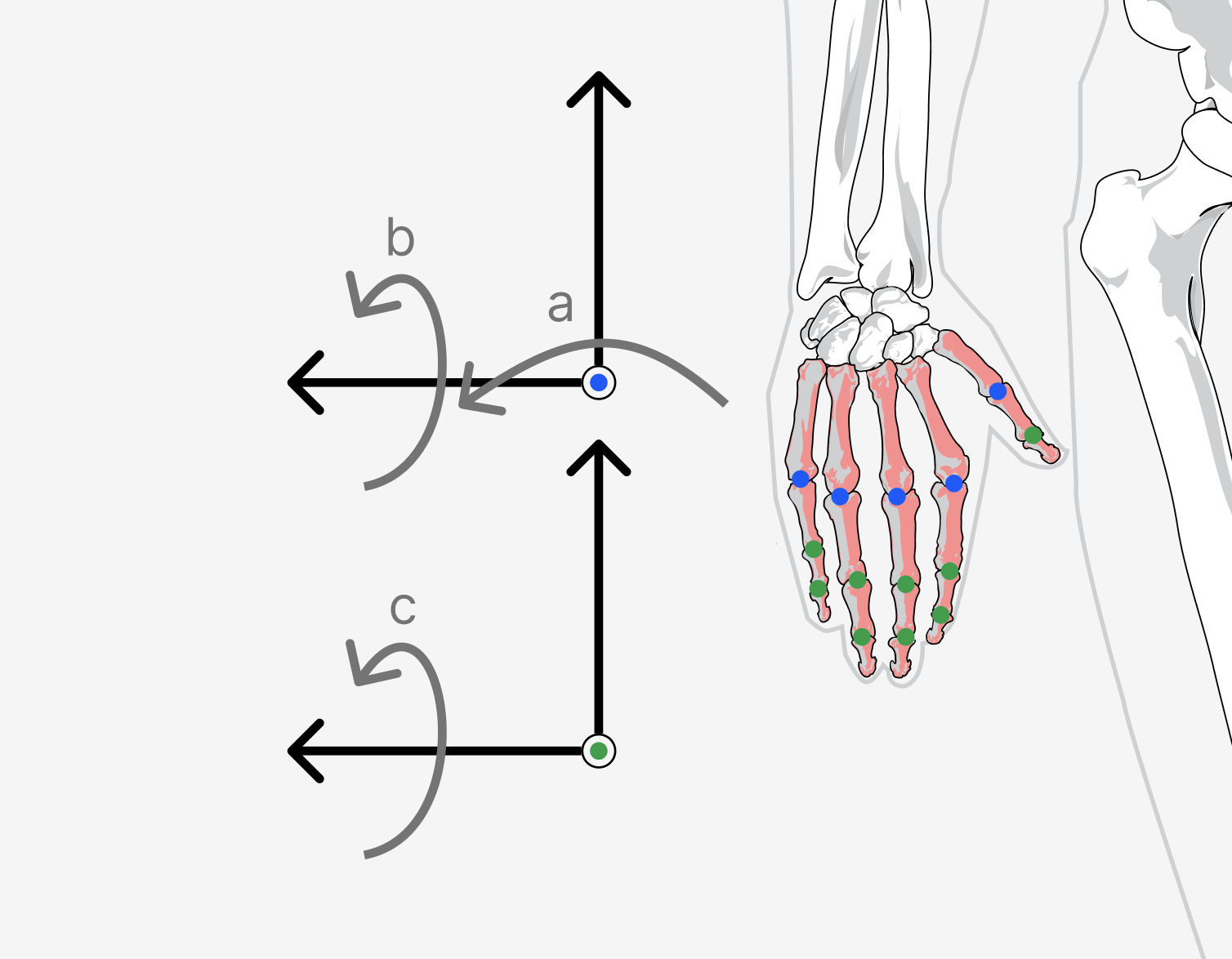}%
  }{%
    Fingers.\\
    (a) adduction/abduction\\
    (b) flexion/extension (MCP)\\
    (c) flexion/extension (PIP/DIP).%
  }{fig:fingers}
  \hspace{\panelsep}
  \atlasimg{%
    \includegraphics[height=\panelh,keepaspectratio]{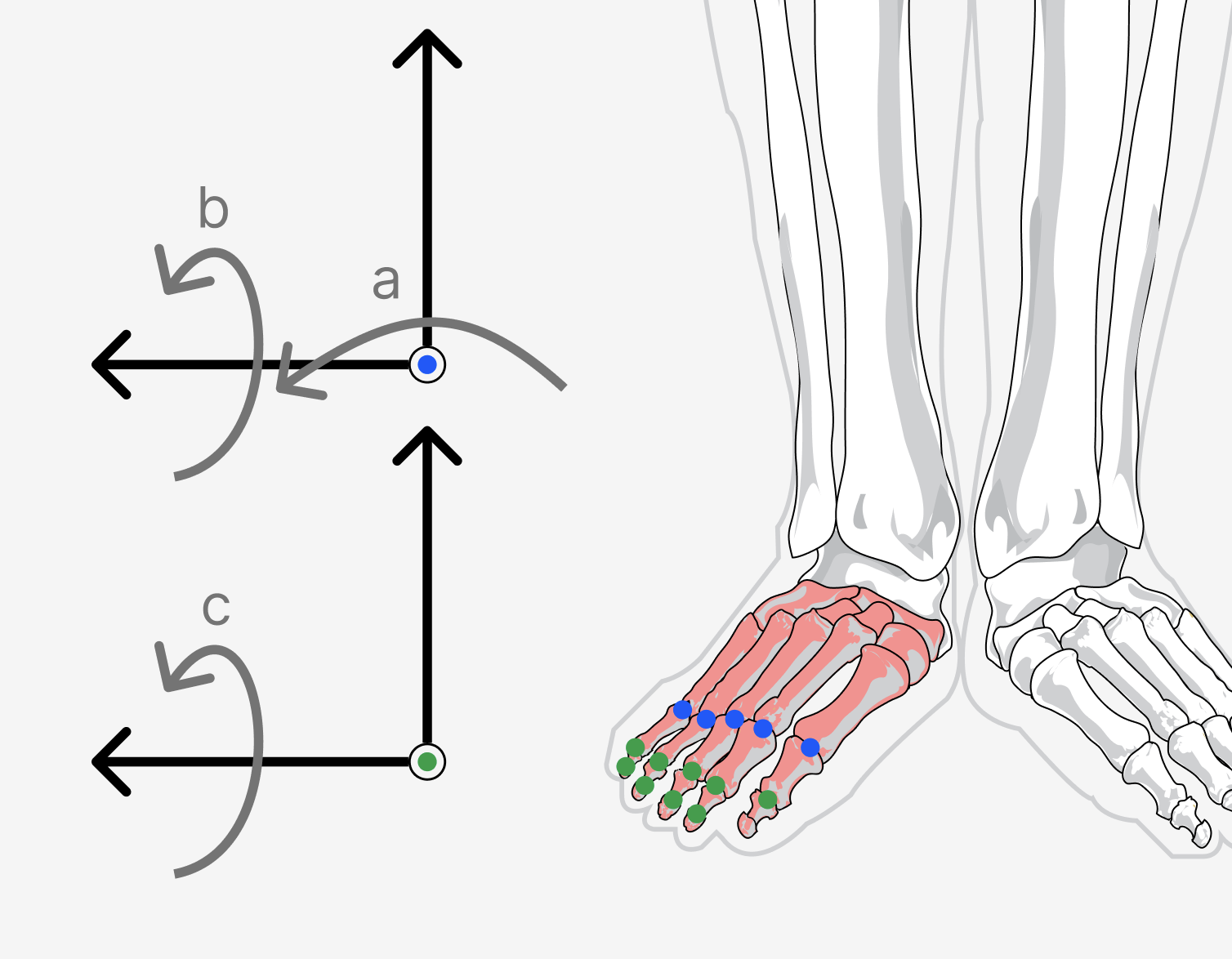}%
  }{%
    Toes.\\
    (a) adduction/abduction\\
    (b) flexion/extension (MTP)\\
    (c) flexion/extension (IP).%
  }{fig:toes}
  \caption{Digits atlas: Finger and toe degrees of freedom. Digits 2--5 have similar structure (MCP/MTP with flexion and adduction, IP joints with flexion only). The thumb and hallux have specialized geometry.}
  \label{fig:digits}
\end{figure}

\FloatBarrier

\subsection{DoF inventory and ROM norms}
\label{subsec:dof_rom_tables}

Figures~\ref{fig:upper_I} through \ref{fig:digits} define the qualitative geometry and axis directions. For quantitative HLAS scoring we also require: (i) a count of how many DoFs are represented at each joint, and (ii) the functional ROM intervals over which those DoFs are used in everyday tasks. Tables~\ref{tab:dof_inventory} and \ref{tab:rom_compact} provide these data.

\paragraph{DoF inventory (Table~\ref{tab:dof_inventory}).}
This table enumerates rotational (R) and translational (T) degrees of freedom per joint, with counts given per side and bilaterally. The total human DoF count is 106R and 4T (the latter from scapulothoracic elevation/depression and protraction/retraction). When computing the DoF sufficiency term $d^{\mathrm{DoF}}_{j,t}$ in HLAS (Eq.~\ref{eq:dof}), we check which subset of these DoFs are implemented and independently actuated on the robot for each task $t$.

\paragraph{ROM norms (Table~\ref{tab:rom_compact}).}
This table distinguishes three ROM categories:
\begin{itemize}[leftmargin=12pt,itemsep=2pt]
\item \textbf{AROM} (active ROM): Range achievable through voluntary muscle contraction.
\item \textbf{PROM} (passive ROM): Range achievable with external assistance (typically 5--10\% larger than AROM).
\item \textbf{Functional ROM}: Subset of AROM used in activities of daily living (ADL). This is the range that matters for HLAS.
\end{itemize}
Values are synthesized primarily from AAOS guidelines and clinical ROM studies \citep{AAOSROM2020,Morrey1981,Palmer1985}. When computing ROM coverage $\rho^{\mathrm{ROM}}_{j,t}$ (Eq.~\ref{eq:rom}), we use the functional intervals $I^{\mathrm{func}}_{j,t}(a)$ so that robots are not penalized for omitting extreme postures (e.g., maximum passive shoulder flexion of 180°) that are rarely used in everyday tasks, but are expected to cover the arcs that matter for gait, stairs, lifting, and reaching.

\begin{table}[t]
\small
\setlength{\tabcolsep}{6pt}\renewcommand{\arraystretch}{1.12}
\centering
\caption{Degrees of freedom represented in the atlas. Rotational (R) and translational (T) DoFs are shown per side where applicable. ST = scapulothoracic, GH = glenohumeral, TC = talocrural, MCP = metacarpophalangeal, MTP = metatarsophalangeal, PIP/DIP = proximal/distal interphalangeal, IP = interphalangeal.}
\label{tab:dof_inventory}
\begin{tabularx}{\linewidth}{@{}l l c c@{}}
\toprule
\textbf{Region} & \textbf{Joint(s)} & \textbf{DoF / side} & \textbf{Both sides} \\
\midrule
Head/Neck & Cervical spine & 3R & 3R \\
Trunk     & Thoracolumbar spine & 3R & 3R \\
\midrule
Upper limb & Shoulder GH                    & 3R       & 6R \\
           & Shoulder girdle ST (SC/AC)     & 1R+2T    & 2R+4T \\
           & Elbow (humeroulnar)            & 1R       & 2R \\
           & Forearm (PRU/DRU)              & 1R       & 2R \\
           & Wrist (RC/MC)                  & 3R       & 6R \\
           & Digits 2--5                    & 16R      & 32R \\
           & Thumb                          & 4R       & 8R \\
\midrule
Lower limb & Hip                            & 3R       & 6R \\
           & Knee                           & 1R       & 2R \\
           & Ankle complex (TC+ST)          & 3R       & 6R \\
           & Hallux                         & 2R       & 4R \\
           & Toes 2--5                      & 12R      & 24R \\
\midrule
\multicolumn{3}{@{}l}{\textbf{Totals}} 
  & \textbf{Both sides:} 106R,\; 4T \\
\bottomrule
\end{tabularx}
\end{table}

\begin{table}[t]
\scriptsize
\setlength{\tabcolsep}{3pt}\renewcommand{\arraystretch}{1.06}
\centering
\caption{Range of motion norms (degrees unless noted). AROM = active, PROM = passive, Func. = functional (activities of daily living). Values synthesized primarily from AAOS tables \citep{AAOSROM2020} and clinical literature \citep{Morrey1981,Palmer1985}. Wrist axial rotation and transverse-plane ankle rotation are small and often coupled; we list approximate arcs for completeness.}
\label{tab:rom_compact}
\begin{tabularx}{\linewidth}{@{}l Y c c c@{}}
\toprule
\textbf{Region} & \textbf{Motion} & \textbf{AROM} & \textbf{PROM} & \textbf{Func.} \\
\midrule
\multicolumn{5}{@{}l}{\textbf{Neck}}\\
Neck & Rot L/R & 80/80 & 85/85 & 60/60 \\ \rowsep
Neck & Flex/Ext & 50/70 & 60/80 & 40/50 \\ \rowsep
Neck & Lat flex L/R & 40/40 & 45/45 & 25/25 \\
\midrule
\multicolumn{5}{@{}l}{\textbf{Shoulder / Scapula}}\\
GH   & Flex/Ext & 180/60 & 180/60 & 120/40 \\ \rowsep
GH   & Abd & 170--180 & 180 & 120 \\ \rowsep
GH   & IR/ER @90 abd & 70/90 & 75/95 & 50/60 \\ \rowsep
ST   & Up rot & 50--60 & 60 & 30 \\ \rowsep
ST   & Elev/Depr (\textit{cm}) & 10--12 / 4--5 & -- & $\sim$50\% max \\ \rowsep
ST   & Pro/Ret (\textit{cm}) & 7--10 / 5--6 & -- & mid \\
\midrule
\multicolumn{5}{@{}l}{\textbf{Elbow / Forearm / Wrist}}\\
Elbow   & Flex/Ext & 150/0 & 150/0 & 30--130 \\ \rowsep
Forearm & Sup/Pro & 80--90 / 80--90 & 90/90 & $\ge$50 / $\ge$50 \\ \rowsep
Wrist   & Flex/Ext & 80/70 & 90/80 & 5/30 \\ \rowsep
Wrist   & Rad/Uln dev & 20/30 & 25/35 & 10/15 \\ \rowsep
Wrist   & Axial rot (small) & $\sim$10/10 & $\sim$15/15 & $\sim$5/5 \\
\midrule
\multicolumn{5}{@{}l}{\textbf{Hand}}\\
Digits 2--5 & MCP flex/abd & 80--90 / 20--25 & 90--100 / 25--30 & MCP flex 60--70 \\ \rowsep
Digits 2--5 & PIP/DIP flex & 100--110 / 70--80 & 110--120 / 80--90 & PIP 70--90, DIP 60--80 \\ \rowsep
Thumb & CMC abd/flex & 45 / 15--20 & 50 / 20--25 & opposition \\ \rowsep
Thumb & MCP/IP flex & 50 / 80 & 60 / 90 & IP 60--80 \\
\midrule
\multicolumn{5}{@{}l}{\textbf{Lower limb}}\\
Hip  & Flex/Ext & 120/20 & 125/30 & $\sim$100 / $\sim$10 \\ \rowsep
Hip  & Abd/Add  & 45/30 & 50/35 & $\sim$20 / $\sim$10 \\ \rowsep
Hip  & IR/ER    & 35/45 & 40/50 & $\sim$15 / $\sim$20 \\ \rowsep
Knee & Flex/Ext & 135/0 & 140/0 & 110/0 \\ \rowsep
Ankle complex (TC+ST) & DF/PF & 20/50 & 25/55 & 10/20 \\ \rowsep
Ankle complex (TC+ST) & Inv/Ev & 20--30 / 5--10 & 30--35 / 10--15 & 5 / 5 \\ \rowsep
Ankle complex (TC+ST) & Axial rot (small) & $\sim$10/10 & $\sim$15/15 & 5/5 \\ \rowsep
Hallux & MTP ext/flex & 70--90 / 30--45 & 90--100 / 45--50 & push-off 45--60 ext \\ \rowsep
Toes 2--5 & MTP/PIP/DIP flex & 40 / 50--70 / 30--60 & +10 each (typ.) & moderate composite \\
\bottomrule
\end{tabularx}
\end{table}

\subsection{Using the atlas in practice}

To apply this atlas when computing HLAS for a specific robot:

\begin{enumerate}[leftmargin=12pt,itemsep=2pt]
\item \textbf{Map robot joints to atlas axes.} Identify which anatomical DoFs each robot actuator implements. For example, a robot ankle with a single sagittal-plane actuator corresponds to the talocrural dorsiflexion/plantarflexion axis (c) in Figure~\ref{fig:lower_II}, while a 2-DoF ankle adds subtalar inversion/eversion (b).

\item \textbf{Verify sign conventions.} Ensure that positive robot joint angles correspond to positive anatomical motions as defined above (e.g., positive ankle angle = dorsiflexion, not plantarflexion). If necessary, apply sign flips or offsets in the control software to align with ISB conventions.

\item \textbf{Check ROM coverage.} For each task-used axis $a \in \mathcal{A}_{j,t}$, compare the robot's safe ROM $I^{\mathrm{rob}}_j(a)$ to the functional human interval $I^{\mathrm{func}}_{j,t}(a)$ from Table~\ref{tab:rom_compact}. Compute overlap as in Eq.~\ref{eq:rom}.

\item \textbf{Verify DoF availability.} For each required axis, confirm that it is independently actuated (not passive or coupled) by commanding small test motions and checking for cross-axis coupling. Set $d^{\mathrm{DoF}}_{j,t} = 1$ only if all task-required axes pass this test (Eq.~\ref{eq:dof}).
\end{enumerate}

By following these steps, designers ensure that their robot specifications are directly comparable to human capabilities in a shared coordinate system, enabling meaningful HLAS scores.

\section{Human Reference and Biomechanical Targets}
\label{sec:biomech}

\subsection{Purpose and data sources}

This section compiles human joint capability data from published biomechanics literature and converts it into quantitative actuator requirements for HLAS. We synthesize representative values for joint moments, mechanical power, angular rates, and metabolic demands across functional tasks, rescale them to a common reference body, and organize them into the per-joint demand fields $T^{\mathrm{hum}}_{j,t}(q,\omega)$ and $P^{\mathrm{hum}}_{j,t}(q,\omega)$ that define Human-Equivalence Envelopes (Section~\ref{sec:hlas}).

\textbf{Important:} We do not collect new human data. All values are extracted from prior biomechanical studies, with references provided for each quantity. Our contribution is the synthesis, normalization, and mapping to task-specific operating bands $\mathcal{R}_{j,t}$ that feed HLAS.

\subsection{Reference body and scaling}

\paragraph{Reference subject specification.}
We adopt a 75\,kg, 175\,cm (5'9'') adult male as the reference subject. This choice aligns with common humanoid target anthropometries (e.g., typical adult male body mass and stature in developed countries) and matches the normalization conventions in canonical gait datasets \citep{Winter2009,Farris2012,Zelik2016}. 

Segment masses, centers of mass, and moments of inertia are computed using de~Leva's adjustments to the Zatsiorsky--Seluyanov regression model \citep{DeLeva1996}, which provides validated parameters for inverse dynamics and biomechanical scaling. These parameters are necessary to (i) convert joint moments from inverse dynamics to actuator torques accounting for gravity and inertia, and (ii) ensure consistent interpretation of normalized literature values.

\paragraph{Normalization and scaling formulas.}
Human joint capabilities in the literature are typically reported as mass-normalized quantities: moments in Nm/kg and mechanical power in W/kg. We convert these to absolute values for the 75\,kg reference via:
\[
T_{\mathrm{abs}} = m \cdot (T/m), \qquad P_{\mathrm{abs}} = m \cdot (P/m),
\]
where $m{=}75$\,kg. For example, an ankle plantarflexion moment of 1.5\,Nm/kg during push-off becomes $T_{\mathrm{abs}}{=}75{\times}1.5{=}112.5$\,Nm.

\paragraph{Generalizing to other body sizes.}
While we use a single reference body for concreteness, the framework generalizes straightforwardly. To target a different mass $m'$ or morphology, substitute $m'$ in the scaling equations and recompute segment parameters using the same de~Leva regressions. The human demand fields $T^{\mathrm{hum}}_{j,t}(q,\omega)$ and $P^{\mathrm{hum}}_{j,t}(q,\omega)$ scale proportionally, so HLAS can be computed for any target body size. Section~\ref{sec:limitations} discusses population variability and extensions to age, gender, and pathological conditions.

\subsection{Task families and joint contributions}

We organize human capability data around six representative task families that span steady-state locomotion, vertical work, manipulation, and safety-relevant extremes:

\begin{enumerate}[leftmargin=12pt,itemsep=2pt]
\item \textbf{Level walking} (1.2--1.4\,m/s): Steady-state locomotion at preferred speed, emphasizing ankle positive power during push-off and knee/hip support torque during stance.

\item \textbf{Moderate running} (2.5--3.0\,m/s): Higher-speed locomotion with increased joint power and duty-cycle demands, testing actuator efficiency and thermal capacity under repeated loading.

\item \textbf{Stair ascent and descent}: Quasi-static vertical work requiring sustained knee and hip torque at low angular rates, with ankle propulsion bursts. Tests thermal endurance and low-speed torque delivery.

\item \textbf{Repetitive lifting} (floor to waist): Manual handling task emphasizing sustained knee and hip extensor torque over minutes-long duty cycles. Relevant for logistics and assistance applications.

\item \textbf{Reaching with payload} (shoulder height, 5\,kg): Upper-limb manipulation under load, testing shoulder and elbow torque with precision endpoint control. Relevant for object handling and assembly tasks.

\item \textbf{Fast hand actions and ballistic extremes}: Rapid finger/wrist motions (4--6\,Hz) for dexterous manipulation, and ballistic shoulder internal rotation during throwing (up to $\sim$7{,}700°/s, $\approx$130\,rad/s) to establish safe rate limits for interaction.
\end{enumerate}

Each task defines a subset of contributing joints $\mathcal{J}_t$ and task-specific operating bands $\mathcal{R}_{j,t}$ over joint angle $q$ and angular rate $\omega$. These bands are derived from phase-resolved kinematic and kinetic data in the cited studies (e.g., gait cycle for walking, reach trajectory for manipulation).

\subsection{Joint-level capability envelopes}

Table~\ref{tab:joint_level_targets} presents per-joint human capability envelopes: moments, positive mechanical power, and angular rates at specific task phases where each joint produces significant positive work. These values define the target fields $T^{\mathrm{hum}}_{j,t}(q,\omega)$ and $P^{\mathrm{hum}}_{j,t}(q,\omega)$ used in the HEE simultaneity test (Eq.~\ref{eq:hee}).

Key observations:

\begin{itemize}[leftmargin=12pt,itemsep=2pt]
\item \textbf{Ankle dominates walking power.} During level walking, the ankle produces roughly 190--260\,W of positive power near push-off (2.5--3.5\,W/kg), far exceeding knee and hip contributions at this phase. This motivates high ankle power delivery in the 8--12\,rad/s band.

\item \textbf{Knee and hip provide stance support.} During early stance and stair ascent, the knee and hip produce sustained torque (35--95\,Nm for knee, 38--75\,Nm for hip) at low angular rates. This requires continuous-torque capability with thermal headroom rather than peak bursts.

\item \textbf{Upper-limb rates span orders of magnitude.} Functional hand motions operate at 4--6\,Hz, while ballistic shoulder internal rotation during throwing reaches 113--134\,rad/s. This wide range informs both torque-mode bandwidth targets and safe rate limits for interaction.
\end{itemize}

\textbf{Measurement context.} All values are reported at task-relevant postures, not isometric extremes. For example, ankle moments are given at plantarflexion angles used during push-off (roughly 0--25°), not at maximum voluntary contraction in a fixed posture. This ensures that robot tests evaluate performance where humans actually operate.

\begin{table*}[t]
\scriptsize
\setlength{\tabcolsep}{4pt}\renewcommand{\arraystretch}{1.08}
\centering
\caption{\textbf{Joint-level human capability envelopes.} Representative joint moments, positive mechanical power, and angular rates for a 75\,kg reference subject at specific task phases where each joint produces significant positive work. These values define the per-joint human demand fields $T^{\mathrm{hum}}_{j,t}(q,\omega)$ and $P^{\mathrm{hum}}_{j,t}(q,\omega)$ used in the HEE simultaneity test (Eq.~\ref{eq:hee}). Normalized values (Nm/kg, W/kg) are shown first, with absolute values (Nm, W) for the 75\,kg reference in parentheses. Compiled from published biomechanics studies as cited.}
\label{tab:joint_level_targets}
\begin{tabularx}{\linewidth}{@{}l c c c@{}}
\toprule
\textbf{Joint / context (phase)} &
\textbf{Moment (Nm/kg $\to$ Nm)} &
\textbf{Power$^{+}$ (W/kg $\to$ W)} &
\textbf{Peak rate (units)} \\
\midrule
Hip (early stance H1, walking) &
0.5--1.0 $\to$ 38--75 &
0.7--1.0 $\to$ 53--75 &
--- \\
Knee (terminal swing, walking) &
0.46--1.27 $\to$ 35--95 &
0.5--0.6 $\to$ 38--45 &
--- \\
Ankle (push-off A2, walking) &
1.2--1.9 $\to$ 90--143 &
2.5--3.5 $\to$ 190--260 &
--- \\
Knee (support, stair ascent) &
$\sim$1.0 $\to$ $\sim$75 &
--- &
--- \\
Ankle (propulsion, stair ascent) &
1.4--1.8 $\to$ 105--135 &
3.0--3.5 $\to$ 225--260 &
--- \\
Shoulder (ballistic IR, throwing) &
--- & --- &
6{,}500--7{,}700°/s ($\approx$113--134\,rad/s) \\
Finger/wrist (rapid open/close) &
--- & --- &
4--6\,Hz \\
\midrule
\multicolumn{4}{l}{\scriptsize \textit{Sources:} \citepalias{Farris2012}, \citepalias{Winter2009}, \citepalias{Zelik2016}, \citepalias{Halsey2012StairClimbing}, \citepalias{Seroyer2010}, \citepalias{Shimoyama1990}.}\\
\bottomrule
\end{tabularx}
\end{table*}

\subsection{Task-level load distribution}

Table~\ref{tab:task_level_targets} summarizes how mechanical and metabolic demands are distributed across joints for each task. This table informs three key HLAS components:

\begin{enumerate}[leftmargin=12pt,itemsep=2pt]
\item \textbf{Task set $\mathcal{T}$}: The rows define the functional tasks used in HLAS scoring.
\item \textbf{Task weights $w_t$}: The relative importance of each task (e.g., $w_{\texttt{Walk}}{=}0.4$ if walking is mission-critical).
\item \textbf{Joint weights $u_{j,t}$}: The distribution of mechanical responsibility within each task (e.g., ankle contributes 50\% of walking positive work, so $u_{\texttt{ankle,Walk}}{=}0.5$).
\end{enumerate}

These weights are derived from the positive mechanical power or torque-time integrals reported in the cited studies. For example, in level walking the ankle produces roughly 45--50\% of the net positive mechanical work over a stride, with the remainder split primarily between the hip (35--40\%) and a smaller contribution from the knee (10--20\%). These proportions become the joint weights $u_{j,t}$ in Eq.~\ref{eq:hlas-expanded}.

\begin{table*}[t]
\scriptsize
\setlength{\tabcolsep}{4pt}\renewcommand{\arraystretch}{1.08}
\centering
\caption{\textbf{Task-level load distribution across joints.} Representative functional tasks and the joints primarily responsible for delivering torque, positive mechanical power, or high angular rate. These joint--task pairings define the task set $\mathcal{T}$, and the power/torque distributions inform the joint weights $u_{j,t}$ and task weights $w_t$ in HLAS (Eq.~\ref{eq:hlas-expanded}). Normalized values (Nm/kg, W/kg) are shown first, with absolute values (Nm, W) for 75\,kg in parentheses. Compiled from published studies as cited.}
\label{tab:task_level_targets}
\begin{tabularx}{\linewidth}{@{}l c c c c@{}}
\toprule
\textbf{Task (typical speed)} &
\textbf{Joint(s)} &
\textbf{Moment (Nm/kg $\to$ Nm)} &
\textbf{Power$^{+}$ (W/kg $\to$ W)} &
\textbf{Peak rate (units)} \\
\midrule
Level walk (1.2--1.4\,m/s) & Ankle (push-off) &
--- & 2.5--3.5 $\to$ 190--260 & --- \\
Level walk (1.2--1.4\,m/s) & Knee/hip (support) &
knee 0.46--1.27 $\to$ 35--95 & --- & --- \\
Moderate run (2.5--3.0\,m/s) & Ankle, hip &
--- & 4--7 $\to$ 300--525 & --- \\
Stair ascent & Knee/hip, ankle &
knee $\sim$1.0 $\to$ $\sim$75 & 3.0--3.5 $\to$ 225--260 & 0.5--2\,rad/s \\
Repetitive lifts (floor$\to$waist) & Knee/hip &
knee 3.0--3.4 $\to$ 225--255 & --- & --- \\
Fast hand actions (open/close) & Finger/wrist &
--- & --- & 4--6\,Hz \\
Ballistic shoulder IR (throw) & Shoulder &
--- & --- & 6{,}500--7{,}700°/s \\
\midrule
\multicolumn{5}{l}{\scriptsize \textit{Sources:} \citepalias{Zelik2016}, \citepalias{Umberger2003}, \citepalias{Winter2009}, \citepalias{Farris2012}, \citepalias{Kram1990}, \citepalias{Gottschall2003}, \citepalias{Halsey2012StairClimbing}, \citepalias{Sarabon2021Knee}, \citepalias{Baltrusch2019}, \citepalias{Shimoyama1990}, \citepalias{Seroyer2010}.} \\
\bottomrule
\end{tabularx}
\end{table*}

\subsection{Task-specific actuator implications}

We now translate the compiled data into concrete actuator requirements, organized by task family.

\paragraph{Level walking (1.2--1.4\,m/s).}
Normative gait datasets consistently show ankle-dominated positive power bursts near push-off, with typical values of 2.5--3.5\,W/kg (190--260\,W for 75\,kg) \citep{Zelik2016}. Preferred walking speed clusters around 1.2--1.4\,m/s \citep{Ralston1958,Umberger2003}, where net metabolic power is 3--4\,W/kg after subtracting standing baseline \citep{Ortega2008,Umberger2003}. Peak joint angular velocities occur at knee terminal swing and ankle push-off, typically 8--12\,rad/s at the ankle during push-off \citep{Mentiplay2018}.

\textbf{Actuator requirements:}
\begin{itemize}[leftmargin=12pt,itemsep=2pt]
\item \textbf{Ankle}: Continuous delivery of 190--260\,W in the 8--12\,rad/s band with ${\ge}70\%$ electrical-to-mechanical efficiency. Must sustain this over hundreds of strides (thermal endurance).
\item \textbf{Knee/hip}: Continuous torque in low-rate bands ($<3$\,rad/s) for quasi-static support during stance, with plateaus around 35--75\,Nm (knee) and 38--75\,Nm (hip).
\end{itemize}

\paragraph{Moderate running (2.5--3.0\,m/s).}
Metabolic power scales with body weight and inversely with ground contact time \citep{Kram1990}. Representative measurements report net metabolic power of 6--9\,W/kg for level treadmill running \citep{Gottschall2003}, with joint power redistribution toward ankle and hip at higher speeds \citep{Farris2012}.

\textbf{Actuator requirements:}
\begin{itemize}[leftmargin=12pt,itemsep=2pt]
\item \textbf{Ankle/hip}: Repeated positive/negative power cycling (300--525\,W peaks) with high efficiency at mid-to-high rates. Must handle duty cycles with brief (100--200\,ms) ground contacts.
\item \textbf{Thermal}: Higher average power than walking, thermal management becomes critical for sustained running.
\end{itemize}

\paragraph{Stairs and vertical work.}
The mechanical power required for vertical ascent is $P_{\mathrm{mech}}/m {=} g v_z$, where $v_z$ is vertical speed. At representative speeds of 0.25--0.35\,m/s, this yields 2.5--3.5\,W/kg mechanical power. Measured metabolic power is higher (gross $\sim$6.0--6.5\,W/kg for 75\,kg) due to swing-phase work and inefficiencies \citep{Halsey2012StairClimbing}.

\textbf{Actuator requirements:}
\begin{itemize}[leftmargin=12pt,itemsep=2pt]
\item \textbf{Knee/hip}: Sustained torque plateaus at low angular rates (0.5--2\,rad/s) with thermal headroom for multi-second duty. Knee moments around 75\,Nm, hip moments 38--75\,Nm.
\item \textbf{Ankle}: Propulsion bursts (225--260\,W, 3.0--3.5\,W/kg) at push-off, similar to walking but with different timing and duty.
\end{itemize}

\paragraph{Repetitive lifting (floor to waist).}
Box-lifting studies report net metabolic power around 5\,W/kg for repetitive lifts in young adults \citep{Baltrusch2019}. Knee extension moments at mid-range angles (60--90° flexion) are approximately 3.0--3.4\,Nm/kg (225--255\,Nm for 75\,kg) during the lift phase \citep{Sarabon2021Knee}.

\textbf{Actuator requirements:}
\begin{itemize}[leftmargin=12pt,itemsep=2pt]
\item \textbf{Knee/hip}: Minutes-long continuous torque near mid-range angles with ${\sim}2{\times}$ peak margin for transients (e.g., 150\,Nm continuous, 300\,Nm burst).
\item \textbf{Thermal}: Must sustain plateau torque without derating over 2--5\,min duty cycles.
\end{itemize}

\paragraph{Upper-limb manipulation and safety bounds.}
Overhead throwing produces extreme shoulder internal rotation rates (6{,}500--7{,}700°/s, $\approx$113--134\,rad/s) \citep{Seroyer2010}, establishing an upper bound for safe joint rates in human-interactive systems. Everyday reaching motions are far slower (peak rates $<$10\,rad/s) but demand high transparency (low backdrive torque) and gentle torque limits. Functional hand rates for finger tapping or rapid open/close are 4--6\,Hz \citep{Shimoyama1990}.

\textbf{Actuator requirements:}
\begin{itemize}[leftmargin=12pt,itemsep=2pt]
\item \textbf{Shoulder/elbow}: Safe rate limits ${\ge}20$\,rad/s for everyday manipulation, higher rates reserved for ballistic motions with safety monitoring. High backdrivability for compliant interaction.
\item \textbf{Wrist/hand}: Torque-mode bandwidth ${\ge}5$--10\,Hz to track rapid motions. Low friction and reflected inertia for dexterous control. Current-limited safety for human contact.
\end{itemize}

\subsection{Summary: from biomechanics to actuator specs}

The compiled human data (Tables~\ref{tab:joint_level_targets} and \ref{tab:task_level_targets}) provide the quantitative anchors for HLAS:

\begin{enumerate}[leftmargin=12pt,itemsep=2pt]
\item \textbf{Target torques and powers} $T^{\mathrm{hum}}_{j,t}(q,\omega)$ and $P^{\mathrm{hum}}_{j,t}(q,\omega)$ define the human envelopes that robots must match in the HEE test (Eq.~\ref{eq:hee}).

\item \textbf{Operating bands} $\mathcal{R}_{j,t}$ specify where these targets apply in $(q,\omega)$ space, derived from phase-resolved task data.

\item \textbf{Joint and task weights} $u_{j,t}$ and $w_t$ reflect the relative contributions of each joint to positive mechanical work, informing the aggregation in Eq.~\ref{eq:hlas-expanded}.

\item \textbf{Efficiency and thermal targets} translate metabolic demands and duty cycles into continuous-safe performance requirements with specified thermal margins.
\end{enumerate}

Section~\ref{sec:hlas} uses these targets to define the Human-Level Actuation Score, and Sections~\ref{sec:hlas_protocols} and \ref{sec:benchmarks} detail how to measure whether a robot meets them.

\section{Human-Level Actuation Score (HLAS) with HEE}
\label{sec:hlas}

\subsection{Motivation and design philosophy}

Comparing humanoid actuation to human capability is straightforward when looking at isolated peaks (e.g., maximum ankle torque at an unspecified posture) but becomes genuinely difficult once we consider where those peaks occur, how long they can be sustained, how cleanly they are rendered under contact, and whether they reflect the operating points humans actually use. Our goal is a single, interpretable score that quantifies ``human-level actuation'' across multiple tasks and joints while preserving enough internal structure to diagnose specific strengths and weaknesses.

We introduce the \emph{Human-Level Actuation Score (HLAS)}, which aggregates six physically motivated factors: workspace coverage (ROM and DoF), simultaneous torque-power delivery (via \emph{Human-Equivalence Envelopes}, HEE), torque-mode bandwidth, task-weighted efficiency, and thermal sustainability. The score is normalized so that a representative human achieves $\mathrm{HLAS}{=}1.0$ on their own operating bands, and it decomposes into task- and joint-level contributions for root-cause analysis.

\textbf{Key design principle:} HLAS credits performance only where it matters. The HEE component (Eq.~\ref{eq:hee}) requires that robots meet human torque \emph{and} power requirements simultaneously at the \emph{same} joint angle and angular rate $(q,\omega)$, weighted by the positive mechanical work humans produce at each point. This simultaneity condition prevents gaming strategies such as achieving high torque at low speed and high speed at low torque, then averaging the two to claim ``human-level'' performance without ever operating at a human-like point.

\subsection{Framework components and notation}

\paragraph{Task set and weights.}
Let $\mathcal{T}$ denote the set of functional tasks evaluated in HLAS (e.g., level walking, stair ascent, reaching with 5\,kg payload). Each task $t \in \mathcal{T}$ carries a nonnegative weight $w_t$ with $\sum_{t\in\mathcal{T}} w_t{=}1$, reflecting mission importance or prevalence. For example, a logistics robot might assign $w_{\texttt{Walk}}{=}0.4$, $w_{\texttt{Stairs}}{=}0.3$, and $w_{\texttt{Lift}}{=}0.3$, while a manipulation-focused platform might emphasize reaching and hand tasks.

\paragraph{Contributing joints and joint weights.}
For each task $t$, let $\mathcal{J}_t$ denote the set of joints that contribute to the task (e.g., $\mathcal{J}_{\texttt{Walk}}{=}\{\text{ankle, knee, hip}\}$). Each joint $j \in \mathcal{J}_t$ receives a weight $u_{j,t}$ that reflects its share of the total positive mechanical work for that task, with $\sum_{j\in\mathcal{J}_t} u_{j,t} {=} 1$ and $u_{j,t} \ge 0$. These weights are derived from the biomechanics data in Section~\ref{sec:biomech}. For instance, in level walking the ankle produces approximately 50\% of the net positive work, so $u_{\texttt{ankle,Walk}} \approx 0.5$.

\paragraph{Operating bands.}
Each joint-task pair $(j,t)$ defines a task-specific operating band $\mathcal{R}_{j,t} \subseteq \{(q,\omega)\}$ in joint angle and angular rate space. These bands are extracted from published biomechanics studies (Section~\ref{sec:biomech}) and capture the posture-rate regions where humans produce positive mechanical power for task $t$ at joint $j$. For example, the ankle push-off band during walking covers roughly $0$--$25^{\circ}$ plantarflexion at 8--12\,rad/s, where the ankle delivers most of its positive work. By restricting evaluation to these bands, we ensure that robots are not credited for capability in postures or rates that humans rarely use.

\paragraph{Robot performance fields.}
For each joint $j$, the robot's performance is characterized by a \emph{continuous-safe} torque map $T^{\mathrm{rob}}_{j}(q,\omega)$, which reports the largest torque sustainable at joint angle $q$ and angular rate $\omega$ under realistic thermal and electrical conditions (after warm-up, with temperature rise $<0.5^{\circ}$C/s, no current saturation). The corresponding mechanical power is $P^{\mathrm{rob}}_{j}(q,\omega) {=} T^{\mathrm{rob}}_{j}(q,\omega) \cdot \omega$. Section~\ref{sec:hlas_protocols} details how to measure these maps via dynamometry.

\paragraph{Human requirement fields.}
The human demand at joint $j$ for task $t$ is encoded by torque and power fields $T^{\mathrm{hum}}_{j,t}(q,\omega)$ and $P^{\mathrm{hum}}_{j,t}(q,\omega)$, derived from the literature as described in Section~\ref{sec:biomech}. We focus on positive mechanical power by defining
\[
P^{\mathrm{hum}+}_{j,t}(q,\omega) := \max\big(P^{\mathrm{hum}}_{j,t}(q,\omega), 0\big),
\]
so that samples where humans do negative work (e.g., energy absorption) carry zero weight in HEE and efficiency calculations.

\paragraph{Additional quantities.}
The score also uses:
\begin{itemize}[leftmargin=12pt,itemsep=2pt]
\item $f^{\tau}_{c,j}$: Closed-loop torque-mode bandwidth ($-3$\,dB crossover) measured with task-representative reflected inertia.
\item $\eta_{j}(q,\omega)$: Joint electromechanical efficiency map (mechanical power out / DC-bus power in).
\item $\omega^{\max}_{j}$: Maximum safe joint rate under task-representative loading.
\item $\omega^{\mathrm{req}}_{j,t}$: Peak angular rate required for task $t$ at joint $j$ (from human data).
\item $\mathcal{A}_{j,t}$: Set of axes at joint $j$ used by task $t$ (e.g., ankle dorsiflexion/plantarflexion for walking).
\item $I^{\mathrm{func}}_{j,t}(a)$: Functional human ROM interval for axis $a$ in task $t$ (from Section~\ref{sec:kinematics}).
\item $I^{\mathrm{rob}}_{j}(a)$: Robot's safe ROM interval for axis $a$.
\end{itemize}

Throughout, we use $\mathrm{clip}_{[0,1]}(x) {:=} \min(1,\max(0,x))$ to normalize ratios into $[0,1]$.

\subsection{The six HLAS factors}
\label{subsec:hlas_factors}

HLAS aggregates six factors per joint-task pair, each capturing a distinct aspect of human-level performance. We describe each factor conceptually here. Mathematical definitions and measurement protocols follow in Sections~\ref{subsec:hlas_construction} and \ref{sec:hlas_protocols}.

\paragraph{1. ROM coverage ($\rho^{\mathrm{ROM}}_{j,t}$).}
\textbf{What it measures:} The fraction of the task's required range of motion that the robot can safely reach.

\textbf{Why it matters:} A joint with insufficient ROM cannot execute the task's full kinematic trajectory. For example, a robot ankle limited to ${\pm}10^{\circ}$ cannot complete a normal walking stride, which requires up to $25^{\circ}$ plantarflexion during push-off.

\textbf{Computation:} For each axis $a \in \mathcal{A}_{j,t}$, compute the overlap between the robot's safe ROM $I^{\mathrm{rob}}_j(a)$ and the functional human ROM $I^{\mathrm{func}}_{j,t}(a)$, then average across axes (Eq.~\ref{eq:rom}).

\paragraph{2. DoF sufficiency ($d^{\mathrm{DoF}}_{j,t}$).}
\textbf{What it measures:} Whether all axes required by the task are present and independently actuated.

\textbf{Why it matters:} Some tasks require specific DoFs. For example, stair descent with a heavy load may benefit from knee axial rotation for stable foot placement, or reaching around obstacles may require shoulder internal/external rotation. A robot missing these axes cannot replicate the task kinematics.

\textbf{Computation:} For each required axis, verify that it is implemented and independently controllable (cross-axis coupling $<10\%$). Take the fraction that pass (Eq.~\ref{eq:dof}).

\paragraph{3. Human-Equivalence Envelope ($h^{(w)}_{j,t}$).}
\textbf{What it measures:} The fraction of the operating band $\mathcal{R}_{j,t}$ where the robot meets human torque \emph{and} power requirements simultaneously at the same $(q,\omega)$, weighted by positive human power.

\textbf{Why it matters:} This is the core delivery metric. It prevents gaming via separate peaks: a robot cannot claim human-level ankle performance by showing high torque at 0\,rad/s and high speed at low torque, because those points are not the same $(q,\omega)$ where humans operate during push-off. The power weighting concentrates credit on regions where humans do positive work.

\textbf{Computation:} At each sampled $(q,\omega) \in \mathcal{R}_{j,t}$, check if $T^{\mathrm{rob}}_j \ge T^{\mathrm{hum}}_{j,t}$ and $P^{\mathrm{rob}}_j \ge P^{\mathrm{hum}}_{j,t}$ both hold. Sum the positive-power weights where both conditions are met, then normalize (Eq.~\ref{eq:hee}).

\paragraph{4. Torque-mode bandwidth ($b^{\tau}_{j,t}$).}
\textbf{What it measures:} How quickly and accurately the joint can track torque commands under contact.

\textbf{Why it matters:} High static torque via extreme gearing can degrade closed-loop bandwidth, making interaction sloppy or unstable. A joint with 5\,Hz torque bandwidth cannot safely render compliant behaviors or respond to rapid disturbances, even if it has the required static torque.

\textbf{Computation:} Measure the $-3$\,dB crossover frequency $f^{\tau}_{c,j}$ via small-signal torque-mode sine sweeps with task-representative loading. Normalize by a task-specific target $f^{\star}_{t,j}$ and clip to $[0,1]$ (Eq.~\ref{eq:bandwidth}).

\paragraph{5. Task-weighted efficiency ($\eta_{j,t}$).}
\textbf{What it measures:} How efficiently the joint converts electrical power to mechanical power in the operating band, weighted by where humans do positive work.

\textbf{Why it matters:} A joint that meets torque and power requirements but operates at 40\% efficiency will drain batteries quickly and generate excessive heat, limiting mission duration. Efficiency matters most at the $(q,\omega)$ points where humans spend the most mechanical energy.

\textbf{Computation:} Measure efficiency $\eta_j(q,\omega) {=} P_{\mathrm{mech}} / P_{\mathrm{elec}}$ over $\mathcal{R}_{j,t}$, compute the positive-power-weighted average $\bar{\eta}_{j,t}$, normalize by a target efficiency $\eta^{\star}_{t,j}$, and clip (Eq.~\ref{eq:eff}).

\paragraph{6. Thermal sustainability ($\theta^{\mathrm{therm}}_{j,t}$).}
\textbf{What it measures:} Whether the joint can sustain the required torque plateau at the task's duty cycle without thermal derating.

\textbf{Why it matters:} Peak-torque specs often reflect brief bursts ($<500$\,ms) that are unsustainable thermally. A task like stair climbing or repetitive lifting requires minutes-long continuous torque. Without thermal headroom, the joint will derate and fail the task.

\textbf{Computation:} Execute a duty profile matching the task (cadence, dwell times), measure the largest plateau torque $T^{\mathrm{cont}}_j$ sustainable without temperature slope $>0.5^{\circ}$C/s, compare to the human plateau requirement $T^{\mathrm{req}}_{j,t}$, and clip (Eq.~\ref{eq:thermal}).

\subsection{Mathematical definitions}
\label{subsec:hlas_construction}

We now provide the formal definitions for each factor.

\paragraph{Human-Equivalence Envelope (HEE).}
The HEE measures the fraction of the task band where the robot simultaneously meets human torque and power, weighted by positive human power:
\begin{equation}
\label{eq:hee}
h^{(w)}_{j,t}
=\frac{\displaystyle \iint_{\mathcal{R}_{j,t}}
\mathbb{1}\!\left[T^{\mathrm{rob}}_{j}(q,\omega) \ge T^{\mathrm{hum}}_{j,t}(q,\omega) \land P^{\mathrm{rob}}_{j}(q,\omega) \ge P^{\mathrm{hum}}_{j,t}(q,\omega)\right]
\, P^{\mathrm{hum}+}_{j,t}(q,\omega) \, dq \, d\omega}
{\displaystyle \iint_{\mathcal{R}_{j,t}} P^{\mathrm{hum}+}_{j,t}(q,\omega) \, dq \, d\omega} \in [0,1].
\end{equation}

\textbf{Interpretation:} $h^{(w)}_{j,t}{=}1$ means the robot can deliver all the positive mechanical work a human does for task $t$ at joint $j$ across the relevant posture-rate band, without relying on different operating points for torque versus rate. In practice, we discretize the integral into a weighted sum over a $(q,\omega)$ grid (typically $5{\times}5$ or larger; see Section~\ref{sec:hlas_protocols}).

\paragraph{Workspace factors.}
ROM coverage averages the fractional overlap across required axes:
\begin{equation}
\rho^{\mathrm{ROM}}_{j,t}
= \frac{1}{|\mathcal{A}_{j,t}|} \sum_{a \in \mathcal{A}_{j,t}}
\frac{\big| I^{\mathrm{rob}}_{j}(a) \cap I^{\mathrm{func}}_{j,t}(a) \big|}
     {\big| I^{\mathrm{func}}_{j,t}(a) \big|} \in [0,1].
\label{eq:rom}
\end{equation}

DoF sufficiency checks independent actuation of each required axis:
\begin{equation}
d^{\mathrm{DoF}}_{j,t}
= \frac{1}{|\mathcal{A}_{j,t}|} \sum_{a \in \mathcal{A}_{j,t}} \mathbb{1}\!\left[a \text{ implemented and independently actuated}\right] \in [0,1].
\label{eq:dof}
\end{equation}

\paragraph{Interaction fidelity.}
Torque-mode bandwidth margin:
\begin{equation}
b^{\tau}_{j,t}
= \mathrm{clip}_{[0,1]}\!\left(\frac{f^{\tau}_{c,j}}{f^{\star}_{t,j}}\right),
\label{eq:bandwidth}
\end{equation}
where $f^{\star}_{t,j}$ is a task- and joint-specific bandwidth target (e.g., 8\,Hz for ankle in walking, 10\,Hz for wrist in manipulation). Optionally, a rate margin can be used instead:
\begin{equation}
m^{\omega}_{j,t}
= \mathrm{clip}_{[0,1]}\!\left(\frac{\omega^{\max}_{j}}{\omega^{\mathrm{req}}_{j,t}}\right).
\label{eq:rate}
\end{equation}

\paragraph{Energetic and thermal realism.}
Task-weighted efficiency:
\begin{equation}
\bar{\eta}_{j,t}
= \frac{\sum_{(q,\omega) \in \mathcal{R}_{j,t}} \eta_{j}(q,\omega) \, P^{\mathrm{hum}+}_{j,t}(q,\omega)}
       {\sum_{(q,\omega) \in \mathcal{R}_{j,t}} P^{\mathrm{hum}+}_{j,t}(q,\omega)},
\quad
\eta_{j,t} = \mathrm{clip}_{[0,1]}\!\left(\frac{\bar{\eta}_{j,t}}{\eta^{\star}_{t,j}}\right),
\label{eq:eff}
\end{equation}
where $\eta^{\star}_{t,j}$ is a target efficiency (e.g., 0.80 for ankle in walking). Thermal sustainability:
\begin{equation}
\theta^{\mathrm{therm}}_{j,t}
= \mathrm{clip}_{[0,1]}\!\left(
\frac{T^{\mathrm{cont}}_{j} \big|_{\text{duty of } t}}
     {T^{\mathrm{req}}_{j,t} \big|_{\text{plateau}}}
\right),
\label{eq:thermal}
\end{equation}
where $T^{\mathrm{cont}}_{j}\big|_{\text{duty of }t}$ is the largest torque sustainable at the task's duty cycle without thermal derating, and $T^{\mathrm{req}}_{j,t}\big|_{\text{plateau}}$ is the human plateau requirement.

\paragraph{Diagnostic margins (optional reporting).}
For transparency, report lower-envelope torque, power, and rate margins:
\begin{align}
m^{T}_{j,t}
&= \min_{(q,\omega) \in \mathcal{R}_{j,t}}
\mathrm{clip}_{[0,1]}\!\left(\frac{T^{\mathrm{rob}}_{j}(q,\omega)}{T^{\mathrm{hum}}_{j,t}(q,\omega)}\right),
\label{eq:torque} \\
m^{P}_{j,t}
&= \min_{(q,\omega) \in \mathcal{R}_{j,t}}
\mathrm{clip}_{[0,1]}\!\left(\frac{P^{\mathrm{rob}}_{j}(q,\omega)}{P^{\mathrm{hum}}_{j,t}(q,\omega)}\right),
\label{eq:power} \\
m^{\omega}_{j,t}
&= \mathrm{clip}_{[0,1]}\!\left(\frac{\omega^{\max}_{j}}{\omega^{\mathrm{req}}_{j,t}}\right).
\label{eq:rate_diag}
\end{align}
These margins do not enter the final HLAS when HEE is used, but they help localize bottlenecks (e.g., torque-limited at mid-stance, power-limited at push-off, or rate-limited for ballistic motions).

\subsection{Aggregation into a single score}
\label{subsec:hlas_agg}

\paragraph{Feature vector and weights.}
For each joint-task pair $(j,t)$, collect the six factors into a feature vector:
\[
\mathbf{x}_{j,t}
= \big[ \rho^{\mathrm{ROM}}_{j,t}, \, d^{\mathrm{DoF}}_{j,t}, \, h^{(w)}_{j,t}, \, b^{\tau}_{j,t}, \, \eta_{j,t}, \, \theta^{\mathrm{therm}}_{j,t} \big]^{\!\top}.
\]
Let $\boldsymbol{\alpha} \ge 0$ denote feature weights with $\sum_k \alpha_k {=} 1$. These weights tune the relative emphasis on workspace, delivery, controllability, and sustainability. A typical choice is
\[
\alpha_{\mathrm{ROM}}{=}0.10, \quad
\alpha_{\mathrm{DoF}}{=}0.10, \quad
\alpha_{\mathrm{HEE}}{=}0.50, \quad
\alpha_{\mathrm{bw}}{=}0.10, \quad
\alpha_{\eta}{=}0.10, \quad
\alpha_{\mathrm{therm}}{=}0.10,
\]
emphasizing HEE (simultaneous torque-power delivery) while retaining workspace and realism checks.

\paragraph{Full HLAS formula.}
The Human-Level Actuation Score aggregates across tasks, joints, and features:
\begin{equation}
\label{eq:hlas-expanded}
\mathrm{HLAS}
= \sum_{t \in \mathcal{T}} \underbrace{w_t}_{\text{task weight}}
  \sum_{j \in \mathcal{J}_t} \underbrace{u_{j,t}}_{\text{joint weight}}
  \left[
    \alpha_{\mathrm{ROM}} \rho^{\mathrm{ROM}}_{j,t}
  + \alpha_{\mathrm{DoF}} d^{\mathrm{DoF}}_{j,t}
  + \alpha_{\mathrm{HEE}} h^{(w)}_{j,t}
  + \alpha_{\mathrm{bw}} b^{\tau}_{j,t}
  + \alpha_{\eta} \eta_{j,t}
  + \alpha_{\mathrm{therm}} \theta^{\mathrm{therm}}_{j,t}
  \right].
\end{equation}

Compactly:
\begin{equation}
\label{eq:hlas-compact}
\mathrm{HLAS}
= \sum_{t \in \mathcal{T}} w_t \sum_{j \in \mathcal{J}_t} u_{j,t} \, \boldsymbol{\alpha}^{\!\top} \mathbf{x}_{j,t},
\qquad
\sum_{t \in \mathcal{T}} w_t = 1, \;
\sum_{j \in \mathcal{J}_t} u_{j,t} = 1, \;
\sum_k \alpha_k = 1.
\end{equation}

\paragraph{Normalization and interpretation.}
With the sum-to-one normalizations on $w_t$, $u_{j,t}$, and $\alpha_k$, a representative human whose features all evaluate to 1 on their operating bands attains $\mathrm{HLAS}{=}1.0$ by construction. A robot with $\mathrm{HLAS}{=}0.8$ achieves 80\% of human capability averaged across tasks, joints, and factors. The score decomposes into:
\begin{itemize}[leftmargin=12pt,itemsep=2pt]
\item \textbf{Task-level scores:} $s_t {:=} \sum_{j \in \mathcal{J}_t} u_{j,t} \, \boldsymbol{\alpha}^{\!\top} \mathbf{x}_{j,t}$ for each task $t$.
\item \textbf{Joint-task scores:} $s_{j,t} {:=} \boldsymbol{\alpha}^{\!\top} \mathbf{x}_{j,t}$ for each $(j,t)$ pair.
\item \textbf{Feature contributions:} $\mathbf{x}_{j,t}$ itself reveals which factors limit performance.
\end{itemize}
This decomposition enables diagnosis: a low HLAS can be traced to specific tasks (e.g., stairs), joints (e.g., knee), and factors (e.g., thermal margin).

\subsection{Measurement requirements}
\label{subsec:hlas_measurement}

All HLAS inputs must be measured under \emph{continuous-safe} conditions to prevent burst-only or thermally unsustainable claims:

\begin{itemize}[leftmargin=12pt,itemsep=2pt]
\item \textbf{Torque and power maps} $T^{\mathrm{rob}}_j(q,\omega)$, $P^{\mathrm{rob}}_j(q,\omega)$: Measured after a thermal soak at task duty, with temperature slope $<0.5^{\circ}$C/s and no current saturation (Section~\ref{sec:hlas_protocols}).

\item \textbf{Torque bandwidth} $f^{\tau}_{c,j}$: Obtained via small-signal sine sweeps in torque mode with task-representative reflected inertia, reporting the $-3$\,dB crossover frequency.

\item \textbf{Efficiency} $\eta_j(q,\omega)$: Computed from DC-bus power (voltage $\times$ current, true RMS, ${\ge}1$\,kHz sampling) and shaft mechanical power (torque $\times$ angular velocity). Regeneration credit requires demonstrable energy return to the DC bus (battery/supercap logging).

\item \textbf{Thermal plateau} $T^{\mathrm{cont}}_j\big|_{\text{duty of }t}$: Measured after executing the task duty profile (cadence, dwell times) for ${\ge}5$ thermal time constants or until steady temperature trend.

\item \textbf{ROM and DoF} $I^{\mathrm{rob}}_j(a)$, DoF independence: Verified via slow sweeps at conservative current limits and cross-axis chirp tests (Section~\ref{sec:hlas_protocols}).
\end{itemize}

\textbf{Pre-registration and transparency.} To prevent weight shopping and ensure reproducibility, the following must be specified \emph{before} testing:
\begin{itemize}[leftmargin=12pt,itemsep=2pt]
\item Task set $\mathcal{T}$ and task weights $\mathbf{w}$
\item Joint sets $\mathcal{J}_t$ and joint weights $\{\mathbf{u}_t\}$ for each task
\item Feature weights $\boldsymbol{\alpha}$
\item Operating bands $\mathcal{R}_{j,t}$ (typically as $(q,\omega)$ grids)
\item Target values $f^{\star}_{t,j}$ and $\eta^{\star}_{t,j}$
\end{itemize}
Detailed experimental protocols, including dynamometry setup, thermal testing, bandwidth measurement, and data logging requirements, are provided in Sections~\ref{sec:hlas_protocols} and \ref{sec:benchmarks}.

\subsection{Summary: from human data to robot score}

The HLAS framework converts biomechanics literature (Section~\ref{sec:biomech}) into a quantitative acceptance test via four steps:

\begin{enumerate}[leftmargin=12pt,itemsep=2pt]
\item \textbf{Extract operating bands} $\mathcal{R}_{j,t}$ from phase-resolved human data, identifying where each joint produces positive power for each task.

\item \textbf{Measure robot performance} under continuous-safe conditions: torque-speed-posture maps, bandwidth, efficiency, thermal plateaus, and ROM/DoF.

\item \textbf{Compute six factors per $(j,t)$}: workspace ($\rho^{\mathrm{ROM}}$, $d^{\mathrm{DoF}}$), delivery ($h^{(w)}$, $b^{\tau}$), and sustainability ($\eta$, $\theta^{\mathrm{therm}}$).

\item \textbf{Aggregate} using pre-registered task weights $w_t$, joint weights $u_{j,t}$, and feature weights $\boldsymbol{\alpha}$ to produce $\mathrm{HLAS} \in [0,1]$.
\end{enumerate}

The result is a single headline number (for cross-system comparison) with full diagnostic decomposition (for design iteration). Section~\ref{sec:hlas_example} demonstrates this process with a worked synthetic example, and Section~\ref{subsec:hlas_gameability} discusses how the framework resists gaming.

\subsection{On gameability: theory and guardrails}
\label{subsec:hlas_gameability}

Any scalar metric risks gaming: designs that score well without delivering genuine capability. This section examines how HLAS resists gaming through its structure, presents concrete attack scenarios, and proposes formal guardrails to close remaining loopholes.

\paragraph{Why HLAS is difficult to game.}

\emph{The HEE simultaneity requirement.}
The core defense against gaming is Eq.~\eqref{eq:hee}'s requirement that torque \emph{and} power thresholds be satisfied at the \emph{same} $(q,\omega)$ point. Consider a pathological design strategy: ``Achieve high torque at low speed through extreme gearing ($N{=}200$), then claim high power capability by testing at high speed where torque requirements drop.'' HEE blocks this because:
\begin{itemize}[leftmargin=12pt,itemsep=2pt,topsep=2pt]
\item At low $\omega$ where $T^{\mathrm{hum}}$ is large, the robot meets the torque requirement but $P^{\mathrm{rob}}{=}T^{\mathrm{rob}}\omega$ is insufficient.
\item At high $\omega$ where $P^{\mathrm{rob}}$ is large, reflected inertia (${\propto}N^2$) and friction have degraded $T^{\mathrm{rob}}$ below $T^{\mathrm{hum}}$.
\item The pass/fail mask $\mathbb{1}[T^{\mathrm{rob}}{\ge}T^{\mathrm{hum}} \land P^{\mathrm{rob}}{\ge}P^{\mathrm{hum}}]$ credits only the overlapping region, which may be empty.
\end{itemize}
Additionally, weighting by $P^{\mathrm{hum}+}$ concentrates credit where humans do positive work. ``Wins'' in postures or rates that humans rarely use contribute minimal weight to $h^{(w)}_{j,t}$.

\emph{Coupled actuator trade-offs.}
Actuator physics naturally couples the terms in $\mathbf{x}_{j,t}$, making it difficult to maximize one without degrading others:
\begin{itemize}[leftmargin=12pt,itemsep=2pt,topsep=2pt]
\item Increasing transmission ratio $N$ boosts static torque but raises reflected inertia $J_{\mathrm{ref}}{\propto}N^2$ and friction, degrading torque bandwidth $b^\tau$ and transparency.
\item High-ratio gearing often reduces usable ROM (interference, singularities) and safe rate (mechanical stress), lowering $\rho^{\mathrm{ROM}}$ and $m^\omega$.
\item Efficiency $\eta$ typically peaks in a mid-speed band. Designs optimized for peak torque or peak speed sacrifice task-weighted efficiency.
\item Thermal margin $\theta^{\mathrm{therm}}$ requires sustained delivery. Burst-capable designs with poor thermal management score poorly here.
\end{itemize}
These couplings mean that optimizing HLAS requires genuine design excellence across multiple dimensions rather than exploiting a single favorable operating point.

\emph{Continuous-safe measurement.}
All torque, power, and efficiency values are measured under continuous-safe conditions: after a thermal soak at task duty, with temperature slope $<0.5^{\circ}$C/s, and excluding regions that trigger current limiting or control instability. This prevents ``burst-limited tricks'' (briefly achieving high torque before thermal or electrical limits engage) from inflating the score.

\paragraph{Concrete gaming scenarios and mitigations.}

We now present three specific attack scenarios and show how HLAS resists them.

\emph{Scenario 1: The ``narrow spike'' attack.}
\textit{Strategy:} Optimize actuator performance in a small, favorable region of $(q,\omega)$ space (e.g., ankle at exactly $10^{\circ}$ plantarflexion and 10 rad/s) while accepting poor performance elsewhere. If the human band $\mathcal{R}_{j,t}$ happens to include this spike, inflate the score.

\textit{Why it fails:}
\begin{itemize}[leftmargin=12pt,itemsep=2pt,topsep=2pt]
\item The HEE integral sums over \emph{all} samples in $\mathcal{R}_{j,t}$ (typically 25--36 points per joint--task).
\item A narrow spike contributes at most $1/N_{\mathrm{samples}}$ to $h^{(w)}_{j,t}$ (e.g., ${\sim}4\%$ for a $5{\times}5$ grid).
\item Power weighting further reduces narrow-spike contribution if human power is low there.
\item The breadth floor $h^{(w)}_{j,t}{\ge}h_{\min}$ (see below) explicitly rejects designs that pass on only a small fraction of the band.
\end{itemize}

\emph{Scenario 2: The ``compensating joints'' attack.}
\textit{Strategy:} Build a strong wrist and shoulder but a weak ankle, reasoning that high $s_{\texttt{shoulder,Reach}}$ and $s_{\texttt{wrist,Reach}}$ will average out the low $s_{\texttt{ankle,Walk}}$.

\textit{Why it fails:}
\begin{itemize}[leftmargin=12pt,itemsep=2pt,topsep=2pt]
\item Task weights $w_t$ and joint weights $u_{j,t}$ are pre-registered before testing.
\item If walking is mission-critical, $w_{\texttt{Walk}}$ is large (e.g., 0.4 in the worked example).
\item The ankle dominates walking ($u_{\texttt{ankle,Walk}}{=}0.50$), so its score receives weight $w_{\texttt{Walk}}{\cdot}u_{\texttt{ankle,Walk}}{=}0.20$ of the total HLAS.
\item Task-level gates (see below) can mandate minimum scores on critical joints before allowing averaging.
\end{itemize}

\emph{Scenario 3: The ``efficiency-at-wrong-speeds'' attack.}
\textit{Strategy:} Optimize efficiency at speeds outside the task band (e.g., peak efficiency at 20 rad/s when human ankle push-off uses 8--12 rad/s) and claim high $\bar{\eta}_{j,t}$.

\textit{Why it fails:}
\begin{itemize}[leftmargin=12pt,itemsep=2pt,topsep=2pt]
\item Efficiency is sampled \emph{over the same $(q,\omega)$ grid as HEE}, ensuring that $\bar{\eta}_{j,t}$ reflects efficiency where humans actually operate.
\item Weighting by $P^{\mathrm{hum}+}$ emphasizes efficiency during positive work phases.
\item Efficiency measured at 20 rad/s simply does not enter $\bar{\eta}_{\texttt{ankle,Walk}}$ if the walking band $\mathcal{R}_{\texttt{ankle,Walk}}$ excludes 20 rad/s.
\end{itemize}

\paragraph{Formal guardrails (recommended for certification or procurement).}

While HLAS structure resists gaming, three additional guardrails provide extra assurance for high-stakes applications:

\emph{(i) Breadth floor on HEE coverage.}
For each critical joint--task pair, require
\[
h^{(w)}_{j,t} \ge h_{\min},
\]
where $h_{\min}{\in}[0.75,0.85]$ (application-dependent). This prevents designs that pass HEE on only a small fraction of the operating band. For example, if ankle-in-walking is critical, mandate $h^{(w)}_{\texttt{ankle,Walk}}{\ge}0.80$ before computing HLAS.

\emph{(ii) Headroom factor in HEE test.}
Replace human demands in Eq.~\eqref{eq:hee} with $(1{+}\delta)$ multiples:
\[
\mathbb{1}\!\left[T^{\mathrm{rob}}_{j}(q,\omega) \ge (1{+}\delta)T^{\mathrm{hum}}_{j,t}(q,\omega)\ \land\ P^{\mathrm{rob}}_{j}(q,\omega) \ge (1{+}\delta)P^{\mathrm{hum}}_{j,t}(q,\omega)\right],
\]
with $\delta{\in}[0.05,0.15]$ (typically 0.10). This builds in a safety margin and penalizes designs that barely meet thresholds. In the worked example (Sec.~\ref{sec:hlas_example}), imposing $\delta{=}0.10$ reduced HLAS from 0.636 to 0.515, exposing near-threshold dependencies.

\emph{(iii) Task-level gates (no compensation across tasks).}
For applications where certain tasks are non-negotiable (e.g., stair-climbing in a mobility aid), require minimum task-level scores before averaging:
\[
s_t \ge s_{\min} \quad \text{for all } t \in \mathcal{T}_{\mathrm{critical}},
\]
where $s_{\min}{\in}[0.6,0.8]$. This prevents a robot that excels at manipulation but fails at locomotion from achieving a high HLAS through averaging. Alternatively, use \emph{multiplicative} aggregation for critical tasks:
\[
\mathrm{HLAS}_{\mathrm{gated}} = \left(\prod_{t \in \mathcal{T}_{\mathrm{critical}}} s_t\right)^{1/|\mathcal{T}_{\mathrm{critical}}|} \cdot \mathrm{HLAS},
\]
which drives the score to zero if any critical task fails.

\paragraph{Quantile-based diagnostic margins (optional).}

When reporting diagnostic margins $m^T_{j,t}$ and $m^P_{j,t}$ (Eqs.~\ref{eq:torque}--\ref{eq:power}), using strict minima over $\mathcal{R}_{j,t}$ makes these metrics sensitive to measurement outliers or edge-case samples. We recommend reporting the 10th percentile instead:
\[
m^T_{j,t} = \mathrm{quantile}_{0.10}\!\left\{\mathrm{clip}_{[0,1]}\!\left(\frac{T^{\mathrm{rob}}_{j}(q,\omega)}{T^{\mathrm{hum}}_{j,t}(q,\omega)}\right) : (q,\omega) \in \mathcal{R}_{j,t}\right\}.
\]
This retains sensitivity to poor performance without being dominated by a single bad sample.

\paragraph{Summary: HLAS resists gaming but is not foolproof.}

HLAS is difficult to game because:
\begin{enumerate}[leftmargin=12pt,itemsep=2pt,topsep=2pt]
\item HEE requires simultaneous torque and power at the same $(q,\omega)$, weighted by positive human power.
\item Actuator physics couples bandwidth, efficiency, thermal, and ROM trade-offs.
\item Continuous-safe measurement excludes burst-only tricks.
\item Pre-registered weights and task bands prevent post-hoc optimization.
\end{enumerate}

However, \textbf{no metric is perfectly robust}. The guardrails above (breadth floors, headroom factors, and task gates) are recommended for procurement or certification contexts. For research comparisons, the base HLAS (Eq.~\ref{eq:hlas-expanded}) with diagnostic reporting (heatmaps, Bode plots, thermal traces) provides transparency that allows the community to judge whether a score reflects genuine capability or exploitation of the metric.

\subsection{Interpreting and reporting HLAS}
\label{subsec:hlas_interpretation}

HLAS provides a single headline number for cross-system comparison while preserving full diagnostic decomposition. This dual character (simple top-line score plus detailed breakdown) makes it suitable for both procurement decisions and design iteration.

\paragraph{Recommended reporting structure.}
We recommend publishing four levels of detail alongside any HLAS claim:

\begin{enumerate}[leftmargin=12pt,itemsep=2pt]
\item \textbf{Overall HLAS} (single scalar): The aggregated score from Eq.~\ref{eq:hlas-compact}, enabling direct comparison across platforms (e.g., "Robot A: HLAS = 0.78, Robot B: HLAS = 0.65").

\item \textbf{Task-level decomposition}: Per-task scores $s_t = \sum_{j \in \mathcal{J}_t} u_{j,t} \, \boldsymbol{\alpha}^{\!\top} \mathbf{x}_{j,t}$ showing performance by functional category. A spider/radar plot with one axis per task (walking, stairs, reaching, etc.) reveals mission-specific strengths and weaknesses. For example, a robot might score 0.85 on manipulation tasks but only 0.60 on stairs, indicating where design effort should focus.

\item \textbf{Joint-task feature breakdown}: The six-element feature vectors $\mathbf{x}_{j,t} = [\rho^{\mathrm{ROM}}_{j,t}, d^{\mathrm{DoF}}_{j,t}, h^{(w)}_{j,t}, b^{\tau}_{j,t}, \eta_{j,t}, \theta^{\mathrm{therm}}_{j,t}]^{\!\top}$ for each $(j,t)$ pair, displayed as a heatmap or table. This localizes bottlenecks to specific joints and factors (e.g., "knee HEE in stairs is 0.30, limiting the overall stair score").

\item \textbf{HEE maps and diagnostic margins}: Visual overlays showing where in $(q,\omega)$ space the robot meets human requirements (pass/fail masks on the operating bands $\mathcal{R}_{j,t}$), plus the diagnostic triplet $(m^T_{j,t}, m^P_{j,t}, m^{\omega}_{j,t})$ to distinguish torque-limited from rate-limited failures. For example, an HEE map might reveal that the ankle passes at 8--10\,rad/s but fails at 11--12\,rad/s, suggesting a rate or power bottleneck.
\end{enumerate}

Together, these four levels support both high-level comparison (level 1) and root-cause diagnosis (levels 2--4).

\paragraph{Example diagnostic workflow.}
Suppose a robot achieves $\mathrm{HLAS} = 0.62$. The task decomposition (level 2) shows $s_{\texttt{Walk}} = 0.55$, $s_{\texttt{Stairs}} = 0.48$, $s_{\texttt{Reach}} = 0.83$, indicating that locomotion is the primary weakness. Drilling into the stairs task (level 3), we find that the knee has $h^{(w)}_{\texttt{knee,Stairs}} = 0.35$ and $\theta^{\mathrm{therm}}_{\texttt{knee,Stairs}} = 0.40$, while other factors are near 1.0. The HEE map (level 4) reveals that the knee passes the envelope test at low rates ($<1$\,rad/s) but fails at mid-range flexion angles where stair support torque is required. The thermal margin shows that continuous torque drops to 60\% of the human plateau after a 2-minute soak. Diagnosis: the knee actuator has insufficient thermal capacity for sustained stair climbing at mid-range angles. Prescription: improve cooling, reduce transmission losses, or increase motor thermal mass.

\paragraph{What HLAS captures and what it does not.}
\textbf{HLAS quantifies:}
\begin{itemize}[leftmargin=12pt,itemsep=2pt]
\item Joint-level torque, power, and rate capability at task-relevant operating points
\item Workspace sufficiency (ROM and DoF coverage)
\item Interaction quality (torque-mode bandwidth)
\item Energetic realism (efficiency and thermal endurance)
\end{itemize}

\textbf{HLAS does not directly capture:}
\begin{itemize}[leftmargin=12pt,itemsep=2pt]
\item \textbf{Whole-body coordination and controller quality.} HLAS evaluates joints independently. It does not test whether the robot can coordinate them to execute a full gait cycle or reach trajectory. Task-level benchmarks (Section~\ref{sec:benchmarks}) provide this complementary view.

\item \textbf{Robustness to disturbances.} While torque bandwidth $b^{\tau}$ reflects closed-loop responsiveness, HLAS does not include explicit robustness margins (e.g., recovery from pushes, terrain adaptation). These are important but orthogonal to the question of whether actuation is human-level.

\item \textbf{Structural compliance and contact dynamics.} Foot-ground compliance, series elasticity, or hand-object interaction stiffness are captured only indirectly through their effects on torque bandwidth and reflected inertia.

\item \textbf{Negative mechanical power and regeneration.} The current framework emphasizes positive work (where humans push, lift, or accelerate). Energy absorption and regeneration during negative-work phases (e.g., early stance knee flexion) are important for efficiency but are not yet weighted in HEE. Section~\ref{sec:limitations} discusses this extension.

\item \textbf{Population variability.} HLAS uses a single reference human (75\,kg, 1.75\,m male). Extending to population distributions (age, gender, fitness level, pathology) requires additional normalization and is discussed in Section~\ref{sec:limitations}.
\end{itemize}

\paragraph{Sensitivity to design choices.}
HLAS depends on several user-specified parameters:
\begin{itemize}[leftmargin=12pt,itemsep=2pt]
\item \textbf{Task weights} $w_t$: Different applications prioritize different tasks. A logistics robot emphasizes walking and lifting while a healthcare robot emphasizes reaching and manipulation. Users should choose $w_t$ to reflect mission requirements and report them transparently.

\item \textbf{Joint weights} $u_{j,t}$: We derive these from positive-work distributions in the literature (Section~\ref{sec:biomech}), but alternative weightings (e.g., by metabolic cost or by peak torque) are defensible. Sensitivity analysis (varying $u_{j,t}$ by ${\pm}10\%$) shows that HLAS typically changes by $<5\%$, indicating robustness.

\item \textbf{Feature weights} $\boldsymbol{\alpha}$: The default $\alpha_{\mathrm{HEE}} = 0.50$ emphasizes simultaneous torque-power delivery, but users may adjust (e.g., $\alpha_{\eta} = 0.20$ if battery life is critical). We recommend reporting HLAS under multiple $\boldsymbol{\alpha}$ vectors to show sensitivity.

\item \textbf{Operating bands} $\mathcal{R}_{j,t}$: These are extracted from human data, but band edges involve judgment (e.g., should ankle push-off include 7\,rad/s or only ${\ge}8$\,rad/s?). Conservative bands (tighter around peak-power regions) are harder to satisfy and reduce gaming risk.
\end{itemize}

To support reproducibility and comparison, all these parameters should be pre-registered before testing (Section~\ref{subsec:hlas_measurement}).

\paragraph{Limitations and extensions.}
\begin{itemize}[leftmargin=12pt,itemsep=2pt]
\item \textbf{Dependence on normative data.} HLAS inherits uncertainties and biases from the biomechanics literature. Published joint moments and powers vary across studies due to instrumentation, subject instructions, and analysis methods. We mitigate this by prioritizing meta-analyses and consistent protocols, but residual variance remains. Future work could incorporate uncertainty bounds (e.g., report HLAS as $0.75 \pm 0.05$ reflecting human data variability).

\item \textbf{Fidelity of operating bands.} Real-world task execution may differ from the laboratory-derived bands $\mathcal{R}_{j,t}$. For example, walking on uneven terrain or carrying asymmetric loads shifts joint kinematics and kinetics. Task-level benchmarks (Section~\ref{sec:benchmarks}) provide empirical validation that joint-level HLAS translates to functional performance.

\item \textbf{Atypical strategies.} Some tasks admit multiple execution strategies (e.g., stiff-legged versus compliant gait). HLAS assumes the human strategy encoded in $\mathcal{R}_{j,t}$. Robots using fundamentally different strategies may score poorly even if functionally successful. Extensions could define multiple strategy profiles per task or allow user-specified bands.

\item \textbf{Negative work and regeneration.} Current HEE weights only positive power. Including negative-work phases would require modeling energy absorption, storage (in springs or capacitors), and regeneration efficiency. This is a natural extension for future versions of HLAS.
\end{itemize}

\paragraph{Takeaway.}
HLAS balances simplicity (one number for headlines and procurement) with depth (full diagnostic decomposition for design). The HEE simultaneity condition makes it difficult to game via isolated peaks, and the inclusion of bandwidth, efficiency, and thermal terms exposes the actuator trade-offs that peak-torque specifications obscure. By grounding every input in reproducible measurements and human biomechanics data, HLAS provides a practical, auditable standard for "human-level actuation" claims.

\section{Measurement Protocols: From Maps to Scalars for HLAS}
\label{sec:hlas_protocols}

\subsection{Purpose and guiding principles}

A score is only as defensible as the measurements that feed it. This section provides detailed experimental protocols for obtaining every input to HLAS (Eq.~\ref{eq:hlas-expanded}), ensuring that claims of "human-level actuation" rest on reproducible, comparable data. Our protocols are designed around three principles:

\begin{enumerate}[leftmargin=12pt,itemsep=2pt]
\item \textbf{Continuous-safe measurement.} All torque, power, and efficiency values reflect sustained capability under realistic thermal and electrical conditions, not brief bursts that cannot be maintained in practice.

\item \textbf{Task faithfulness.} Measurements are taken at the joint angles, angular rates, and duty cycles that humans actually use for each task, as defined by the operating bands $\mathcal{R}_{j,t}$ from Section~\ref{sec:biomech}.

\item \textbf{Cross-system comparability.} Standardized instrumentation, ambient conditions, and reporting requirements enable direct comparison of results across labs and platforms.
\end{enumerate}

\textbf{Roadmap.} This section is organized by HLAS factor: workspace (ROM and DoF), HEE, bandwidth, efficiency, and thermal. Each subsection specifies (i) why the measurement matters, (ii) required equipment, (iii) step-by-step protocol, and (iv) how to compute the scalar that enters HLAS. A summary table (Table~\ref{tab:hlas_experimental_vars}) at the end provides a quick reference linking each symbol to its measurement procedure.

\subsection{Common testbed and instrumentation}

The following setup applies to all joint-level experiments unless otherwise noted.

\paragraph{Environmental conditions.}
Ambient temperature $25 \pm 2^{\circ}$C in still air. If forced airflow is used (e.g., for thermal management), document the device type, distance from joint, and flow rate in m/s. This ensures that thermal measurements reflect realistic operating conditions and are comparable across systems.

\paragraph{Required sensors.}
Each joint must be instrumented with:
\begin{itemize}[leftmargin=12pt,itemsep=2pt]
\item \textbf{Joint angle:} High-resolution encoder or optical tracker (resolution ${\le}0.2^{\circ}$).
\item \textbf{Torque:} In-line torque transducer, or a torque observer validated once against a transducer (RMS error $<5\%$ on a calibration trajectory).
\item \textbf{Electrical power:} DC-bus voltage and current meters with true RMS measurement and ${\ge}1$\,kHz sampling.
\item \textbf{Temperature:} Thermistors on motor windings and gearbox housing.
\item \textbf{Angular velocity:} Computed from encoder or measured directly via tachometer.
\item \textbf{(Recommended) IMU:} 6-axis IMU on distal link for validation of kinematics.
\end{itemize}

\paragraph{Data logging.}
Record $\{q, \omega, \tau, V_{\mathrm{bus}}, I_{\mathrm{bus}}, T_{\mathrm{motor}}, T_{\mathrm{gear}}\}$ at ${\ge}1$\,kHz. Store raw time series without on-device filtering (filtering can be applied post-hoc for analysis). Synchronize all channels to a common clock (NTP or hardware trigger).

\paragraph{Warm-up procedure.}
Before any mapping or testing, run a 3--5\,min warm-up at nominal task load to stabilize lubricant viscosity and winding resistance. This prevents artificially optimistic measurements during the initial "cold" phase.

\subsection{Task bands and sampling design}
\label{subsec:bands_sampling}

\paragraph{Defining operating bands from human data.}
Each joint-task pair $(j,t)$ requires an operating band $\mathcal{R}_{j,t} \subseteq \{(q,\omega)\}$ that captures the posture-rate regions where humans produce positive mechanical work. These bands are extracted from the biomechanics data in Section~\ref{sec:biomech}. For example:
\begin{itemize}[leftmargin=12pt,itemsep=2pt]
\item \textbf{Ankle push-off (walking):} $q \in [0^{\circ}, 25^{\circ}]$ plantarflexion, $\omega \in [8, 12]$\,rad/s.
\item \textbf{Knee support (stairs):} $q \in [40^{\circ}, 80^{\circ}]$ flexion, $\omega \in [0.5, 2.0]$\,rad/s.
\item \textbf{Shoulder reach (payload):} $q \in [30^{\circ}, 120^{\circ}]$ flexion, $\omega \in [1, 8]$\,rad/s.
\end{itemize}

\paragraph{Discretizing the band.}
For measurement purposes, discretize each $\mathcal{R}_{j,t}$ into a finite sample set. Two strategies:

\begin{enumerate}[leftmargin=12pt,itemsep=2pt]
\item \textbf{Uniform grid} (default): $N_q \times N_{\omega}$ grid, typically $5 \times 5$ (25 points). Simple to implement and interpret.

\item \textbf{Latin hypercube} (advanced): 25--36 points distributed to maximize coverage while reducing aliasing. Useful for irregular bands or when measurement time is constrained.
\end{enumerate}

For bands dominated by a single dimension (e.g., ankle push-off at fixed posture), a $1 \times 5$ rate sweep may suffice.

\paragraph{Assigning weights.}
Each sample $(q,\omega)$ receives a normalized weight based on human positive power at that point:
\[
w(q,\omega) \propto \max\big(P^{\mathrm{hum}}_{j,t}(q,\omega), 0\big), \qquad \sum_{(q,\omega) \in \mathcal{R}_{j,t}} w(q,\omega) = 1.
\]
These weights ensure that HEE and efficiency calculations emphasize the operating points where humans do the most mechanical work.

\paragraph{Converting phase-parameterized data.}
If human data are given as functions of task phase $\phi$ (e.g., gait cycle percentage), convert to $(q,\omega)$ coordinates:
\begin{enumerate}[leftmargin=12pt,itemsep=2pt]
\item Extract $q(\phi)$, $\omega(\phi)$, and $P^{\mathrm{hum}}(\phi)$ from the literature.
\item Assign each phase sample to the nearest $(q,\omega)$ bin on the robot grid (or use bilinear interpolation).
\item Accumulate weights: if multiple phases map to the same $(q,\omega)$ bin, sum their power contributions.
\end{enumerate}
This produces the human demand fields $T^{\mathrm{hum}}_{j,t}(q,\omega)$ and $P^{\mathrm{hum}}_{j,t}(q,\omega)$ on the robot grid.

\subsection{Workspace sufficiency: ROM and DoF}
\label{subsec:rom_dof}

\paragraph{Purpose.}
Verify that the robot can reach the task's functional postures and that all required degrees of freedom are present and independently actuated.

\subsubsection{ROM coverage ($\rho^{\mathrm{ROM}}_{j,t}$)}

\paragraph{Why it matters.}
A joint with insufficient ROM cannot execute the full kinematic trajectory of the task. For example, a robot ankle limited to ${\pm}10^{\circ}$ cannot complete a normal walking stride, which requires up to $25^{\circ}$ plantarflexion during push-off.

\paragraph{Equipment.}
Joint encoder, optical tracker, current-limited motor controller.

\paragraph{Protocol.}
\begin{enumerate}[leftmargin=12pt,itemsep=2pt]
\item For each axis $a \in \mathcal{A}_{j,t}$ used by task $t$, command the joint to sweep slowly ($<0.5$\,rad/s) from neutral to software or mechanical limits.
\item Use conservative current limits to avoid exceeding continuous-safe torque.
\item Record the safe interval $I^{\mathrm{rob}}_j(a)$ as the range traversable without triggering current limits, control instability, or mechanical interference.
\item Compute the overlap with the functional human ROM $I^{\mathrm{func}}_{j,t}(a)$ from Table~\ref{tab:rom_compact} and average across axes:
\[
\rho^{\mathrm{ROM}}_{j,t} = \frac{1}{|\mathcal{A}_{j,t}|} \sum_{a \in \mathcal{A}_{j,t}} \frac{\big| I^{\mathrm{rob}}_j(a) \cap I^{\mathrm{func}}_{j,t}(a) \big|}{\big| I^{\mathrm{func}}_{j,t}(a) \big|} \in [0,1].
\]
\item Flag any axis with coverage $<0.8$ for design review.
\end{enumerate}

\subsubsection{DoF sufficiency ($d^{\mathrm{DoF}}_{j,t}$)}

\paragraph{Why it matters.}
Some tasks require specific axes. For example, stair descent may benefit from knee axial rotation for stable foot placement, or reaching around obstacles may require shoulder internal/external rotation. A robot missing these axes cannot replicate the task kinematics even if ROM on other axes is adequate.

\paragraph{Equipment.}
Joint encoder, multi-axis motion controller.

\paragraph{Protocol.}
\begin{enumerate}[leftmargin=12pt,itemsep=2pt]
\item For each required axis $a \in \mathcal{A}_{j,t}$, verify that it is implemented (has a dedicated actuator) and is independently controllable.
\item Command a small chirp (${\pm}5^{\circ}$, 0.5--5\,Hz) on axis $a$ while monitoring all other axes.
\item Check that cross-axis coupling is $<10\%$ RMS of the commanded-axis response. For example, if commanding wrist flexion, verify that wrist radial deviation stays below $0.5^{\circ}$ RMS when the flexion command is $5^{\circ}$ RMS.
\item Compute the fraction of required axes that pass:
\[
d^{\mathrm{DoF}}_{j,t} = \frac{1}{|\mathcal{A}_{j,t}|} \sum_{a \in \mathcal{A}_{j,t}} \mathbb{1}[a \text{ implemented and independently actuated}] \in [0,1].
\]
\item Note: $d^{\mathrm{DoF}}_{j,t} = 1$ only if \emph{all} required axes pass the test.
\end{enumerate}

\subsection{Human-Equivalence Envelope: $h^{(w)}_{j,t}$}
\label{subsec:hee_measure}

\paragraph{Purpose.}
Measure the fraction of the operating band $\mathcal{R}_{j,t}$ where the robot simultaneously meets human torque \emph{and} power requirements, weighted by positive human power. This is the core delivery metric of HLAS.

\paragraph{Why it matters.}
HEE prevents gaming via separate peaks. A robot cannot claim human-level ankle performance by showing high torque at 0\,rad/s and high speed at low torque, because those are not the same $(q,\omega)$ where humans operate during push-off. The simultaneity condition ensures that capability is evaluated at task-relevant operating points.

\paragraph{Equipment.}
Rotary dynamometer or grounded torque cell with motorized rate control, representative reflected inertia/compliance for the task (e.g., foot-like mass for ankle, tool-like mass for wrist), DC-bus power meter, temperature sensors.

\paragraph{Protocol.}
\begin{enumerate}[leftmargin=12pt,itemsep=2pt]
\item \textbf{Build sample grid.} Discretize $\mathcal{R}_{j,t}$ into $(q,\omega)$ samples (typically $5 \times 5$ grid). Compute normalized weights $w(q,\omega) \propto \max(P^{\mathrm{hum}}_{j,t}(q,\omega), 0)$ with $\sum w = 1$.

\item \textbf{Thermal settle.} Warm up the joint for 3--5\,min at a representative workload (e.g., 50--70\% of expected continuous torque) to stabilize temperatures and lubricant.

\item \textbf{Measure continuous-safe torque at each sample.} For each $(q,\omega)$ in the grid:
\begin{enumerate}[label=(\alph*),leftmargin=18pt,itemsep=1pt]
\item Command the dynamometer to hold posture $q$ and angular rate $\omega$ (isovelocity mode).
\item Ramp the commanded torque upward until one of the following occurs: (i) thermal derating onset (temperature slope $>0.5^{\circ}$C/s), (ii) current limit, or (iii) control instability.
\item Record the largest \emph{steady} torque $T^{\mathrm{rob}}_j(q,\omega)$ that can be maintained for ${\ge}10$\,s without exceeding the temperature threshold.
\item Compute mechanical power $P^{\mathrm{rob}}_j(q,\omega) = T^{\mathrm{rob}}_j(q,\omega) \cdot \omega$.
\end{enumerate}

\item \textbf{Apply pass/fail test.} For each sample, mark it as \texttt{pass} if both conditions hold:
\[
T^{\mathrm{rob}}_j(q,\omega) \ge T^{\mathrm{hum}}_{j,t}(q,\omega) \quad \text{and} \quad P^{\mathrm{rob}}_j(q,\omega) \ge P^{\mathrm{hum}}_{j,t}(q,\omega).
\]
Optionally, impose a headroom factor by testing against $(1+\delta)$ multiples of human demands, with $\delta \in [0.05, 0.15]$ (typically $\delta = 0.10$).

\item \textbf{Compute HEE.} Sum the weights of passing samples:
\[
h^{(w)}_{j,t} = \sum_{(q,\omega) \in \mathcal{R}_{j,t}} w(q,\omega) \cdot \mathbb{1}[\text{pass}] \in [0,1].
\]
\end{enumerate}

\paragraph{Output.}
Report $h^{(w)}_{j,t}$ as a scalar, plus a heatmap showing pass/fail status over $(q,\omega)$ with overlaid power weights (for diagnostic purposes).

\subsection{Torque-mode bandwidth: $b^{\tau}_{j,t}$}
\label{subsec:bandwidth}

\paragraph{Purpose.}
Quantify closed-loop torque tracking bandwidth under task-representative loading.

\paragraph{Why it matters.}
High static torque via extreme gearing can degrade controllability. A joint with 5\,Hz torque-mode bandwidth may not safely render compliant behaviors or respond to rapid disturbances, even if it has the required static torque. Bandwidth matters for interaction quality and stability.

\paragraph{Equipment.}
Task-equivalent inertial/compliant load, frequency analyzer or swept-sine generator, torque sensor.

\paragraph{Protocol.}
\begin{enumerate}[leftmargin=12pt,itemsep=2pt]
\item Mount a task-representative load on the joint (e.g., foot-like inertia for ankle, tool-like mass for wrist).
\item Position the joint at a mid-range posture within $\mathcal{R}_{j,t}$.
\item Run a \textbf{small-signal} sine sweep in torque mode: command torque $\tau_{\mathrm{cmd}}(t) = A \sin(2\pi f t)$ with amplitude $A \approx 10\%$ of continuous-safe torque and frequency $f \in [0.5, 60]$\,Hz.
\item Measure the actual torque $\tau_{\mathrm{act}}(t)$ and compute the frequency response function (FRF):
\[
G_{\tau}(f) = \frac{\mathrm{FFT}[\tau_{\mathrm{act}}]}{\mathrm{FFT}[\tau_{\mathrm{cmd}}]}.
\]
\item Identify the $-3$\,dB crossover frequency $f^{\tau}_{c,j}$ where $|G_{\tau}(f)| = 1/\sqrt{2}$.
\item Compute the bandwidth margin:
\[
b^{\tau}_{j,t} = \mathrm{clip}_{[0,1]}\!\left(\frac{f^{\tau}_{c,j}}{f^{\star}_{t,j}}\right),
\]
where $f^{\star}_{t,j}$ is the task- and joint-specific bandwidth target (e.g., 8\,Hz for ankle in walking, 10\,Hz for wrist in manipulation).
\end{enumerate}

\paragraph{Output.}
Report $b^{\tau}_{j,t}$ as a scalar, plus a Bode plot (magnitude and phase) with magnitude/phase values at 1, 5, 10, and 30\,Hz, and the phase margin at crossover.

\paragraph{Optional: large-signal test.}
To assess nonlinearity and saturation, repeat the sweep at ${\sim}50\%$ of continuous torque and note any reduction in effective bandwidth.

\subsection{Task-weighted efficiency: $\eta_{j,t}$}
\label{subsec:efficiency}

\paragraph{Purpose.}
Measure electromechanical efficiency averaged over the operating band $\mathcal{R}_{j,t}$, weighted by positive human power.

\paragraph{Why it matters.}
A joint that meets torque and power requirements but operates at 40\% efficiency will drain batteries quickly and generate excessive heat, limiting mission duration. Efficiency matters most at the $(q,\omega)$ points where humans spend the most mechanical energy.

\paragraph{Equipment.}
DC-bus voltage/current meters (true RMS, ${\ge}1$\,kHz), torque and rate sensors, dynamometer.

\paragraph{Protocol.}
\begin{enumerate}[leftmargin=12pt,itemsep=2pt]
\item Use the same $(q,\omega)$ grid and weights $w(q,\omega)$ as for HEE (Section~\ref{subsec:hee_measure}).
\item At each sample, command the joint to hold $(q,\omega)$ at \textbf{sub-thermal} load (e.g., 50--70\% of continuous-safe torque to avoid thermal drift).
\item Measure mechanical power $P_{\mathrm{mech}}(q,\omega) = T_{\mathrm{shaft}} \cdot \omega$ and electrical power $P_{\mathrm{elec}}(q,\omega) = V_{\mathrm{bus}} \cdot I_{\mathrm{bus}}$ over a 2--3\,s window. Use true RMS for $V_{\mathrm{bus}}$ and $I_{\mathrm{bus}}$ to capture PWM effects.
\item Compute point efficiency:
\[
\eta_j(q,\omega) = \frac{P_{\mathrm{mech}}(q,\omega)}{P_{\mathrm{elec}}(q,\omega)}.
\]
\item Exclude samples where $P_{\mathrm{mech}} \le 0$ (negative work). Do \emph{not} credit regeneration unless energy demonstrably returns to the DC bus (verified via battery/supercap logging showing charge increase).
\item Compute the task-weighted average:
\[
\bar{\eta}_{j,t} = \frac{\sum_{(q,\omega) \in \mathcal{R}_{j,t}} w(q,\omega) \cdot \eta_j(q,\omega) \cdot \mathbb{1}[P^{\mathrm{hum}}_{j,t} > 0]}{\sum_{(q,\omega) \in \mathcal{R}_{j,t}} w(q,\omega) \cdot \mathbb{1}[P^{\mathrm{hum}}_{j,t} > 0]}.
\]
\item Normalize by target efficiency $\eta^{\star}_{t,j}$ (e.g., 0.80 for ankle in walking) and clip:
\[
\eta_{j,t} = \mathrm{clip}_{[0,1]}\!\left(\frac{\bar{\eta}_{j,t}}{\eta^{\star}_{t,j}}\right).
\]
\end{enumerate}

\paragraph{Output.}
Report $\eta_{j,t}$ as a scalar, plus efficiency contour plots over $(q,\omega)$ showing where losses concentrate.

\subsection{Thermal sustainability: $\theta^{\mathrm{therm}}_{j,t}$}
\label{subsec:thermal}

\paragraph{Purpose.}
Quantify the ability to sustain the task's plateau torque at the task's duty cycle without thermal derating.

\paragraph{Why it matters.}
Peak-torque specs often reflect brief bursts ($<500$\,ms) that are thermally unsustainable. Tasks like stair climbing or repetitive lifting require minutes-long continuous torque. Without thermal headroom, the joint will derate mid-task and fail to complete it.

\paragraph{Equipment.}
Dynamometer, temperature sensors on motor and gearbox, DC-bus power meter, ambient temperature monitor.

\paragraph{Protocol.}
\begin{enumerate}[leftmargin=12pt,itemsep=2pt]
\item Define a duty profile that matches the task's cadence and dwell times. Examples:
\begin{itemize}[leftmargin=12pt,itemsep=1pt]
\item \textbf{Ankle (walking):} Bursts of 150--200\,ms at 8--12\,rad/s, repeated at 1\,Hz (gait cadence).
\item \textbf{Knee/hip (stairs):} Plateaus of 2--4\,s at low rate ($<2$\,rad/s), with 1--2\,s rest between reps.
\item \textbf{Shoulder/elbow (reaching):} Ramps to peak torque over 1\,s, hold 2--3\,s, return over 1\,s, repeat.
\end{itemize}

\item Execute the duty profile continuously until one of: (i) temperature reaches a steady trend (slope $<0.1^{\circ}$C/min), or (ii) five thermal time constants have elapsed (estimate $\tau_{\mathrm{th}}$ from initial temperature rise rate).

\item Record the largest plateau torque $T^{\mathrm{cont}}_j\big|_{\text{duty of }t}$ that can be sustained without temperature slope exceeding $0.5^{\circ}$C/s during the plateau phase.

\item Compute the thermal margin:
\[
\theta^{\mathrm{therm}}_{j,t} = \mathrm{clip}_{[0,1]}\!\left(\frac{T^{\mathrm{cont}}_j\big|_{\text{duty of }t}}{T^{\mathrm{req}}_{j,t}\big|_{\text{plateau}}}\right),
\]
where $T^{\mathrm{req}}_{j,t}\big|_{\text{plateau}}$ is the human plateau torque requirement from Section~\ref{sec:biomech}.

\item Document: ambient temperature, airflow (if any), time-to-derate (if derating occurred), and final motor/gearbox temperatures.
\end{enumerate}

\paragraph{Output.}
Report $\theta^{\mathrm{therm}}_{j,t}$ as a scalar, plus time traces of torque, current, DC-bus power, and temperatures with $T^{\mathrm{cont}}_j\big|_{\text{duty}}$ marked.

\subsection{Quality control and pre-registration}
\label{subsec:qc}

\paragraph{Uncertainty quantification.}
To assess measurement repeatability:
\begin{itemize}[leftmargin=12pt,itemsep=2pt]
\item Repeat 3 samples at randomly chosen $(q,\omega)$ points per joint-task.
\item Compute bootstrapped 95\% confidence intervals for $h^{(w)}_{j,t}$, $\bar{\eta}_{j,t}$, and $f^{\tau}_{c,j}$.
\item For ROM and DoF, report the minimum per-axis coverage and standard deviation across axes.
\end{itemize}

\paragraph{Sanity checks.}
Perform the following validation tests:
\begin{enumerate}[leftmargin=12pt,itemsep=2pt]
\item \textbf{Power balance:} Verify that $\int P_{\mathrm{mech}} \, dt \le \int P_{\mathrm{elec}} \, dt$ (mechanical energy out cannot exceed electrical energy in minus losses).

\item \textbf{No-load inflation:} Confirm that bandwidth with task-representative load is lower than (or equal to) no-load bandwidth. Higher loaded bandwidth indicates measurement error.

\item \textbf{Observer validation (if using torque observer):} Check that observer error stays $<5\%$ RMS against the transducer on a calibration trajectory spanning the full ROM.
\end{enumerate}

\paragraph{Pre-registration to prevent gaming.}
Publish the following \emph{before} conducting measurements:
\begin{itemize}[leftmargin=12pt,itemsep=2pt]
\item Task set $\mathcal{T}$ and task weights $\mathbf{w} = \{w_t\}$
\item Joint sets $\mathcal{J}_t$ and joint weights $\{\mathbf{u}_t\} = \{u_{j,t}\}$ for each task
\item Feature weights $\boldsymbol{\alpha} = [\alpha_{\mathrm{ROM}}, \alpha_{\mathrm{DoF}}, \alpha_{\mathrm{HEE}}, \alpha_{\mathrm{bw}}, \alpha_{\eta}, \alpha_{\mathrm{therm}}]^{\!\top}$
\item Operating bands $\mathcal{R}_{j,t}$ (as $(q,\omega)$ grids or convex hulls)
\item Target values $f^{\star}_{t,j}$ (bandwidth) and $\eta^{\star}_{t,j}$ (efficiency) for each joint-task
\end{itemize}
This prevents "weight shopping" where parameters are tuned after seeing test results to maximize the score.

\subsection{Required data products}
\label{subsec:artifacts}

When reporting HLAS, publish the following alongside the headline score:

\begin{enumerate}[leftmargin=12pt,itemsep=2pt]
\item \textbf{HEE heatmaps:} Pass/fail masks over $(q,\omega)$ for each $(j,t)$, with overlaid power-weight contours showing where human positive work concentrates.

\item \textbf{Torque-mode Bode plots:} Magnitude and phase vs. frequency for each joint, with $f^{\tau}_{c,j}$ and phase margin at crossover marked. Include a table of $b^{\tau}_{j,t}$ values.

\item \textbf{Efficiency contours:} $\eta_j(q,\omega)$ maps for each joint, plus bar charts of task-weighted averages $\bar{\eta}_{j,t}$.

\item \textbf{Thermal duty traces:} Time series of commanded torque, measured torque, current, DC-bus power, and motor/gearbox temperatures during duty tests, with $T^{\mathrm{cont}}_j\big|_{\text{duty}}$ highlighted.

\item \textbf{ROM coverage visuals:} Diagrams showing robot ROM $I^{\mathrm{rob}}_j(a)$ overlaid on functional human intervals $I^{\mathrm{func}}_{j,t}(a)$ for each axis.

\item \textbf{Raw logs:} Time-series data at ${\ge}1$\,kHz for all sensor channels, with README specifying units and sampling rates.
\end{enumerate}

These artifacts enable independent verification and facilitate comparison across systems.

\subsection{Summary: from human data to robot measurements}
\label{subsec:human_numbers}

The human demand fields $T^{\mathrm{hum}}_{j,t}(q,\omega)$ and $P^{\mathrm{hum}}_{j,t}(q,\omega)$ are derived from Section~\ref{sec:biomech} and placed on the robot's $(q,\omega)$ grid via interpolation if needed. For direct computation:
\[
T^{\mathrm{hum}}_{j,t}(q,\omega) = \frac{P^{\mathrm{hum}}_{j,t}(q,\omega)}{\omega}, \qquad
w(q,\omega) \propto \max\big(P^{\mathrm{hum}}_{j,t}(q,\omega), 0\big), \quad \sum w = 1.
\]

Optionally, enforce a conservative headroom factor by scaling human demands by $(1+\delta)$ with $\delta \in [0.05, 0.15]$ inside the HEE test (Eq.~\ref{eq:hee}). This builds in a safety margin and reduces sensitivity to measurement noise.

\subsection{Quick reference: experimental inputs to HLAS}

Table~\ref{tab:hlas_experimental_vars} provides a complete mapping from HLAS symbols to measurement procedures, serving as a quick reference for practitioners.

\begin{small}
\setlength{\tabcolsep}{5pt}
\renewcommand{\arraystretch}{1.12}
\begin{longtable}{@{}L{0.25\linewidth} L{0.20\linewidth} L{0.53\linewidth}@{}}
\caption{Experimentally obtained inputs for Eq.\,\ref{eq:hlas-expanded}. Each row specifies the HLAS symbol, the corresponding equation, and the minimal measurement procedure. Optional diagnostics (bottom three rows) are recommended for reporting but not required in the main score when HEE is used.}
\label{tab:hlas_experimental_vars}\\
\toprule
\textbf{Factor} & \textbf{Symbol (Eq.)} & \textbf{Measurement procedure $\rightarrow$ scalar used in HLAS} \\
\midrule
\endfirsthead
\toprule
\textbf{Factor} & \textbf{Symbol (Eq.)} & \textbf{Measurement procedure $\rightarrow$ scalar used in HLAS} \\
\midrule
\endhead
\midrule \multicolumn{3}{r}{\emph{Continued on next page}} \\
\endfoot
\bottomrule
\endlastfoot

\textbf{ROM coverage} & $\rho^{\mathrm{ROM}}_{j,t}$ (Eq.~\ref{eq:rom}) &
Current-limited slow sweeps ($<0.5$\,rad/s) on each axis $a \in \mathcal{A}_{j,t}$ to obtain safe ROM $I^{\mathrm{rob}}_j(a)$; intersect with functional human ROM $I^{\mathrm{func}}_{j,t}(a)$ and average across axes
$\rightarrow \rho^{\mathrm{ROM}}_{j,t} = \frac{1}{|\mathcal{A}_{j,t}|} \sum_a \frac{|I^{\mathrm{rob}}_j \cap I^{\mathrm{func}}_{j,t}|}{|I^{\mathrm{func}}_{j,t}|}$. \\[4pt]

\textbf{DoF sufficiency} & $d^{\mathrm{DoF}}_{j,t}$ (Eq.~\ref{eq:dof}) &
Command chirps (${\pm}5^{\circ}$, 0.5--5\,Hz) on each axis; verify independent actuation (cross-axis coupling $<10\%$ RMS)
$\rightarrow d^{\mathrm{DoF}}_{j,t} = \text{fraction of axes that pass}$ (1.0 only if all pass). \\[4pt]

\textbf{HEE} & $h^{(w)}_{j,t}$ (Eq.~\ref{eq:hee}) &
On $(q,\omega)$ grid over $\mathcal{R}_{j,t}$: compute human demands $T^{\mathrm{hum}}$, $P^{\mathrm{hum}}$ and weights $w \propto \max(P^{\mathrm{hum}}, 0)$; measure continuous-safe robot torque $T^{\mathrm{rob}}$ (10\,s holds, temp slope $<0.5^{\circ}$C/s) and power $P^{\mathrm{rob}} = T^{\mathrm{rob}} \omega$; mark pass if \emph{both} $T^{\mathrm{rob}} \ge T^{\mathrm{hum}}$ and $P^{\mathrm{rob}} \ge P^{\mathrm{hum}}$ at same $(q,\omega)$
$\rightarrow h^{(w)}_{j,t} = \sum w \cdot \mathbb{1}[\text{pass}]$. \\[4pt]

\textbf{Torque bandwidth} & $b^{\tau}_{j,t}$ (Eq.~\ref{eq:bandwidth}) &
Torque-mode sine sweep (0.5--60\,Hz, amplitude ${\sim}10\%$ continuous torque) with task-representative load; identify $-3$\,dB crossover $f^{\tau}_{c,j}$
$\rightarrow b^{\tau}_{j,t} = \mathrm{clip}_{[0,1]}(f^{\tau}_{c,j} / f^{\star}_{t,j})$. \\[4pt]

\textbf{Task-weighted efficiency} & $\eta_{j,t}$ (Eq.~\ref{eq:eff}) &
At $(q,\omega)$ samples (sub-thermal loads), measure $P_{\mathrm{mech}} = T_{\mathrm{shaft}} \omega$ and $P_{\mathrm{elec}} = V_{\mathrm{bus}} I_{\mathrm{bus}}$ (true RMS); compute $\eta_j = P_{\mathrm{mech}} / P_{\mathrm{elec}}$; average with positive-power weights
$\rightarrow \bar{\eta}_{j,t} = \sum w \eta_j$; normalize
$\rightarrow \eta_{j,t} = \mathrm{clip}_{[0,1]}(\bar{\eta}_{j,t} / \eta^{\star}_{t,j})$. \\[4pt]

\textbf{Thermal sustainability} & $\theta^{\mathrm{therm}}_{j,t}$ (Eq.~\ref{eq:thermal}) &
Execute task duty profile (cadence, dwell) until thermal steady state; record largest plateau torque $T^{\mathrm{cont}}_j\big|_{\text{duty}}$ sustainable without temp slope $>0.5^{\circ}$C/s
$\rightarrow \theta^{\mathrm{therm}}_{j,t} = \mathrm{clip}_{[0,1]}(T^{\mathrm{cont}}_j\big|_{\text{duty}} / T^{\mathrm{req}}_{j,t}\big|_{\text{plateau}})$. \\[6pt]

\multicolumn{3}{@{}l}{\textit{Optional diagnostics (recommended for reporting; not required in HLAS):}} \\[2pt]

\textbf{Torque margin} & $m^T_{j,t}$ (Eq.~\ref{eq:torque}) &
Lower envelope over $\mathcal{R}_{j,t}$
$\rightarrow m^T_{j,t} = \min_{(q,\omega)} \mathrm{clip}_{[0,1]}(T^{\mathrm{rob}} / T^{\mathrm{hum}})$. \\[4pt]

\textbf{Power margin} & $m^P_{j,t}$ (Eq.~\ref{eq:power}) &
Lower envelope over $\mathcal{R}_{j,t}$
$\rightarrow m^P_{j,t} = \min_{(q,\omega)} \mathrm{clip}_{[0,1]}(P^{\mathrm{rob}} / P^{\mathrm{hum}})$. \\[4pt]

\textbf{Rate margin} & $m^{\omega}_{j,t}$ (Eq.~\ref{eq:rate_diag}) &
Maximum safe joint rate with task load
$\rightarrow m^{\omega}_{j,t} = \mathrm{clip}_{[0,1]}(\omega^{\max}_j / \omega^{\mathrm{req}}_{j,t})$. \\
\end{longtable}
\end{small}

\section{Benchmark Protocols}
\label{sec:benchmarks}

\subsection{Purpose: bridging joint capabilities to task performance}

This section provides standardized benchmarks that bridge joint-level actuator capabilities (measured via the protocols in Section~\ref{sec:hlas_protocols}) to task-level functional performance. The benchmarks serve two complementary purposes:

\begin{enumerate}[leftmargin=12pt,itemsep=2pt]
\item \textbf{Joint-level validation:} Dynamometer tests under controlled conditions establish actuator and transmission limits: continuous-safe torque maps, bandwidth, efficiency, thermal plateaus, and transparency parameters. These tests produce the inputs to HLAS (Eq.~\ref{eq:hlas-expanded}).

\item \textbf{Task-level validation:} Whole-robot trials verify that joint-level capabilities translate into human-relevant functional performance on representative tasks (walking, lifting, reaching, manipulation). These tests do not enter HLAS directly but provide empirical evidence that the score reflects real-world capability.
\end{enumerate}

Together, joint and task benchmarks make HLAS claims auditable and comparable across systems, while exposing how actuator trade-offs (e.g., torque density versus bandwidth) manifest in integrated performance.

\paragraph{Scope and organization.}
This section is organized into joint-level benchmarks (Section~\ref{subsec:joint_bench}) and task-level benchmarks (Section~\ref{subsec:task_bench}). Each benchmark specifies: (i) purpose, (ii) required apparatus, (iii) step-by-step protocol, and (iv) reporting requirements. A summary at the end (Section~\ref{subsec:benchmark_summary}) explains how benchmark outputs feed HLAS and diagnostic metrics.

\subsection{Common experimental conditions}

Unless stated otherwise, all benchmarks are conducted under the following standardized conditions to ensure comparability:

\paragraph{Environmental.}
Ambient temperature $25 \pm 2^{\circ}$C in still air. If forced airflow (cooling fans, HVAC) is used, document: (i) device type and model, (ii) distance from joint or robot, and (iii) measured flow rate in m/s at the joint surface. Natural convection (still air) is preferred for reproducibility.

\paragraph{Instrumentation.}
All joint-level tests require the sensors specified in Section~\ref{sec:hlas_protocols}: high-resolution encoders, torque sensors (or validated observers), DC-bus power meters (true RMS, ${\ge}1$\,kHz), and temperature sensors on motor windings and gearbox housing. Task-level tests additionally use motion capture (or on-board state estimation), force/torque sensors (instrumented handles, force plates, insole sensors), and synchronized data logging.

\paragraph{Data logging.}
Record the signal vector $\{q, \omega, \tau, V_{\mathrm{bus}}, I_{\mathrm{bus}}, T_{\mathrm{motor}}, T_{\mathrm{gear}}\}$ at ${\ge}1$\,kHz for all joints. Store raw time series without on-device filtering (post-processing is permitted for analysis). Synchronize all channels to a common clock (NTP for networked sensors, hardware trigger for local acquisition).

\paragraph{Calibration.}
Provide calibration certificates or in-lab cross-calibration for torque sensors, angle encoders, and electrical power meters. Document uncertainty (e.g., ${\pm}0.5\%$ full-scale for torque, ${\pm}0.2^{\circ}$ for angle).

\subsection{Task set and benchmark mapping}

Recall from Section~\ref{sec:biomech} that our task library $\mathcal{T}$ contains four core functional tasks:

\begin{itemize}[leftmargin=14pt,itemsep=2pt]
\item $t_{\mathrm{gait}}$: Level walking (1.2--1.4\,m/s) and stair ascent/descent (17--18\,cm rise), emphasizing ankle positive power and knee/hip support torque.
\item $t_{\mathrm{lift}}$: Repetitive lifts from floor to waist height (15--25\,kg), emphasizing sustained knee/hip torque and thermal endurance.
\item $t_{\mathrm{reach}}$: Shoulder-height reaching with 5\,kg payload at moderate speed (${\sim}0.6$\,m/s), emphasizing shoulder/elbow torque with endpoint precision.
\item $t_{\mathrm{hand}}$: Fast hand actions including standardized grasps (power, key, tip) and rapid open/close (4--6\,Hz), emphasizing low-torque bandwidth and force control.
\end{itemize}

Each task $t$ maps to one or more benchmarks in Section~\ref{subsec:task_bench}. Optional safety-oriented tasks (moderate running at 2.5--3.0\,m/s, ballistic shoulder internal rotation during throwing) are used to set rate requirements $\omega^{\mathrm{req}}_{j,t}$ but are not required for basic HLAS computation.

\subsection{Joint-level dynamometer benchmarks}
\label{subsec:joint_bench}

\paragraph{Purpose.}
Establish joint-level performance limits under controlled loading, providing the continuous-safe torque maps $T^{\mathrm{rob}}_j(q,\omega)$, efficiency $\eta_j(q,\omega)$, bandwidth $f^{\tau}_{c,j}$, and thermal plateaus $T^{\mathrm{cont}}_j\big|_{\text{duty}}$ that enter HLAS.

\paragraph{Required apparatus.}
\begin{itemize}[leftmargin=12pt,itemsep=2pt]
\item Rotary dynamometer or grounded torque cell with motorized rate control
\item Posture fixture with angle resolution ${\le}0.2^{\circ}$
\item In-line torque transducer (or validated torque observer with RMS error $<5\%$)
\item Load emulator capable of realizing task-representative reflected inertia and compliance
\item Calibration certificates for torque, angle, and electrical power measurements
\end{itemize}

\subsubsection{Benchmark 1: Isometric plateaus}

\paragraph{Purpose.}
Quantify continuous-safe torque at key postures and characterize steady-state ripple and thermal rise.

\paragraph{Protocol.}
\begin{enumerate}[leftmargin=12pt,itemsep=2pt]
\item Select test postures from the task bands $\mathcal{R}_{j,t}$ (e.g., knee at $60^{\circ}$ flexion for stair support, ankle at $10^{\circ}$ plantarflexion for push-off).
\item For each posture, command a step torque and ramp until thermal derating onset, current limit, or control saturation.
\item Settle at the largest torque maintainable for ${\ge}10$\,s with temperature slope $<0.5^{\circ}$C/s.
\item Repeat 3 times per posture and average.
\end{enumerate}

\paragraph{Report.}
\begin{itemize}[leftmargin=12pt,itemsep=2pt]
\item Continuous-safe torque (mean over last 3\,s of hold)
\item RMS torque ripple (after removing DC component)
\item Angle hold error (RMS deviation from commanded posture)
\item Temperature rise from baseline (motor and gearbox)
\end{itemize}
This defines the posture-specific isometric point $T^{\mathrm{rob}}_j(q,0)$ in the torque-speed map.

\subsubsection{Benchmark 2: Isovelocity sweeps}

\paragraph{Purpose.}
Obtain torque-speed curves $T^{\mathrm{rob}}_j(\omega)$, derived power $P^{\mathrm{rob}}_j(\omega) = \omega T^{\mathrm{rob}}_j(\omega)$, and efficiency $\eta_j(\omega)$ under controlled thermal conditions.

\paragraph{Protocol.}
\begin{enumerate}[leftmargin=12pt,itemsep=2pt]
\item Position the joint at mid-range within the task band $\mathcal{R}_{j,t}$.
\item Perform isovelocity sweeps over task-relevant rates:
\begin{itemize}[leftmargin=12pt,itemsep=1pt]
\item Upper-limb joints: $\omega \in [0, 20]$\,rad/s
\item Lower-limb joints: $\omega \in [0, 12]$\,rad/s (extend if design targets higher rates)
\end{itemize}
\item At each rate $\omega$, ramp commanded torque to the largest steady value supportable for ${\ge}5$\,s without temperature slope $>0.5^{\circ}$C/s or current saturation.
\item Record continuous-safe torque $T^{\mathrm{rob}}_j(q,\omega)$ and compute mechanical power $P^{\mathrm{rob}}_j = T^{\mathrm{rob}}_j \cdot \omega$.
\item Measure DC-bus power $P_{\mathrm{elec}} = V_{\mathrm{bus}} \cdot I_{\mathrm{bus}}$ (true RMS) and compute efficiency $\eta_j = P_{\mathrm{mech}} / P_{\mathrm{elec}}$.
\item Optionally, capture short-term peak ($<500$\,ms) by applying a brief torque burst and logging maximum non-derated torque.
\end{enumerate}

\paragraph{Report.}
\begin{itemize}[leftmargin=12pt,itemsep=2pt]
\item Curves: $T^{\mathrm{rob}}_j(\omega)$, $P^{\mathrm{rob}}_j(\omega)$, $\eta_j(\omega)$
\item Thermal state at each point (motor/gear temperatures)
\item Peak (sub-500\,ms) torque/power as dashed overlay on continuous-safe curves
\item Annotate regions where thermal, current, or control limits bind
\end{itemize}

\subsubsection{Benchmark 3: Transparency and backdrivability}

\paragraph{Purpose.}
Identify reflected inertia $J_{\mathrm{ref}}$, friction parameters (Coulomb $f_c$, viscous $b$), and gravity-compensated backdrive torque $\tau_{\mathrm{bd}}$. These inform safe interaction limits and stable impedance ranges.

\paragraph{Protocol.}
\begin{enumerate}[leftmargin=12pt,itemsep=2pt]
\item Enable gravity compensation (feedforward model only, no feedback torque).
\item Set commanded torque to zero or zero-mean (no active assistance/resistance).
\item Impose small manual perturbations across the ROM and record $\tau$, $\omega$, and $\dot{\omega}$.
\item Fit a standard friction model via least squares with robust loss:
\[
\tau \approx J_{\mathrm{ref}} \dot{\omega} + b \omega + f_c \, \mathrm{sign}(\omega).
\]
\item Additionally, perform slow sinusoidal backdrive (${\le}0.5$\,Hz, ${\pm}30^{\circ}$ amplitude) and report the 95th percentile of $|\tau|$ after removing gravity.
\end{enumerate}

\paragraph{Report.}
\begin{itemize}[leftmargin=12pt,itemsep=2pt]
\item Gravity-compensated backdrive torque $\tau_{\mathrm{bd}}$: median and 95th percentile
\item Identified parameters $(J_{\mathrm{ref}}, b, f_c)$ with 95\% confidence intervals
\item Residual plot (e.g., Bland-Altman) comparing model predictions to measurements
\item Discuss implications for stable impedance range (Z-width)
\end{itemize}

\subsubsection{Benchmark 4: Torque-mode bandwidth}

\paragraph{Purpose.}
Quantify closed-loop torque tracking bandwidth and stability margins with realistic loading, providing $f^{\tau}_{c,j}$ for the bandwidth margin $b^{\tau}_{j,t}$ in Eq.~\ref{eq:bandwidth}.

\paragraph{Protocol.}
\begin{enumerate}[leftmargin=12pt,itemsep=2pt]
\item Mount a task-representative inertial/compliant load.
\item Position at mid-range posture within $\mathcal{R}_{j,t}$.
\item Run \textbf{small-signal} sine sweep (0.5--60\,Hz, amplitude ${\sim}10\%$ of continuous torque) in torque mode. Record commanded and actual torque.
\item Compute frequency response function $G_{\tau}(f) = \mathrm{FFT}[\tau_{\mathrm{act}}] / \mathrm{FFT}[\tau_{\mathrm{cmd}}]$.
\item Identify $-3$\,dB crossover frequency $f^{\tau}_{c,j}$ where $|G_{\tau}(f)| = 1/\sqrt{2}$.
\item Optionally, run \textbf{large-signal} multisine or PRBS excitation (30--50\% of continuous torque) to assess nonlinearity and saturation.
\item Repeat with and without the task-representative load to show the effect of loading.
\end{enumerate}

\paragraph{Report.}
\begin{itemize}[leftmargin=12pt,itemsep=2pt]
\item Small-signal $-3$\,dB crossover $f^{\tau}_{c,j}$
\item Large-signal effective crossover (if nonlinearity appears)
\item Phase margin at crossover
\item Bode plots (magnitude and phase) with values at 1, 5, 10, 30\,Hz
\item Comparison of loaded vs. no-load bandwidth
\end{itemize}

\subsubsection{Benchmark 5: Thermal duty and derating}

\paragraph{Purpose.}
Measure time-to-derate and continuous torque/power trajectories under task-representative duty cycles, providing $T^{\mathrm{cont}}_j\big|_{\text{duty of }t}$ for the thermal margin $\theta^{\mathrm{therm}}_{j,t}$ in Eq.~\ref{eq:thermal}.

\paragraph{Protocol.}
\begin{enumerate}[leftmargin=12pt,itemsep=2pt]
\item Select a representative workload profile for each joint-task pair:
\begin{itemize}[leftmargin=12pt,itemsep=1pt]
\item $t_{\mathrm{gait}}$, ankle: Push-off bursts at 8--12\,rad/s, 150--200\,ms duration, 1\,Hz cadence
\item $t_{\mathrm{gait}}$, knee/hip: Support plateaus at 1--2\,rad/s, ${\sim}70\%$ of continuous torque, 2--4\,s holds
\item $t_{\mathrm{reach}}$, shoulder/elbow: Ramps to peak over 1\,s, hold 2--3\,s, return over 1\,s, repeat
\end{itemize}
\item Execute the duty profile continuously for ${\ge}5$ thermal time constants or until temperature reaches a steady trend (slope $<0.1^{\circ}$C/min).
\item If thermal derating occurs, reduce commanded torque to the largest sustainable level and continue logging.
\item Record the steady-state plateau torque $T^{\mathrm{cont}}_j\big|_{\text{duty}}$ (largest torque with temperature slope $<0.5^{\circ}$C/s).
\end{enumerate}

\paragraph{Report.}
\begin{itemize}[leftmargin=12pt,itemsep=2pt]
\item Time traces: $T_{\mathrm{cont}}(t)$, $P_{\mathrm{cont}}(t)$, $T_{\mathrm{motor}}(t)$, $T_{\mathrm{gear}}(t)$
\item Time-to-derate (if any)
\item Steady plateau torque $T^{\mathrm{cont}}_j\big|_{\text{duty}}$ and comparison to human requirement $T^{\mathrm{req}}_{j,t}\big|_{\text{plateau}}$
\item Ambient temperature and airflow specification
\end{itemize}

\subsection{Task-level benchmarks}
\label{subsec:task_bench}

\paragraph{Purpose.}
Verify that joint-level capabilities translate into functional task performance, validating the human-derived operating bands $\mathcal{R}_{j,t}$ and exposing whole-body coordination issues not captured by single-joint tests.

\paragraph{Common setup for all task benchmarks.}
\begin{itemize}[leftmargin=12pt,itemsep=2pt]
\item Motion capture system (or on-board state estimator) logging end-effector kinematics at ${\ge}100$\,Hz
\item Force/torque sensors: instrumented handles/boxes, force plates, or insole sensors as appropriate
\item Synchronize robot signals with external sensors (NTP or hardware trigger)
\item Conduct ${\ge}3$ trials per condition, report mean ${\pm}$ SD and 5th/95th percentiles where variability matters
\end{itemize}

\subsubsection{Benchmark 6: Gait module ($t_{\mathrm{gait}}$)}

\paragraph{Purpose.}
Instantiate the gait task $t_{\mathrm{gait}}$ (level walking and stairs) and verify ankle positive power, knee/hip support torque, and footfall dynamics in human-like conditions.

\paragraph{Apparatus.}
Instrumented treadmill with force plates, stair rig (17--18\,cm rise, ${\sim}28$\,cm tread), motion capture or on-board state estimation, insole sensors (optional).

\paragraph{Protocol.}
\begin{enumerate}[leftmargin=12pt,itemsep=2pt]
\item \textbf{Level walking:} Walk at 1.2--1.4\,m/s on instrumented treadmill for 2\,min. Use nominal human-like kinematics from Section~\ref{sec:biomech}. Do not exceed continuous-safe limits from dynamometer tests.

\item \textbf{Stair ascent:} Ascend stairs (17--18\,cm rise) for 1\,min at self-selected comfortable cadence.

\item \textbf{Stair descent:} Descend stairs for 1\,min.

\item \textbf{Optional: Moderate running:} Run at 2.5--3.0\,m/s for 1\,min to test higher-rate ankle/hip power (extends $t_{\mathrm{gait}}$ to running regime).
\end{enumerate}

\paragraph{Report.}
\begin{itemize}[leftmargin=12pt,itemsep=2pt]
\item \textbf{Ankle performance:} Cycle-averaged positive work (J/stride), peak positive power (W), angular rates during push-off
\item \textbf{Knee/hip performance:} Peak extension moments (Nm), support torque duty (fraction of stride above 70\% continuous torque)
\item \textbf{Gait quality:} Duty factor, stride frequency, stride length
\item \textbf{Efficiency:} Estimated mechanical-to-electrical efficiency over stance (from DC-bus accounting)
\item \textbf{Footfall dynamics:} Peak vertical GRF (N), GRF timing relative to heel strike, any missed steps or instabilities
\item \textbf{Thermal:} Joint temperatures at end of trial; note any derating
\end{itemize}

\paragraph{Validation.}
Compare measured ankle positive work, knee support torque, and joint angular rates to the human bands $\mathcal{R}_{j,t_{\mathrm{gait}}}$ used in HEE. This confirms that dynamometer-derived capabilities translate to integrated locomotion.

\subsubsection{Benchmark 7: Lift and carry ($t_{\mathrm{lift}}$)}

\paragraph{Purpose.}
Instantiate the lifting task $t_{\mathrm{lift}}$ and assess knee/hip torque endurance, thermal behavior, and base stability under repetitive manual handling.

\paragraph{Apparatus.}
Instrumented boxes (15\,kg and 25\,kg with handles and IMUs), force plates, motion capture, floor markers for standardized foot placement.

\paragraph{Protocol.}
\begin{enumerate}[leftmargin=12pt,itemsep=2pt]
\item Mark foot placement positions and trunk posture targets to standardize lifting strategy (reduce inter-trial variance).
\item Lift box from floor to waist height (${\sim}1.0$\,m), carry 5\,m, set down, return. Repeat continuously for 2\,min.
\item Perform separate trials with 15\,kg and 25\,kg boxes.
\item Do not exceed continuous-safe limits. If thermal derating begins, note time and continue at reduced torque.
\end{enumerate}

\paragraph{Report.}
\begin{itemize}[leftmargin=12pt,itemsep=2pt]
\item \textbf{Knee/hip performance:} Torque duty cycles (fraction of time above 70\% continuous torque), peak torques during lift phase
\item \textbf{Thermal:} Temperature trajectories, whether they stay within continuous-safe plateau from Benchmark 5, continuous torque at end of 2\,min trial
\item \textbf{Stability:} Center-of-pressure (CoP) margin (minimum distance to support polygon edge), any foot slips or path deviations
\item \textbf{Success rate:} Fraction of lifts completed without assistance, drops, or instability
\end{itemize}

\paragraph{Validation.}
Compare end-of-trial continuous torque to $T^{\mathrm{cont}}_j\big|_{\text{duty of }t_{\mathrm{lift}}}$ from thermal duty tests. Verify that thermal margins predicted from single-joint tests hold during integrated task execution.

\subsubsection{Benchmark 8: Reach with payload ($t_{\mathrm{reach}}$)}

\paragraph{Purpose.}
Instantiate the reaching task $t_{\mathrm{reach}}$ and test shoulder/elbow torque delivery with endpoint precision under a 5\,kg payload at moderate speed.

\paragraph{Apparatus.}
5\,kg instrumented payload (e.g., tool handle with force/torque sensor and IMU), motion capture or on-board endpoint tracking, visual targets at shoulder height.

\paragraph{Protocol.}
\begin{enumerate}[leftmargin=12pt,itemsep=2pt]
\item From neutral pose, execute point-to-point reaches to shoulder height (${\sim}1.4$\,m above ground).
\item Target endpoint speed ${\approx}0.6$\,m/s (moderate, human-like speed).
\item Perform 20 repetitions with randomized target directions within a $30^{\circ}$ cone, matching the kinematic band $\mathcal{R}_{j,t_{\mathrm{reach}}}$ from Section~\ref{sec:biomech}.
\end{enumerate}

\paragraph{Report.}
\begin{itemize}[leftmargin=12pt,itemsep=2pt]
\item \textbf{Joint performance:} Peak shoulder and elbow torques (Nm), torque profiles during acceleration/deceleration phases
\item \textbf{Endpoint accuracy:} RMS position error (target $<2$\,cm), 95th percentile error
\item \textbf{Settling:} Time to settle within 2\,cm of target, overshoot percentage
\item \textbf{Success rate:} Fraction of reaches completed without torque saturation, instability, or collision
\item \textbf{Efficiency:} Average mechanical power and electrical power during reach phase
\end{itemize}

\paragraph{Validation.}
Verify that measured shoulder/elbow torques and rates fall within the bands $\mathcal{R}_{j,t_{\mathrm{reach}}}$ used for HEE, confirming that joint-level capabilities support the intended task kinematics.

\subsubsection{Benchmark 9: Hand dexterity ($t_{\mathrm{hand}}$)}

\paragraph{Purpose.}
Instantiate the hand task $t_{\mathrm{hand}}$ and characterize low-torque, high-bandwidth manipulation: grasp force control, rapid open/close, and stiction/friction limits.

\paragraph{Apparatus.}
Instrumented grasp objects (cylinders, blocks, small tools with embedded force sensors), high-speed camera or finger tracking (60--120\,fps), commanded force profiles.

\paragraph{Protocol.}
\begin{enumerate}[leftmargin=12pt,itemsep=2pt]
\item \textbf{Standardized grasps:} Execute power, key, and tip pinch grasps on instrumented objects. Ramp commanded force from 0 to 20\,N over 2\,s, hold 2\,s, release. Repeat 5 times per grasp type.

\item \textbf{Rapid open/close:} Perform full-excursion finger flexion/extension cycles at 1, 2, 4, and 6\,Hz for 10\,s each, spanning the 4--6\,Hz functional range from Section~\ref{sec:biomech}.
\end{enumerate}

\paragraph{Report.}
\begin{itemize}[leftmargin=12pt,itemsep=2pt]
\item \textbf{Force bandwidth:} Frequency at $-3$\,dB of commanded-to-measured force FRF
\item \textbf{Minimum controllable force:} Smallest force achievable above sensor noise floor (target $<1$\,N)
\item \textbf{Stiction:} Breakaway force (force to initiate motion from rest), hysteresis in force-displacement curves
\item \textbf{Repeatability:} Cycle-to-cycle variation in peak force, timing, and position (report standard deviation)
\item \textbf{Success rate at 6\,Hz:} Fraction of cycles that reach full flexion/extension without missing or stalling
\end{itemize}

\paragraph{Validation.}
Confirm that measured force bandwidth and achievable rates meet or exceed the 4--6\,Hz functional hand rates from Section~\ref{sec:biomech}, validating that wrist/hand actuators support dexterous manipulation.

\subsection{Quality control and data release}
\label{subsec:benchmark_qc}

To ensure reproducibility and enable cross-system comparison, publish the following with every benchmark report:

\begin{enumerate}[leftmargin=12pt,itemsep=2pt]
\item \textbf{Raw time series} (${\ge}1$\,kHz for joint-level, ${\ge}100$\,Hz for task-level) with README specifying channel names, units, and sampling rates.

\item \textbf{Configuration file} documenting: posture ranges, payloads, controller gains, PID parameters, force/impedance control settings, and any airflow or thermal management details.

\item \textbf{Calibration notes} for torque, angle, force, and electrical power sensors, including uncertainty estimates.

\item \textbf{Statistical summaries:} Where stochasticity matters (foot placement, hand contact forces, endpoint errors), report distributions (median, 5th/95th percentiles, interquartile range) in addition to means.

\item \textbf{Failure logs:} Document any trials where torque saturation, thermal derating, control instability, or mechanical interference occurred. Explain whether failures were due to actuator limits, controller limits, or experimental setup.
\end{enumerate}

\subsection{Summary: how benchmarks feed HLAS and diagnostics}
\label{subsec:benchmark_summary}

\paragraph{Joint-level benchmarks produce HLAS inputs.}
Benchmarks 1--5 yield the experimental scalars that enter Eq.~\ref{eq:hlas-expanded}:
\begin{itemize}[leftmargin=12pt,itemsep=2pt]
\item \textbf{Isometric plateaus \& isovelocity sweeps} (Benchmarks 1--2): Continuous-safe torque maps $T^{\mathrm{rob}}_j(q,\omega)$, derived power $P^{\mathrm{rob}}_j(q,\omega)$, and efficiency $\eta_j(q,\omega)$ over the operating bands $\mathcal{R}_{j,t}$.
\item \textbf{Transparency test} (Benchmark 3): Reflected inertia $J_{\mathrm{ref}}$, friction parameters $(b, f_c)$, and backdrive torque $\tau_{\mathrm{bd}}$ as diagnostic indicators of safe impedance range.
\item \textbf{Torque bandwidth} (Benchmark 4): Closed-loop $-3$\,dB crossover $f^{\tau}_{c,j}$ feeding the bandwidth margin $b^{\tau}_{j,t}$ (Eq.~\ref{eq:bandwidth}).
\item \textbf{Thermal duty} (Benchmark 5): Continuous plateau torque $T^{\mathrm{cont}}_j\big|_{\text{duty}}$ at task-representative duty cycles, feeding the thermal margin $\theta^{\mathrm{therm}}_{j,t}$ (Eq.~\ref{eq:thermal}).
\end{itemize}

\paragraph{Task-level benchmarks validate HLAS and expose integration issues.}
Benchmarks 6--9 do \emph{not} directly enter HLAS but serve two critical functions:

\begin{enumerate}[leftmargin=12pt,itemsep=2pt]
\item \textbf{Empirical validation:} Task benchmarks confirm that joint-level capabilities (measured on dynamometers) translate to functional performance on integrated robots. For example, if dynamometer tests show $h^{(w)}_{\texttt{ankle,gait}} = 0.85$ but the gait benchmark reveals that the robot cannot sustain 2\,min of walking due to thermal derating, this exposes a gap between component capability and system integration.

\item \textbf{Whole-body coordination:} Task benchmarks capture effects not visible in single-joint tests: foot-ground interaction, base stability (CoP margins), controller coordination across joints, and thermal coupling between neighboring actuators. A high HLAS joint score with poor task performance suggests controller limitations, kinematic redundancy issues, or structural compliance problems rather than actuator inadequacy.
\end{enumerate}

\paragraph{Recommended reporting structure.}
When publishing HLAS, include:
\begin{itemize}[leftmargin=12pt,itemsep=2pt]
\item The headline HLAS score (Eq.~\ref{eq:hlas-compact}) with task/joint/factor decomposition
\item Joint-level benchmark results (Benchmarks 1--5) with all required data products from Section~\ref{subsec:benchmark_qc}
\item Task-level benchmark results (Benchmarks 6--9) demonstrating that HLAS translates to functional capability
\item Diagnostic discussion of any discrepancies between joint-level scores and task-level performance
\end{itemize}

This complete picture makes "human-level actuation" claims auditable, reproducible, and interpretable.

\section{Worked Example: Computing HLAS for a Theoretical Robot}
\label{sec:hlas_example}

\subsection{Purpose and scope}

This section provides a complete, step-by-step HLAS computation for a hypothetical humanoid robot, illustrating how measurement protocols (Section~\ref{sec:hlas_protocols}) and aggregation formulas (Section~\ref{sec:hlas}) combine to produce a single interpretable score with full diagnostic decomposition. All values are synthetic but internally consistent and representative of achievable actuator performance.

\textbf{Note on reference data.} Human capability data exhibit measurement and population variability across studies. The values here (e.g., 2.0--2.5\,W/kg for ankle push-off) are representative but may differ from multi-study ranges in Section~\ref{sec:biomech} (e.g., 2.5--3.5\,W/kg). HLAS requires internal consistency within an evaluation and when comparing robots. The framework is agnostic to which specific reference dataset is chosen. The computational methods and interpretive logic demonstrated here apply regardless of reference values selected.

The example demonstrates how to: (1) construct per-joint, per-task feature vectors $\mathbf{x}_{j,t}$ from measurements; (2) apply the HEE (Human-Equivalence Envelope) test for simultaneous torque-power delivery; (3) aggregate joint-task scores into task-level and overall HLAS; (4) interpret the decomposition to diagnose performance; and (5) assess sensitivity to design choices (headroom factors, weight selection).

\textbf{Scope:} This pedagogical example uses simplified kinematics (one DoF per joint) and three tasks. Real implementations would include the full DoF atlas (Section~\ref{sec:kinematics}) and application-specific task sets.

\subsection{Robot specification}
\label{subsec:example_setup}

\paragraph{Kinematic structure.}
Consider a humanoid robot with 12 actuated joints:
\begin{itemize}[leftmargin=12pt,itemsep=2pt]
\item \textbf{Lower limbs (6 DoF):} Hip, knee, and ankle on each leg. Each joint actuates the primary axis for locomotion: hip flexion/extension, knee flexion/extension, ankle dorsiflexion/plantarflexion.
\item \textbf{Upper limbs (6 DoF):} Shoulder, elbow, and wrist on each arm. Each joint actuates the primary axis for manipulation: shoulder flexion/extension, elbow flexion/extension, wrist flexion/extension.
\end{itemize}

\textbf{Simplification:} While the anatomical atlas (Section~\ref{sec:kinematics}) specifies that ankle and wrist are 3-DoF complexes, this example treats each as a single actuated DoF to keep the computation manageable. A full implementation would include all required axes per joint.

\paragraph{Actuator performance (measured via Section~\ref{sec:hlas_protocols}).}
Continuous-safe torque maps $T^{\mathrm{rob}}_j(q,\omega)$ were obtained via dynamometer testing (Section~\ref{subsec:joint_bench}). Key characteristics:
\begin{itemize}[leftmargin=12pt,itemsep=2pt]
\item \textbf{Lower-limb joints:} Torque decreases mildly with speed due to gearbox losses (e.g., ankle: 140\,Nm at 0\,rad/s, 110\,Nm at 12\,rad/s).
\item \textbf{Upper-limb joints:} Relatively flat torque-speed curves due to lower gear ratios (e.g., wrist: 25\,Nm at 0\,rad/s, 22\,Nm at 15\,rad/s).
\item \textbf{Efficiency:} Electromechanical efficiency $\eta_j(q,\omega)$ ranges from 0.70 to 0.82 over positive-work bands, with peak efficiency at mid-speeds (5--10\,rad/s for lower limbs, 8--12\,rad/s for upper limbs).
\end{itemize}

\paragraph{Torque-mode bandwidth (measured via Section~\ref{subsec:bandwidth}).}
Closed-loop $-3$\,dB crossover frequencies under task-representative loading:
\[
\begin{aligned}
&\text{Ankle: } f^{\tau}_{c,\text{ankle}} = 12\,\text{Hz}, \quad
\text{Knee: } f^{\tau}_{c,\text{knee}} = 7\,\text{Hz}, \quad
\text{Hip: } f^{\tau}_{c,\text{hip}} = 6\,\text{Hz}, \\
&\text{Shoulder: } f^{\tau}_{c,\text{shoulder}} = 9\,\text{Hz}, \quad
\text{Elbow: } f^{\tau}_{c,\text{elbow}} = 10\,\text{Hz}, \quad
\text{Wrist: } f^{\tau}_{c,\text{wrist}} = 12\,\text{Hz}.
\end{aligned}
\]

\paragraph{Known limitations.}
\begin{itemize}[leftmargin=12pt,itemsep=2pt]
\item \textbf{Wrist ROM:} Robot achieves ${\pm}12^{\circ}$ flexion/extension vs. functional human requirement of ${\pm}15^{\circ}$, giving $\rho^{\mathrm{ROM}}_{\text{wrist,Reach}} = 0.80$.
\item \textbf{Hip/knee HEE in stairs:} Low HEE coverage (${\sim}0.085$) due to insufficient torque at low angular rates where stair support occurs. This is a known design trade-off (optimized for walking speeds over quasi-static support).
\end{itemize}

\subsection{Task set, weights, and targets}
\label{subsec:example_tasks}

\paragraph{Task selection and weights.}
We evaluate three tasks from $\mathcal{T}$ (Section~\ref{sec:biomech}):

\begin{itemize}[leftmargin=12pt,itemsep=2pt]
\item \textbf{\texttt{Walk}:} Level walking at 1.2--1.4\,m/s, emphasizing ankle positive power during push-off and knee/hip support during stance. Task weight $w_{\texttt{Walk}} = 0.4$ (highest priority for this general-purpose platform).

\item \textbf{\texttt{Stairs}:} Stair ascent and descent (17--18\,cm rise), emphasizing sustained knee/hip torque at low rates and ankle propulsion bursts. Task weight $w_{\texttt{Stairs}} = 0.3$.

\item \textbf{\texttt{Reach}:} Shoulder-height reaching with 5\,kg payload at ${\sim}0.6$\,m/s, emphasizing shoulder/elbow torque with endpoint precision. Task weight $w_{\texttt{Reach}} = 0.3$.
\end{itemize}

Weights sum to unity: $w_{\texttt{Walk}} + w_{\texttt{Stairs}} + w_{\texttt{Reach}} = 1.0$.

\paragraph{Joint weights within tasks.}
Joint weights $u_{j,t}$ reflect each joint's share of positive mechanical work for the task, derived from the biomechanics data in Section~\ref{sec:biomech}:

\begin{itemize}[leftmargin=12pt,itemsep=2pt]
\item \textbf{\texttt{Walk}:} Ankle 0.50, knee 0.30, hip 0.20 (ankle-dominant positive work).
\item \textbf{\texttt{Stairs}:} Ankle 0.10, knee 0.50, hip 0.40 (knee/hip-dominant quasi-static support).
\item \textbf{\texttt{Reach}:} Shoulder 0.60, elbow 0.30, wrist 0.10 (shoulder-dominant endpoint control).
\end{itemize}

Each task's joint weights sum to unity: $\sum_{j \in \mathcal{J}_t} u_{j,t} = 1$.

\paragraph{Feature weights.}
We use the default feature weights from Section~\ref{sec:hlas}, emphasizing HEE (simultaneous torque-power delivery) while retaining workspace and realism checks:
\[
\alpha_{\mathrm{ROM}} = 0.10, \quad
\alpha_{\mathrm{DoF}} = 0.10, \quad
\alpha_{\mathrm{HEE}} = 0.50, \quad
\alpha_{\mathrm{bw}} = 0.10, \quad
\alpha_{\eta} = 0.10, \quad
\alpha_{\mathrm{therm}} = 0.10.
\]
These weights sum to unity: $\sum_k \alpha_k = 1$.

\paragraph{Normalization targets.}
Bandwidth and efficiency targets are specified per task and joint based on the biomechanical demands in Section~\ref{sec:biomech}:

\textbf{Bandwidth targets $f^{\star}_{t,j}$ (Hz):}
\begin{itemize}[leftmargin=12pt,itemsep=2pt]
\item \texttt{Walk}: ankle 8, knee 6, hip 5
\item \texttt{Stairs}: ankle 6, knee 6, hip 6
\item \texttt{Reach}: shoulder 8, elbow 8, wrist 10
\end{itemize}

\textbf{Efficiency targets $\eta^{\star}_t$ (dimensionless):}
\begin{itemize}[leftmargin=12pt,itemsep=2pt]
\item \texttt{Walk}: 0.80
\item \texttt{Stairs}: 0.75 (lower due to low-speed operation with higher friction)
\item \texttt{Reach}: 0.80
\end{itemize}

\subsection{Detailed HEE calculation: ankle in walking}
\label{subsec:example_hee_detail}

To illustrate the HEE computation concretely, we walk through ankle push-off during \texttt{Walk} in detail.

\paragraph{Operating band and sampling.}
The ankle push-off band for walking covers posture $q = 10^{\circ}$ plantarflexion (fixed) and angular rates $\omega \in \{8, 9, 10, 11, 12\}$\,rad/s (5-point rate sweep). This simplified 1D sweep captures the dominant positive-work region from Section~\ref{sec:biomech}.

\paragraph{Human requirements.}
From Table~\ref{tab:joint_level_targets}, human ankle push-off requires moments of 1.2--1.9\,Nm/kg (90--143\,Nm for 75\,kg) and positive power of 2.0--2.5\,W/kg (150--188\,W for 75\,kg). These values vary with rate and are interpolated onto our 5-point grid:

\begin{center}
\small
\begin{tabular}{@{}r r r r@{}}
\toprule
$\omega$ (rad/s) & $T^{\mathrm{hum}}$ (Nm) & $P^{\mathrm{hum}}$ (W) & $w$ (normalized) \\
\midrule
8  & 30 & 240 & 0.151 \\
9  & 32 & 288 & 0.181 \\
10 & 34 & 340 & 0.214 \\
11 & 33 & 363 & 0.228 \\
12 & 30 & 360 & 0.226 \\
\bottomrule
\end{tabular}
\end{center}

Weights $w(\omega)$ are proportional to $P^{\mathrm{hum}}$ and sum to 1: $w(\omega) \propto P^{\mathrm{hum}}(\omega)$, $\sum w = 1$.

\paragraph{Robot performance.}
Continuous-safe torque $T^{\mathrm{rob}}_{\text{ankle}}(\omega)$ was measured via isovelocity sweeps (Benchmark 2, Section~\ref{subsec:joint_bench}), holding $q = 10^{\circ}$ and ramping torque at each rate until thermal or current limits. Derived power $P^{\mathrm{rob}} = T^{\mathrm{rob}} \cdot \omega$:

\begin{center}
\small
\begin{tabular}{@{}r r r@{}}
\toprule
$\omega$ (rad/s) & $T^{\mathrm{rob}}$ (Nm) & $P^{\mathrm{rob}}$ (W) \\
\midrule
8  & 36 & 288 \\
9  & 35 & 315 \\
10 & 34 & 340 \\
11 & 30 & 330 \\
12 & 27 & 324 \\
\bottomrule
\end{tabular}
\end{center}

\paragraph{Pass/fail test and HEE.}
For each sample, check if \emph{both} $T^{\mathrm{rob}} \ge T^{\mathrm{hum}}$ and $P^{\mathrm{rob}} \ge P^{\mathrm{hum}}$:

\begin{center}
\small
\begin{tabular}{@{}r c c c@{}}
\toprule
$\omega$ (rad/s) & Torque OK? & Power OK? & Pass? \\
\midrule
8  & $36 \ge 30$ \checkmark & $288 \ge 240$ \checkmark & \textbf{yes} \\
9  & $35 \ge 32$ \checkmark & $315 \ge 288$ \checkmark & \textbf{yes} \\
10 & $34 \ge 34$ \checkmark & $340 \ge 340$ \checkmark & \textbf{yes} \\
11 & $30 \ge 33$ \texttimes & $330 \ge 363$ \texttimes & \textbf{no} \\
12 & $27 \ge 30$ \texttimes & $324 \ge 360$ \texttimes & \textbf{no} \\
\bottomrule
\end{tabular}
\end{center}

Sum the weights of passing samples:
\[
h^{(w)}_{\text{ankle,Walk}} = w(8) + w(9) + w(10) = 0.151 + 0.181 + 0.214 = \boxed{0.546}.
\]

\paragraph{Interpretation.}
The robot meets human ankle requirements at 8, 9, and 10\,rad/s (covering 54.6\% of the positive-work-weighted band) but fails at 11 and 12\,rad/s due to insufficient torque at higher rates. This HEE score of 0.546 reflects partial capability: the robot can handle moderate-speed walking but may struggle with faster cadences.

\subsection{Complete feature vectors for all joint-task pairs}
\label{subsec:example_joint_scores}

Table~\ref{tab:joint_scores} presents the complete feature vectors $\mathbf{x}_{j,t} = [\rho^{\mathrm{ROM}}_{j,t}, d^{\mathrm{DoF}}_{j,t}, h^{(w)}_{j,t}, b^{\tau}_{j,t}, \eta_{j,t}, \theta^{\mathrm{therm}}_{j,t}]^{\!\top}$ for all nine joint-task pairs, along with the computed joint-task scores $s_{j,t} = \boldsymbol{\alpha}^{\!\top} \mathbf{x}_{j,t}$.

\begin{table}[t]
\centering
\small
\caption{Feature vectors and joint-task scores. Each row shows the six HLAS factors for one $(j,t)$ pair and the weighted-sum score $s_{j,t} = \boldsymbol{\alpha}^{\!\top} \mathbf{x}_{j,t}$. All values are dimensionless and in $[0,1]$.}
\label{tab:joint_scores}
\begin{tabular}{@{}l l c c c c c c c@{}}
\toprule
Task & Joint & $\rho^{\mathrm{ROM}}$ & $d^{\mathrm{DoF}}$ & $h^{(w)}$ & $b^{\tau}$ & $\eta$ & $\theta^{\mathrm{therm}}$ & $s_{j,t}$ \\
\midrule
\texttt{Walk} & ankle & 0.880 & 1.000 & 0.546 & 1.000 & 0.977 & 1.000 & 0.758 \\
\texttt{Walk} & knee  & 0.900 & 1.000 & 0.284 & 1.000 & 0.920 & 0.960 & 0.620 \\
\texttt{Walk} & hip   & 1.000 & 1.000 & 0.087 & 1.000 & 0.902 & 0.956 & 0.529 \\
\midrule
\texttt{Stairs} & ankle & 1.000 & 1.000 & 0.290 & 1.000 & 1.000 & 0.970 & 0.642 \\
\texttt{Stairs} & knee  & 0.888 & 1.000 & 0.087 & 1.000 & 0.984 & 0.950 & 0.526 \\
\texttt{Stairs} & hip   & 0.933 & 1.000 & 0.085 & 1.000 & 0.971 & 0.971 & 0.530 \\
\midrule
\texttt{Reach} & shoulder & 0.967 & 1.000 & 0.397 & 1.000 & 0.971 & 0.967 & 0.689 \\
\texttt{Reach} & elbow    & 1.000 & 1.000 & 0.385 & 1.000 & 0.996 & 1.000 & 0.692 \\
\texttt{Reach} & wrist    & 0.800 & 1.000 & 0.375 & 1.000 & 0.946 & 1.000 & 0.662 \\
\bottomrule
\end{tabular}
\end{table}

\paragraph{Key observations from Table~\ref{tab:joint_scores}:}
\begin{itemize}[leftmargin=12pt,itemsep=2pt]
\item \textbf{All joints have full DoF coverage} ($d^{\mathrm{DoF}} = 1.0$): required axes are implemented and independently actuated.
\item \textbf{Bandwidth margins are all 1.0}: measured bandwidths meet or exceed targets, reflecting appropriate gear-ratio choices.
\item \textbf{ROM is high (${\ge}0.80$)} with one exception: wrist in \texttt{Reach} is limited to 0.80 due to mechanical constraints.
\item \textbf{HEE varies widely} (0.085 to 0.546): this is the dominant source of variation. Hip and knee in \texttt{Stairs} have very low HEE (${\sim}0.087$), indicating insufficient torque at the low rates where stair support occurs.
\item \textbf{Efficiency and thermal margins are strong} (${\ge}0.90$): good actuator/transmission design with adequate cooling.
\end{itemize}

\paragraph{How entries were computed (representative examples):}
\begin{itemize}[leftmargin=12pt,itemsep=2pt]
\item \textbf{ROM:} Ankle in \texttt{Walk} requires functional plantarflexion $[0, 25]^{\circ}$. Robot achieves $[-5, 22]^{\circ}$, giving overlap $22 / 25 = 0.88$.
\item \textbf{HEE:} Computed as shown in Section~\ref{subsec:example_hee_detail} for ankle; analogous for other joints.
\item \textbf{Bandwidth:} $b^{\tau}_{j,t} = \mathrm{clip}_{[0,1]}(f^{\tau}_{c,j} / f^{\star}_{t,j})$. Example: knee in \texttt{Stairs} has $f^{\tau}_{c,\text{knee}} = 7$\,Hz and $f^{\star}_{\texttt{Stairs},\text{knee}} = 6$\,Hz, giving $7/6 = 1.17 \to 1.0$ after clipping.
\item \textbf{Efficiency:} Task-weighted average efficiency $\bar{\eta}_{j,t}$ divided by target $\eta^{\star}_t$ and clipped. Example: ankle in \texttt{Walk} has $\bar{\eta} = 0.781$ and $\eta^{\star} = 0.80$, giving $0.781 / 0.80 = 0.976 \approx 0.977$.
\item \textbf{Thermal:} Ratio of continuous-safe plateau torque to human requirement. Example: knee in \texttt{Walk} sustains 48\,Nm at walking duty vs. 50\,Nm human requirement, giving $48 / 50 = 0.96$.
\end{itemize}

\subsection{Visual diagnostic tools}
\label{subsec:example_visuals}

\begin{figure}[t]
  \centering
  \includegraphics[width=0.8\linewidth]{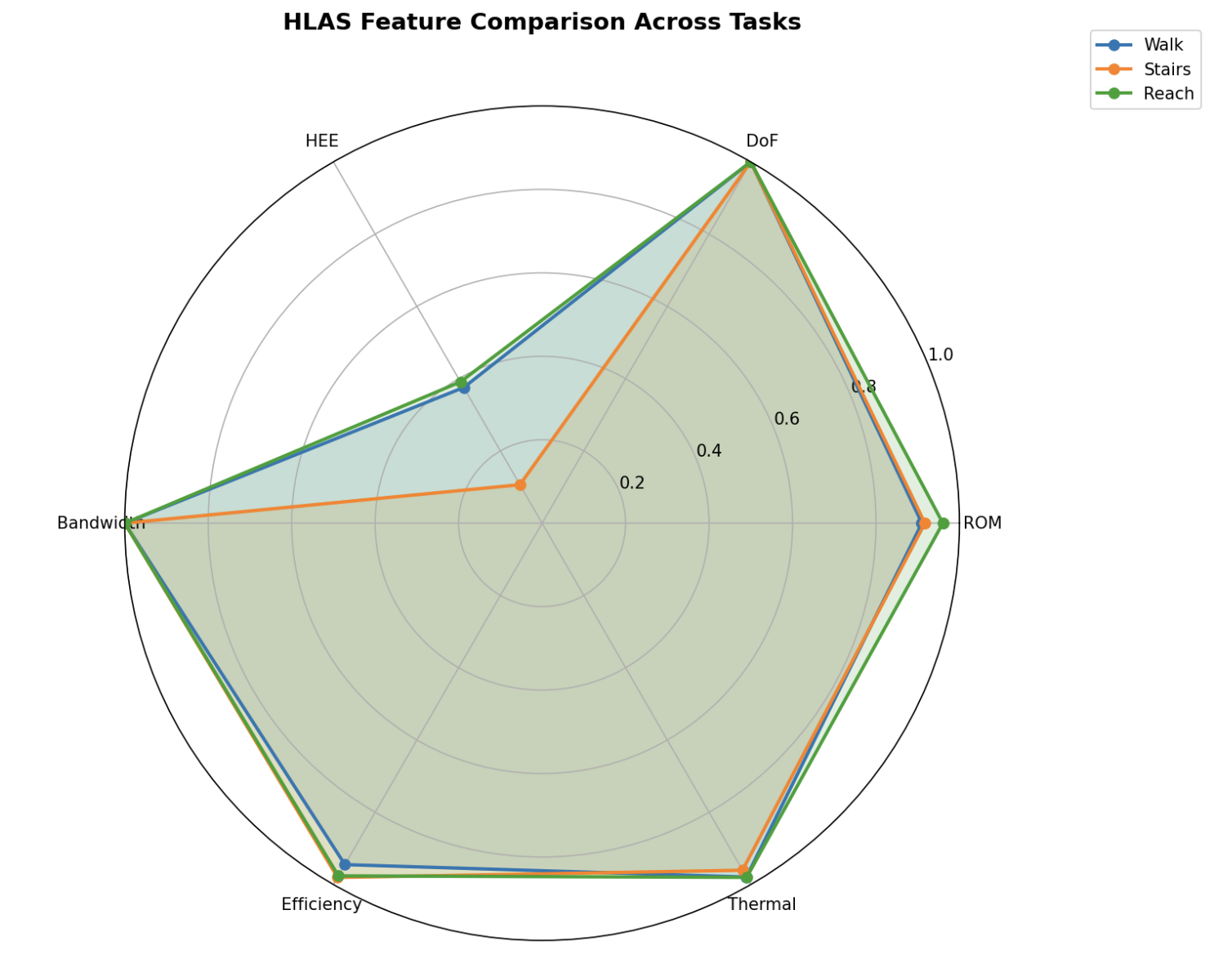}
  \caption{Radar/spider plot of the six HLAS features aggregated by task. Each axis represents one feature (ROM, DoF, HEE, bandwidth, efficiency, thermal). Values closer to the perimeter are better. \texttt{Walk} and \texttt{Reach} show balanced profiles, while \texttt{Stairs} is weak in HEE (pulled inward) due to low knee/hip torque at slow rates.}
  \label{fig:spider}
\end{figure}

\begin{figure}[t]
  \centering
  \includegraphics[width=\linewidth]{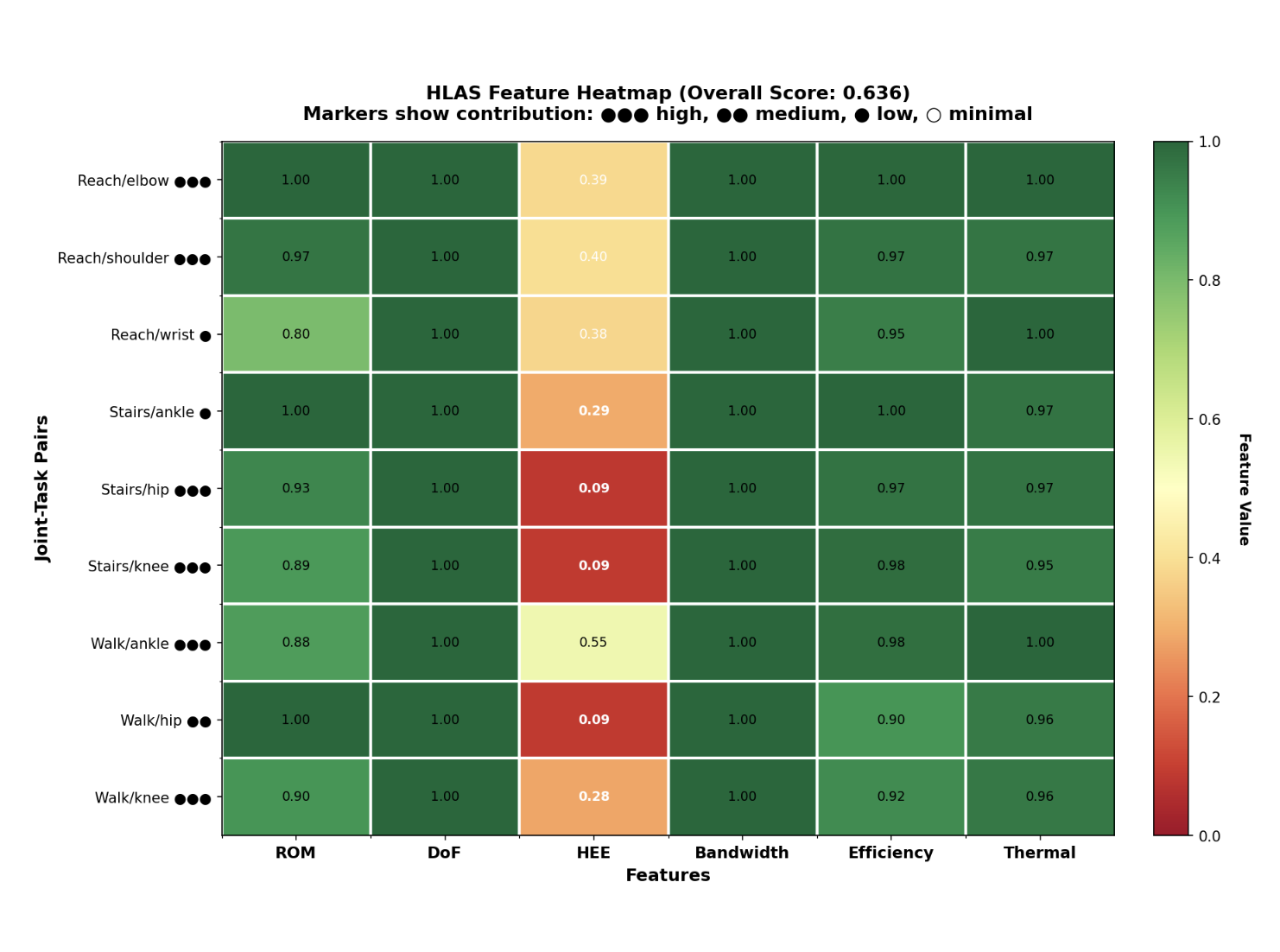}
  \caption{Heatmap of all six HLAS features for each joint-task pair. Darker cells indicate lower values. The heatmap localizes bottlenecks: knee and hip HEE in \texttt{Stairs} are dark (${\sim}0.085$), and wrist ROM in \texttt{Reach} is moderately dark (0.80). All other factors are light (strong performance).}
  \label{fig:feature-heatmap}
\end{figure}

Figures~\ref{fig:spider} and \ref{fig:feature-heatmap} provide complementary visual interpretations of Table~\ref{tab:joint_scores}.

\paragraph{Interpretation of Figure~\ref{fig:spider}.}
The radar plot shows task-level feature aggregates (averaging across joints within each task). \texttt{Stairs} is noticeably weaker in HEE, pulling the polygon inward on that axis. This reflects the low HEE scores for knee and hip in stair support. Bandwidth, efficiency, and thermal margins are strong for all tasks (polygon extends to the perimeter on those axes).

\paragraph{Interpretation of Figure~\ref{fig:feature-heatmap}.}
The heatmap localizes weaknesses to specific $(j,t)$ pairs. The dark cells for knee and hip HEE in \texttt{Stairs} (values ${\sim}0.085$) immediately identify the primary bottleneck. The moderately dark cell for wrist ROM in \texttt{Reach} (0.80) is a secondary issue. All other cells are light, indicating strong performance on those factors.

\subsection{Aggregation to final HLAS}
\label{subsec:example_agg}

\paragraph{Step 1: Task-level scores.}
Aggregate joint-task scores within each task using joint weights $u_{j,t}$:
\[
s_t = \sum_{j \in \mathcal{J}_t} u_{j,t} \, s_{j,t}.
\]

\textbf{\texttt{Walk}:}
\[
s_{\texttt{Walk}} = 0.50 \times 0.758 + 0.30 \times 0.620 + 0.20 \times 0.529 = 0.379 + 0.186 + 0.106 = \boxed{0.671}.
\]

\textbf{\texttt{Stairs}:}
\[
s_{\texttt{Stairs}} = 0.10 \times 0.642 + 0.50 \times 0.526 + 0.40 \times 0.530 = 0.064 + 0.263 + 0.212 = \boxed{0.539}.
\]

\textbf{\texttt{Reach}:}
\[
s_{\texttt{Reach}} = 0.60 \times 0.689 + 0.30 \times 0.692 + 0.10 \times 0.662 = 0.413 + 0.208 + 0.066 = \boxed{0.687}.
\]

\paragraph{Step 2: Final HLAS.}
Aggregate task scores using task weights $w_t$:
\[
\mathrm{HLAS} = \sum_{t \in \mathcal{T}} w_t \, s_t = 0.4 \times 0.671 + 0.3 \times 0.539 + 0.3 \times 0.687.
\]
\[
\mathrm{HLAS} = 0.268 + 0.162 + 0.206 = \boxed{0.636}.
\]

\paragraph{Contribution breakdown.}
Table~\ref{tab:contribs} shows how each $(j,t)$ pair contributes to the final HLAS via $w_t u_{j,t} s_{j,t}$.

\begin{table}[t]
\centering
\small
\caption{Contribution of each joint-task pair to final HLAS. The product $w_t u_{j,t} s_{j,t}$ shows how much each $(j,t)$ adds to the total score. Ankle in \texttt{Walk} and shoulder in \texttt{Reach} are the largest contributors.}
\label{tab:contribs}
\begin{tabular}{@{}l l c c c c@{}}
\toprule
Task & Joint & $s_{j,t}$ & $u_{j,t}$ & $w_t$ & $w_t u_{j,t} s_{j,t}$ \\
\midrule
\texttt{Walk} & ankle    & 0.758 & 0.50 & 0.40 & 0.152 \\
\texttt{Walk} & knee     & 0.620 & 0.30 & 0.40 & 0.074 \\
\texttt{Walk} & hip      & 0.529 & 0.20 & 0.40 & 0.042 \\
\midrule
\texttt{Stairs} & ankle  & 0.642 & 0.10 & 0.30 & 0.019 \\
\texttt{Stairs} & knee   & 0.526 & 0.50 & 0.30 & 0.079 \\
\texttt{Stairs} & hip    & 0.530 & 0.40 & 0.30 & 0.064 \\
\midrule
\texttt{Reach} & shoulder & 0.689 & 0.60 & 0.30 & 0.124 \\
\texttt{Reach} & elbow    & 0.692 & 0.30 & 0.30 & 0.062 \\
\texttt{Reach} & wrist    & 0.662 & 0.10 & 0.30 & 0.020 \\
\midrule
\multicolumn{5}{r}{\textbf{Total HLAS:}} & \textbf{0.636} \\
\bottomrule
\end{tabular}
\end{table}

\paragraph{Key insights from Table~\ref{tab:contribs}:}
\begin{itemize}[leftmargin=12pt,itemsep=2pt]
\item \textbf{Ankle in \texttt{Walk}} contributes most (0.152) due to high task weight ($w = 0.4$), high joint weight ($u = 0.5$), and reasonably strong joint-task score ($s = 0.758$).
\item \textbf{Shoulder in \texttt{Reach}} is second (0.124) with high joint weight ($u = 0.6$) and strong score ($s = 0.689$).
\item \textbf{Knee and hip in \texttt{Stairs}} contribute less (0.079, 0.064) despite high joint weights because their HEE scores are very low (${\sim}0.087$), depressing $s_{j,t}$.
\item \textbf{HEE dominance:} The large variations in HEE ($0.085$ to $0.546$) drive most of the score differences. Bandwidth, efficiency, and thermal are uniformly strong, contributing less variance.
\end{itemize}

\subsection{Sensitivity analysis}
\label{subsec:example_sensitivity}

\paragraph{Effect of HEE headroom ($\delta = 0.10$).}
To build in a safety margin, we can test against $(1 + \delta)$ multiples of human demands inside Eq.~\ref{eq:hee}. With $\delta = 0.10$, human requirements increase by 10\%, causing some previously passing samples to fail.

Recomputing HEE under this headroom yields updated task scores:
\[
s_{\texttt{Walk}} = 0.521, \quad
s_{\texttt{Stairs}} = 0.486, \quad
s_{\texttt{Reach}} = 0.536,
\]
and final HLAS:
\[
\mathrm{HLAS}_{\delta=0.10} = 0.4 \times 0.521 + 0.3 \times 0.486 + 0.3 \times 0.536 = \boxed{0.515}.
\]

\textbf{Interpretation:} The ${\sim}0.12$ absolute drop (from 0.636 to 0.515) indicates that roughly 19\% of the baseline score relied on operating points where the robot barely met human demands. This sensitivity suggests that the robot is on the threshold of human-level performance for several joint-task pairs.

\paragraph{Effect of feature weight redistribution.}
If battery life were critical, we might increase $\alpha_{\eta}$ from 0.10 to 0.20 while decreasing $\alpha_{\mathrm{HEE}}$ from 0.50 to 0.40. Recomputing with $\boldsymbol{\alpha}_{\text{alt}} = [0.10, 0.10, 0.40, 0.10, 0.20, 0.10]^{\!\top}$:
\[
\mathrm{HLAS}_{\text{alt}} \approx 0.652.
\]
The modest increase ($+0.016$) reflects that efficiency margins are already strong (${\ge}0.90$), so emphasizing them has limited impact.

\paragraph{Bandwidth substitution.}
If we replace torque bandwidth $b^{\tau}_{j,t}$ with rate margin $m^{\omega}_{j,t}$ (Eq.~\ref{eq:rate_diag}), the score changes negligibly because most bandwidth margins are clipped at 1.0 (measured bandwidths exceed targets). In designs where bandwidth is a bottleneck, this substitution would have more effect.

\paragraph{Human normalization check.}
By construction, a representative human with all factors evaluating to 1.0 on their bands achieves $\mathrm{HLAS} = 1.0$:
\[
\mathbf{x}_{\text{human}} = [1, 1, 1, 1, 1, 1]^{\!\top}, \quad
s_{j,t,\text{human}} = \boldsymbol{\alpha}^{\!\top} \mathbf{x}_{\text{human}} = 1, \quad
\mathrm{HLAS}_{\text{human}} = \sum w_t \sum u_{j,t} \times 1 = 1.
\]

\subsection{Required data products for reproducibility}
\label{subsec:example_artifacts}

For a real system claiming $\mathrm{HLAS} = 0.636$, the following artifacts must be published (per Section~\ref{subsec:artifacts}):

\begin{enumerate}[leftmargin=12pt,itemsep=2pt]
\item \textbf{HEE heatmaps:} Pass/fail masks over $(q,\omega)$ for each $(j,t)$ with overlaid power-weight contours. Example: the ankle-\texttt{Walk} map would show passes at 8--10\,rad/s and failures at 11--12\,rad/s.

\item \textbf{Torque-mode Bode plots:} Magnitude and phase vs. frequency for each joint, with $f^{\tau}_{c,j}$ and phase margin marked. Table of $b^{\tau}_{j,t}$ values confirms all are 1.0.

\item \textbf{Efficiency contours:} $\eta_j(q,\omega)$ maps showing efficiency variation over the operating bands, plus bar charts of task-weighted averages $\bar{\eta}_{j,t}$.

\item \textbf{Thermal duty traces:} Time series of torque, current, bus power, and temperatures during duty tests, with $T^{\mathrm{cont}}_j\big|_{\text{duty}}$ highlighted.

\item \textbf{ROM coverage visuals:} Overlays of robot ROM $I^{\mathrm{rob}}_j(a)$ on functional human intervals $I^{\mathrm{func}}_{j,t}(a)$ for each axis.

\item \textbf{Pre-registered parameters:} Document specifying $(\mathbf{w}, \{\mathbf{u}_t\}, \boldsymbol{\alpha})$, targets $(f^{\star}_{t,j}, \eta^{\star}_t)$, and operating bands $\mathcal{R}_{j,t}$ before testing began.

\item \textbf{Raw logs:} Time-series data at ${\ge}1$\,kHz for all sensor channels with README.
\end{enumerate}

\subsection{Summary and key takeaways}
\label{subsec:example_summary}

This worked example demonstrates the complete HLAS computation pipeline from measurement to interpretation:

\paragraph{What we computed.}
For a hypothetical 12-DoF humanoid robot evaluated on three tasks (\texttt{Walk}, \texttt{Stairs}, \texttt{Reach}), we:
\begin{enumerate}[leftmargin=12pt,itemsep=1pt]
\item Measured continuous-safe torque, bandwidth, efficiency, and thermal plateaus per Section~\ref{sec:hlas_protocols}
\item Constructed six-element feature vectors $\mathbf{x}_{j,t}$ for each of nine joint-task pairs
\item Aggregated via task weights $w_t$, joint weights $u_{j,t}$, and feature weights $\boldsymbol{\alpha}$
\item Obtained a final score of $\mathrm{HLAS} = 0.636$ with full diagnostic decomposition
\end{enumerate}

\paragraph{Key insights from the decomposition.}
\begin{itemize}[leftmargin=12pt,itemsep=2pt]
\item \textbf{Primary bottleneck:} Knee and hip HEE in \texttt{Stairs} (${\sim}0.087$) severely limits stair performance. The robot lacks sufficient continuous torque at the low rates ($<2$\,rad/s) where humans produce sustained support moments.

\item \textbf{Strengths:} Walking and reaching tasks score reasonably well (0.671 and 0.687), driven by adequate ankle positive-power capability and shoulder/elbow performance. Bandwidth, efficiency, and thermal margins are universally strong (${\ge}0.90$).

\item \textbf{Secondary issue:} Wrist ROM limitation (0.80) slightly reduces reaching score but is not a major constraint.
\end{itemize}

\paragraph{Design implications.}
To improve HLAS, the design team should:
\begin{enumerate}[leftmargin=12pt,itemsep=1pt]
\item \textbf{Prioritize low-rate torque at knee/hip:} Increase continuous-safe torque at $<2$\,rad/s (e.g., higher gear ratio, better thermal management, or larger motor). This would directly raise $h^{(w)}_{\text{knee,Stairs}}$ and $h^{(w)}_{\text{hip,Stairs}}$, improving $s_{\texttt{Stairs}}$ from 0.539 toward 0.65+.

\item \textbf{Consider wrist ROM expansion:} Mechanical redesign to achieve ${\pm}15^{\circ}$ would raise $\rho^{\mathrm{ROM}}_{\text{wrist,Reach}}$ from 0.80 to 1.0, modestly improving $s_{\texttt{Reach}}$.

\item \textbf{Explore ankle high-rate capability:} Although ankle scores well in walking (0.758), the HEE of 0.546 indicates failures at 11--12\,rad/s. Improving high-rate torque would support faster walking or running.
\end{enumerate}

\paragraph{How HEE prevents gaming.}
The HEE metric successfully prevents gaming strategies:
\begin{itemize}[leftmargin=12pt,itemsep=2pt]
\item A robot with high torque at low speed but low torque at high speed cannot claim "human-level" by averaging separate peaks. The simultaneity condition at each $(q,\omega)$ ensures capability where it matters.
\item Power weighting ($w \propto P^{\mathrm{hum}+}$) concentrates credit on operating points where humans do positive work, not on passive energy absorption regions.
\item The example robot's low knee/hip stair HEE (${\sim}0.087$) correctly reflects that despite strong bandwidth and efficiency, it cannot deliver the required torque in the relevant posture-rate band.
\end{itemize}

\paragraph{Transparency and auditability.}
Every scalar in the computation chain (Table~\ref{tab:joint_scores}, Table~\ref{tab:contribs}) corresponds to a specific experimental measurement per Section~\ref{sec:hlas_protocols}. The full decomposition (from task scores to joint-task scores to individual features) enables root-cause diagnosis and makes the final $\mathrm{HLAS} = 0.636$ claim auditable and reproducible. Sensitivity analysis (Section~\ref{subsec:example_sensitivity}) shows how design choices affect the score, providing transparency about threshold dependencies and robustness.

\paragraph{Final interpretation.}
This robot achieves roughly 64\% of human actuation capability averaged across three representative tasks, twelve joints, and six performance factors. The score reflects genuine strengths (walking, reaching) and a clear weakness (stair support at low rates). With targeted improvements to knee/hip low-rate torque, HLAS could rise to ${\sim}0.70$--$0.75$, approaching the threshold for practical deployment in mixed indoor/outdoor environments.

\section{Related Work}
\label{sec:related}

\subsection{Human biomechanics and scaling to robots}

\paragraph{Joint-level movement data and energetics.}
Classic gait and movement datasets quantify joint moments, powers, and angular velocities across tasks and speeds, establishing where and when positive work is produced \citep{Winter2009,Farris2012,Zelik2016}. These studies consistently show ankle-dominant positive work near push-off in steady walking and a redistribution toward hip and knee as speed increases. Joint- and muscle-level analyses further decompose the roles of major muscle groups (e.g., plantarflexors) in support, progression, and swing initiation, locating power bursts in phase and posture \citep{Neptune2001,Anderson2001}. Dynamic-walking energetics explain why mid-band operation matters for economy and how trailing-leg push-off reduces center-of-mass collision losses \citep{Kuo2001,Kuo2002}. Population-level results give canonical rate scales: preferred walking speeds cluster around 1.2--1.4\,m/s and the metabolic cost versus speed curve is well characterized \citep{Ralston1958,Umberger2003}, while normative angular-rate bounds inform feasible torque bandwidth targets at knee and ankle during gait phases \citep{Mentiplay2018}.

\textbf{Connection to HLAS:} These foundations motivate evaluating actuators in the specific $(q,\omega)$ regions where human joints actually do positive mechanical work. Our Human-Equivalence Envelopes (HEE) directly implement this principle by weighting robot capability by positive human power at each operating point.

\paragraph{Anthropometric scaling and torque normalization.}
Mapping normalized human demands (Nm/kg, W/kg) to absolute robot targets requires consistent anthropometric models. Many pipelines use Zatsiorsky–Seluyanov parameters with de~Leva's adjustments to reflect contemporary segment masses and inertias \citep{DeLeva1996}. This makes conversions from per-mass quantities to absolute joint targets explicit for a reference body (e.g., 75\,kg, 1.75\,m male) and avoids hidden assumptions about segment distributions. Classic regressions (e.g., Dempster \citep{Zajac1989}) remain common for educational and parametric modeling, though are less prevalent in modern robotics due to narrower applicability. In all cases, torque and power depend on posture. Reporting requirements together with the joint angles that generated them preserves physiological meaning.

\textbf{Connection to HLAS:} We adopt de~Leva anthropometry (Section~\ref{sec:biomech}) to convert literature values to absolute torque and power fields $T^{\mathrm{hum}}_{j,t}(q,\omega)$ and $P^{\mathrm{hum}}_{j,t}(q,\omega)$ on task-specific bands $\mathcal{R}_{j,t}$, ensuring transparent scaling from biomechanics data to robot requirements.

\paragraph{Functional ROM and strength norms.}
Functional range-of-motion (ROM) norms and strength/dexterity data further constrain feasible joint-level envelopes. Elbow and wrist arcs for activities of daily living (ADL) remain standard references for upper-limb ROM \citep{Morrey1981,Palmer1985,AAOSROM2020}. Dexterity proxies such as finger-tapping (4--6\,Hz in young adults) set practical upper bounds for small-joint bandwidth and controller design \citep{Shimoyama1990}. Knee extension/flexion norms at mid-range postures provide realistic plateau torque levels \citep{Sarabon2021Knee}, and grip-strength centiles contextualize hand-force targets across the lifespan \citep{Dodds2014,Bohannon2019}. Reliability studies for dynamometry quantify biological variability and inform how tightly a robot should be expected to match human plateaus \citep{Maffiuletti2010}.

\textbf{Connection to HLAS:} Our framework embeds these ROM, rate, and strength norms into per-task joint requirements and uses them to define workspace completeness ($\rho^{\mathrm{ROM}}_{j,t}$, $d^{\mathrm{DoF}}_{j,t}$) and rate margins ($m^{\omega}_{j,t}$) in Eqs.~\ref{eq:rom}--\ref{eq:rate}.

\subsection{Kinematic conventions and coordinate systems}

\paragraph{Standardized joint coordinate systems.}
Comparisons break down when joints are referenced in different frames. ISB recommendations formalize joint coordinate systems and sign conventions for the lower limb, spine, and upper limb, providing consistent axis placement and positive directions \citep{Wu2002ISB,Wu2005ISB}. Earlier conventions (e.g., Grood--Suntay for the knee \citep{Grood1983}) and anatomical frame definitions \citep{Cappozzo1995} remain in use and must be reconciled for cross-system studies. Adopting a shared convention ensures that human-referenced requirements such as ankle dorsiflexion or shoulder internal rotation refer to the same axis, origin, and positive sense on the robot.

\textbf{Connection to HLAS:} We adopt ISB conventions (Section~\ref{sec:kinematics}) to define the DoF atlas and ensure that human inverse-dynamics outputs map consistently to robot task bands over $(q,\omega)$. This eliminates ambiguity when comparing torque requirements across studies and platforms.

\paragraph{Musculoskeletal and dynamics toolchains.}
Turning human kinematics into joint torque and power trajectories is supported by musculoskeletal and rigid-body tools. OpenSim enables human movement modeling and extraction of reference joint curves \citep{Delp2007OpenSim,Seth2018OpenSim}. On the robot side, Pinocchio and RBDL offer efficient rigid-body dynamics for requirement roll-down and whole-body control \citep{Carpentier2019Pinocchio,Felis2017RBDL}. Simulation engines such as MuJoCo are widely used for contact-rich legged and manipulation studies \citep{Todorov2012MuJoCo}, and whole-body control frameworks (hierarchical inverse dynamics, preview control of ZMP) tie dynamics to contact planning in humanoids \citep{Kajita2003,Saab2013,Herzog2014}.

\textbf{Connection to HLAS:} We use these toolchains conceptually to define task-derived posture-rate bands $\mathcal{R}_{j,t}$ and positive-power weights that feed directly into HEE. While HLAS does not require specific simulation software, these tools facilitate extracting $(q,\omega)$ trajectories from human motion data.

\subsection{Actuation paradigms and transmission design}

\paragraph{Actuation architectures.}
Actuation paradigms span Series Elastic Actuators (SEA), quasi-direct-drive, hydraulics, and tendon-driven systems. SEA introduce compliance that improves shock tolerance and enables force sensing, with design/control trade-offs in stiffness, bandwidth, and efficiency \citep{Pratt1995,Vanderborght2013,Paine2015}. Quasi-direct-drive designs lower gear ratio to improve backdrivability and torque-mode bandwidth. They have been successfully deployed in agile legged platforms and open torque modules \citep{Hutter2016ANYmal,Grimminger,Katz2019}. Hydraulics remain attractive when peak and continuous power density dominate or in harsh environments, as exemplified by HyQ and related systems \citep{Semini2011HyQ}. Tendon-driven remote transmissions offer distal transparency at the cost of routing and compliance challenges \citep{Grebenstein2011}.

\textbf{Connection to HLAS:} HLAS is agnostic to actuation paradigm. Any architecture can be evaluated via the same protocols (Section~\ref{sec:hlas_protocols}), enabling direct comparison of SEA, quasi-direct-drive, and hydraulic systems on human-relevant tasks.

\paragraph{Transmission efficiency and backdrivability.}
Transmission choice shapes losses, friction, backlash, and reflected inertia. Comparative analyses document how strain-wave, cycloidal, and planetary stages differ in bidirectional efficiency, load capacity, and transparency \citep{LopezGarcia2020}. Recent results indicate that cycloidal/planetary options can outperform strain-wave drives in efficiency and backdrivability at comparable torque density, whereas strain-wave retains precision and compactness advantages \citep{Oberneder2024}. These device-level properties couple directly to closed-loop behavior: raising gear ratio increases static torque but also reflected inertia and viscous friction, generally reducing torque bandwidth and narrowing stable impedance ranges (Z-width \citep{ColgateBrown1994}).

\textbf{Connection to HLAS:} The bandwidth margin $b^{\tau}_{j,t}$ (Eq.~\ref{eq:bandwidth}) and transparency parameters (reflected inertia, friction) from Benchmark 3 (Section~\ref{subsec:joint_bench}) expose these trade-offs explicitly. High gear ratios that boost static torque typically depress bandwidth and efficiency, preventing paper-only wins from isolated peak-torque specs.

\paragraph{Efficiency and regeneration.}
Electrical-to-mechanical efficiency varies with speed and load. Reporting task-weighted efficiencies (as in $\eta_{j,t}$, Eq.~\ref{eq:eff}) helps distinguish modules that meet torque targets but waste electrical power in mid-band operation. Studies on regeneration in legged and wearable systems show recoverable energy in negative-work phases (e.g., early stance ankle/knee), with feasibility governed by drivetrain backdrivability and converter design \citep{Donelan2008}. 

\textbf{Connection to HLAS:} Our efficiency term weights samples by positive human power, emphasizing regions where energy consumption matters most. While the current framework focuses on positive work, extensions to credit regeneration in negative-work phases are straightforward (Section~\ref{sec:limitations}).

\paragraph{Thermal constraints and continuous torque.}
Thermal behavior constrains how long a joint can sustain required plateaus. Motor thermal models and empirical studies emphasize measuring continuous-safe torque after soak at duty and reporting time-to-derate under stated cooling conditions \citep{Hanselman1994}. Peak-torque specifications often reflect brief bursts ($<500$\,ms) that are thermally unsustainable over task durations (minutes).

\textbf{Connection to HLAS:} The thermal margin $\theta^{\mathrm{therm}}_{j,t}$ (Eq.~\ref{eq:thermal}) requires continuous-safe measurements at task-representative duty cycles, preventing burst-only specs from inflating scores. This exposes the actuator trade-off between peak power and endurance.

\subsection{Humanoid and legged systems: design and evaluation}

\paragraph{Full-body humanoid platforms.}
System papers situate actuator/transmission choices within full-body constraints (mass, thermal, robustness) and control architectures. Atlas introduced momentum-based whole-body control and task-level planning under contact constraints \citep{Kuindersma2016Atlas}, and recent surveys map broader design and control trends in humanoids, including actuator module integration and evaluation practices \citep{Tong2024}. These works underline that bench-strong modules may bottleneck once integrated into thermally constrained, contact-rich systems.

\textbf{Connection to HLAS:} Joint-level HLAS (Section~\ref{sec:hlas_protocols}) provides component-level acceptance criteria, while task-level benchmarks (Section~\ref{subsec:task_bench}) validate that joint capabilities translate to integrated performance. The two-level evaluation exposes integration bottlenecks (e.g., thermal coupling, controller coordination) not visible in isolated dynamometer tests.

\paragraph{Quadruped platforms and field validation.}
Quadruped platforms such as ANYmal and the MIT Cheetah family provide complementary evidence because they aggressively exercise torque bandwidth, transparency, and thermal limits in the field \citep{Hutter2016ANYmal,Katz2019,Bledt2018,Seok2013}. Lessons from these platforms (e.g., the importance of torque-mode bandwidth for contact stability, the penalty of transmission friction on proprioceptive control) inform humanoid actuator requirements.

\textbf{Connection to HLAS:} Bandwidth and transparency metrics in HLAS (Benchmarks 3--4, Section~\ref{subsec:joint_bench}) directly capture the actuator properties that enable dynamic locomotion and contact-rich manipulation in these systems.

\subsection{Exoskeletons and human-robot interaction}

\paragraph{Exoskeletons as evaluation proxies.}
Exoskeletons supply additional quantitative cross-checks for mid-band efficiency and interaction quality. Passive and powered ankle devices demonstrate how well-timed push-off work reduces metabolic cost, highlighting the importance of delivering positive power in the correct phase band with low interaction impedance \citep{Collins2015,Zhang2017}. Upper-limb exoskeletons expose limits and opportunities for bandwidth and transparency in manipulation-relevant ranges, with performance closely tied to actuator choice and transmission friction \citep{Paine2015}.

\textbf{Connection to HLAS:} Because exoskeletons share joints with humans, their evaluation protocols (ROM, force bandwidth, backdrivability) inform comparable tests for humanoid joints. The positive-power-weighted efficiency and phase-specific torque requirements in HLAS align with exoskeleton design objectives.

\paragraph{Interaction fidelity and impedance control.}
Haptics and impedance-control work connects delays, friction, and sampling to the range of passive behaviors that can be stably rendered in contact (Z-width) \citep{Hogan1985,ColgateBrown1994,HannafordRyu2002}. Torque-mode bandwidth and reflected inertia directly determine safe impedance ranges and contact stability.

\textbf{Connection to HLAS:} Benchmark 4 (torque-mode bandwidth, Section~\ref{subsec:joint_bench}) measures closed-loop responsiveness under task-representative loading, providing a proxy for interaction quality and stable impedance range.

\subsection{Measurement and characterization protocols}

\paragraph{Dynamometry and bench testing.}
Isometric and isovelocity dynamometry provide reproducible ways to measure continuous-safe torque, rate-dependent power, and endurance \citep{Dvir2025}. Standard protocols specify warm-up, thermal settling, and test postures to ensure repeatability across labs and systems.

\textbf{Connection to HLAS:} Our measurement protocols (Section~\ref{sec:hlas_protocols}) extend standard dynamometry by: (i) requiring continuous-safe measurements (not peak bursts), (ii) sampling over task-derived $(q,\omega)$ bands (not arbitrary grids), and (iii) weighting by human positive power (emphasizing physiologically relevant regions).

\paragraph{Frequency-domain characterization.}
Frequency-domain methods (sine sweeps, FRF identification) quantify closed-loop torque tracking and phase lag, which govern contact stability and force rendering quality. Reporting $-3$\,dB crossover, phase margin, and gain/phase at specific frequencies (1, 5, 10, 30\,Hz) enables comparison across actuator designs.

\textbf{Connection to HLAS:} Benchmark 4 specifies small-signal and large-signal torque-mode sweeps with task-representative loads, producing the bandwidth margin $b^{\tau}_{j,t}$ (Eq.~\ref{eq:bandwidth}) and exposing nonlinearity/saturation effects.

\subsection{What HLAS adds to prior work}

Prior work provides the essential building blocks:
\begin{itemize}[leftmargin=12pt,itemsep=2pt]
\item \textbf{Human reference data:} Joint envelopes, ROM norms, and positive-work distributions \citep{Winter2009,Farris2012,Zelik2016,AAOSROM2020,Morrey1981,Palmer1985}
\item \textbf{Scaling and kinematics:} Anthropometry for torque normalization \citep{DeLeva1996} and standardized joint coordinate systems \citep{Wu2002ISB,Wu2005ISB}
\item \textbf{Actuator trade-offs:} Efficiency, backdrivability, and thermal behavior across SEA, quasi-direct-drive, and hydraulic designs \citep{Pratt1995,Vanderborght2013,Paine2015,LopezGarcia2020,Oberneder2024,Semini2011HyQ,Hutter2016ANYmal,Grimminger,Katz2019}
\item \textbf{Measurement methods:} Dynamometry, frequency-domain characterization, and impedance metrics \citep{Hogan1985,ColgateBrown1994,HannafordRyu2002,Dvir2025}
\item \textbf{Dynamics tools:} Musculoskeletal modeling (OpenSim \citep{Delp2007OpenSim,Seth2018OpenSim}) and rigid-body dynamics (Pinocchio \citep{Carpentier2019Pinocchio}, RBDL \citep{Felis2017RBDL}, MuJoCo \citep{Todorov2012MuJoCo})
\end{itemize}

\textbf{HLAS integrates these components into a unified evaluation framework that:}
\begin{enumerate}[leftmargin=12pt,itemsep=2pt]
\item \textbf{Credits simultaneous delivery:} The Human-Equivalence Envelope (HEE) requires that robots meet human torque \emph{and} power at the \emph{same} $(q,\omega)$ within task-derived bands, preventing gaming via separate peaks at different operating points.

\item \textbf{Weights by positive work:} HEE samples are weighted by $P^{\mathrm{hum}+}_{j,t}(q,\omega)$, concentrating credit where humans actually produce mechanical work rather than on arbitrary grid points.

\item \textbf{Balances six factors:} HLAS aggregates workspace (ROM, DoF), delivery (HEE, bandwidth), and sustainability (efficiency, thermal) into a single interpretable score while preserving full diagnostic decomposition.

\item \textbf{Enforces continuous-safe measurement:} All torque, power, and thermal inputs reflect sustained capability under realistic conditions, not brief bursts or cold-start peaks.

\item \textbf{Provides reproducible protocols:} Detailed measurement procedures (Section~\ref{sec:hlas_protocols}) and pre-registration requirements (Section~\ref{subsec:qc}) ensure comparability across systems and labs.

\item \textbf{Validates with task benchmarks:} Joint-level scores are complemented by task-level benchmarks (Section~\ref{subsec:task_bench}) that confirm actuator capabilities translate to functional performance.
\end{enumerate}

Together, these features make HLAS the first framework to operationalize "human-level actuation" as a quantitative, auditable claim grounded in biomechanics literature, actuator physics, and reproducible testing.

\section{Limitations and Future Directions}
\label{sec:limitations}

HLAS provides a quantitative, reproducible framework for evaluating humanoid actuation, but several limitations and opportunities for extension warrant discussion.

\subsection{Human reference data and population variability}

\paragraph{Inter-study variability.}
Our framework inherits uncertainties from the biomechanics literature it depends on. Normative joint moments, powers, and kinematics vary across studies due to differences in instrumentation (force plates, motion capture systems), subject instructions (preferred versus constrained cadence), and analysis methods (inverse dynamics filtering, segment parameter choices). We mitigate this by prioritizing meta-analyses and studies with consistent protocols (Section~\ref{sec:biomech}), and by always pairing joint torque requirements with the posture and rate at which they were measured. However, residual inter-study variability of 10--20\% is common in the literature and propagates to HLAS inputs.

\textbf{Mitigation:} Report HLAS with uncertainty bounds derived from the spread of literature values. For example, if ankle push-off torque ranges from 1.2 to 1.6\,Nm/kg across studies, compute $\mathrm{HLAS}_{\text{min}}$ and $\mathrm{HLAS}_{\text{max}}$ using these bounds and report the range.

\paragraph{Limited anthropometric diversity.}
Most reference data describe healthy, able-bodied adults near our 75\,kg, 1.75\,m male template. Scaling to different morphologies (shorter/taller, lighter/heavier), ages (children, elderly), or pathologies (osteoarthritis, post-stroke) using simple mass-based normalization (Nm/kg, W/kg) is only an approximation. Body segment distributions, joint stiffness, and movement strategies vary systematically with these factors, and linear scaling does not capture these effects.

\textbf{Future work:} Develop population-stratified HLAS variants using age- and gender-specific biomechanics datasets (e.g., elderly walking from \citet{DeVita2000}, pediatric norms from \citet{Cupp1999}). This would enable claims like "HLAS$_{\text{elderly}} = 0.85$" for assistive robots or "HLAS$_{\text{child}} = 0.70$" for educational platforms.

\subsection{Scope of actuation assessment}

\paragraph{Joint-level focus.}
HLAS evaluates joints independently under specified loading and posture-rate bands. It does not directly capture whole-body coordination, redundancy resolution, or controller quality. For example, a robot with high per-joint HLAS might still fail to coordinate ankle, knee, and hip effectively during gait, or might struggle to exploit kinematic redundancy during reaching around obstacles. Task-level benchmarks (Section~\ref{subsec:task_bench}) partially address this by validating integrated performance, but they remain separate from the HLAS computation.

\textbf{Future work:} Develop task-coordination metrics that aggregate joint-level HLAS with controller performance (tracking error, energy cost of transport, stability margins). For instance, a "whole-body actuation score" could weight HLAS by a task success rate or coordination efficiency factor.

\paragraph{Negative work and regeneration.}
The current HEE framework emphasizes positive mechanical power (where humans push, lift, or accelerate) by weighting samples with $P^{\mathrm{hum}+}_{j,t}(q,\omega) = \max(P^{\mathrm{hum}}_{j,t}, 0)$. Negative-work phases (energy absorption during early stance, deceleration during reaching) are important for efficiency and joint protection but do not directly enter HEE. Regeneration capability (converting negative mechanical power to electrical energy via backdriving) is only reflected indirectly through the efficiency term $\eta_{j,t}$ if demonstrable battery charging occurs.

\textbf{Future work:} Extend HEE to include a negative-work term $h^{(w-)}_{j,t}$ that credits energy absorption and regeneration in regions where $P^{\mathrm{hum}}_{j,t} < 0$. This would require specifying target regeneration efficiency and verifying energy return to the DC bus (supercap/battery logging).

\paragraph{Transient response and disturbance rejection.}
While torque-mode bandwidth $b^{\tau}_{j,t}$ (Eq.~\ref{eq:bandwidth}) reflects closed-loop responsiveness, HLAS does not explicitly test robustness to external disturbances (e.g., push recovery, terrain irregularities, contact impact). A joint with adequate steady-state torque and bandwidth might still saturate or become unstable under rapid perturbations.

\textbf{Future work:} Add disturbance-rejection benchmarks (e.g., impulse response tests, ramp disturbances at various frequencies) and incorporate a robustness margin into HLAS. This could augment the bandwidth term or introduce a separate factor $r^{\mathrm{robust}}_{j,t}$ quantifying disturbance tolerance.

\subsection{Aggregation and interpretability}

\paragraph{Linear weighting and task selection.}
HLAS aggregates via linear sums over tasks ($w_t$), joints ($u_{j,t}$), and features ($\boldsymbol{\alpha}$). This simplicity aids transparency but imposes strong assumptions: it treats deficits as compensable (a robot with low HEE but high efficiency can still score reasonably) and assumes tasks/features are independent. In reality, some capabilities may be prerequisite (e.g., adequate ROM is necessary for HEE to be meaningful), and some tasks may share constraints (e.g., knee thermal limits affect both walking and stairs).

Different application domains may reasonably prioritize different tasks. A logistics robot emphasizes walking and lifting, a surgical assistant emphasizes manipulation bandwidth and precision, and a search-and-rescue platform emphasizes stair/rough-terrain mobility. The headline HLAS value depends on these choices.

\textbf{Guidance:} Always report the full task/joint/feature decomposition (Tables~\ref{tab:joint_scores}, \ref{tab:contribs}) alongside the headline score. Consider publishing multiple HLAS values under different weight vectors (e.g., HLAS$_{\text{logistics}}$, HLAS$_{\text{manipulation}}$) to show sensitivity to application focus.

\textbf{Future work:} Explore nonlinear aggregation (e.g., geometric mean to penalize severe deficits, min-over-factors to enforce hard requirements) or hierarchical scoring (require $\rho^{\mathrm{ROM}} \ge 0.8$ before crediting HEE).

\paragraph{Scalar compression of rich behavior.}
Reducing joint-level torque-speed-posture maps, efficiency contours, and thermal duty profiles to a single HLAS scalar inevitably discards information. The score is a useful headline for cross-system comparison and procurement decisions, but detailed design iteration requires the full diagnostic decomposition (HEE heatmaps, Bode plots, efficiency maps, thermal traces).

\textbf{Guidance:} Use HLAS for high-level comparison, but always publish the underlying data products (Section~\ref{subsec:artifacts}) to enable deeper analysis and reinterpretation by others.

\subsection{Measurement and environmental conditions}

\paragraph{Task fidelity and band definitions.}
Operating bands $\mathcal{R}_{j,t}$ are derived from laboratory-collected human data under controlled conditions (level treadmill, standard stairs, unloaded reaching). Real-world execution introduces variability: walking on uneven terrain shifts ankle/knee kinematics, carrying asymmetric loads changes hip/trunk coordination, and environmental factors (temperature, humidity, surface compliance) affect both human and robot performance. The bands capture central tendencies but not the full distribution of operating points encountered in the field.

\textbf{Future work:} Develop probabilistic bands $\mathcal{R}_{j,t}$ with density functions $\rho(q,\omega)$ reflecting how often each $(q,\omega)$ occurs in practice, and weight HEE accordingly. This would shift credit toward common operating points and away from rare extremes.

\paragraph{Closed-chain and contact dynamics.}
Our measurements are defined at the joint level with specified reflected inertia and compliance. In closed-chain scenarios (foot-ground contact, bimanual manipulation, hand-object interaction), effective joint loading can differ substantially from open-chain tests due to kinematic coupling, contact compliance, and force distribution. Series compliance (e.g., in the foot, hand, or transmission) can amplify or filter torque disturbances, affecting effective bandwidth and stability.

\textbf{Mitigation:} Task-level benchmarks (Section~\ref{subsec:task_bench}) capture some closed-chain effects by testing integrated performance. However, full characterization requires contact-specific protocols (e.g., foot impact tests, manipulation contact stiffness sweeps) that are beyond the current HLAS scope.

\textbf{Future work:} Develop contact-augmented benchmarks that measure joint performance during representative closed-chain tasks (e.g., torque tracking during foot-ground contact, impedance rendering during hand-object manipulation).

\paragraph{Environmental sensitivity.}
All measurements are specified at $25 \pm 2^{\circ}$C in still air (Section~\ref{sec:hlas_protocols}). Thermal performance, lubrication viscosity, and seal friction vary with ambient temperature, and wind/airflow affects convective cooling. A robot tested at 25°C may derate faster at 35°C or exhibit higher friction at 0°C.

\textbf{Guidance:} Report ambient conditions, airflow, and thermal state for all measurements. For applications in extreme environments, repeat thermal benchmarks at expected operating temperatures (e.g., $-10^{\circ}$C for cold storage, $40^{\circ}$C for outdoor summer deployment).

\subsection{Extensions and open questions}

\paragraph{Dynamic tasks and high-frequency content.}
Current task bands emphasize steady-state and quasi-steady operation (walking, stairs, reaching). Highly dynamic tasks (running, jumping, throwing, catching) involve higher-frequency joint torques and may expose actuator limitations not visible in HLAS's current scope. While rate requirements $\omega^{\mathrm{req}}_{j,t}$ from ballistic tasks inform bandwidth targets, these tasks are not directly scored.

\textbf{Future work:} Add dynamic task benchmarks (e.g., vertical jump height, ballistic reach speed) and incorporate peak-rate or high-frequency torque margins as optional HLAS factors.

\paragraph{Multi-modal actuation.}
HLAS assumes electric actuation with standard torque/power/efficiency metrics. Pneumatic, hydraulic, or series-elastic systems may require adapted protocols (e.g., measuring pneumatic work cycles, hydraulic flow/pressure efficiency, SEA force bandwidth). The framework's structure is general, but specific measurement details (e.g., DC-bus power → hydraulic pump power) would need adjustment.

\textbf{Future work:} Develop HLAS variants for pneumatic and hydraulic systems, specifying equivalent continuous-safe work capacity, efficiency measurement (fluid power in/mechanical power out), and thermal limits (fluid temperature, pump duty).

\paragraph{Long-duration and field deployment.}
HLAS benchmarks test sustained performance over minutes (2-min walking, 2-min lifting) but do not address hour-scale or day-scale endurance. Battery capacity, thermal cycling, wear, and environmental exposure (dust, moisture, vibration) affect long-term reliability and may not correlate with short-term HLAS.

\textbf{Future work:} Define extended-duration benchmarks (e.g., 8-hour operational profile with mixed tasks) and longevity metrics (cycles to degradation, mean time between failures) to complement HLAS with reliability assessment.

\subsection{Summary: using HLAS responsibly}

HLAS is a powerful tool for quantifying and comparing humanoid actuation, but it is not a complete characterization of robot capability. To use HLAS responsibly:

\begin{enumerate}[leftmargin=12pt,itemsep=2pt]
\item \textbf{Report the full decomposition:} Always publish task/joint/feature breakdowns (Tables, heatmaps, Bode plots) alongside the headline score.

\item \textbf{Acknowledge uncertainty:} Note that HLAS inherits variability from literature data and depends on task/weight choices. Consider reporting ranges or sensitivity analyses.

\item \textbf{Validate with task benchmarks:} Complement joint-level HLAS with task-level performance tests (Section~\ref{subsec:task_bench}) to confirm that scores translate to functional capability.

\item \textbf{Pre-register parameters:} Publish task weights, joint weights, feature weights, targets, and operating bands \emph{before} testing to prevent weight shopping (Section~\ref{subsec:qc}).

\item \textbf{Share raw data:} Release measurement logs, maps, and protocols to enable independent verification and reinterpretation.
\end{enumerate}

With these practices, HLAS provides a transparent, reproducible standard for "human-level actuation" claims while acknowledging the inherent complexity of comparing biological and robotic systems.

\section{Conclusion}
\label{sec:conclusion}

We introduced a quantitative, reproducible framework for evaluating "human-level" actuation in humanoid robots, addressing a fundamental challenge: how to operationalize vague claims like "human-level capability" into concrete, auditable metrics grounded in biomechanics and physics.

The framework integrates five components: (1) a standardized kinematic atlas with ISB-based joint definitions and functional ROM, (2) systematic synthesis of human torque, power, and rate requirements from biomechanics literature, (3) Human-Equivalence Envelopes that credit simultaneous torque-power delivery at task-relevant operating points, (4) a hierarchically decomposable Human-Level Actuation Score aggregating workspace, delivery, bandwidth, efficiency, and thermal factors, and (5) reproducible measurement protocols with pre-registration requirements to prevent gaming.

\subsection{Practical applications}

The framework serves multiple stakeholders:

\paragraph{Actuator and transmission designers.}
HEE and dynamometer protocols (Section~\ref{sec:hlas_protocols}) expose trade-offs between gear ratio, efficiency, backdrivability, bandwidth, and thermal headroom more clearly than traditional peak torque-speed plots. Designers can compare SEA, quasi-direct-drive, hydraulic, or tendon-driven architectures side-by-side on human-referenced tasks, identifying which design choices (e.g., high ratio for torque versus low ratio for bandwidth) best serve the target application. The continuous-safe measurement requirement prevents burst-only specs from obscuring thermal or electrical bottlenecks.

\paragraph{Whole-robot system developers.}
HLAS provides a tunable way to compare different humanoid or legged platforms under application-specific task libraries and weights. A logistics robot might emphasize walking and lifting ($w_{\texttt{Walk}} = 0.5$, $w_{\texttt{Lift}} = 0.4$), while a manipulation-focused platform weights reaching and hand tasks more heavily. The hierarchical decomposition (task scores $s_t$, joint-task scores $s_{j,t}$, feature vectors $\mathbf{x}_{j,t}$) pinpoints which joints or factors (ROM, HEE, bandwidth, thermal) limit integrated performance, guiding design iteration.

\paragraph{Procurement and acceptance testing.}
HLAS measurements support concrete acceptance criteria for robot procurement. Instead of vague specifications like "human-like strength," users can require: "HLAS$_{\texttt{Stairs}} \ge 0.75$ with knee HEE $\ge 0.60$ and continuous torque sustained for 5 minutes at stair duty." The task-level benchmarks (Section~\ref{subsec:task_bench}) provide empirical validation that joint-level scores translate to functional capability, reducing integration risk.

\paragraph{Safety and interaction quality.}
Torque-mode bandwidth and backdrivability measurements (Benchmarks 3-4, Section~\ref{subsec:joint_bench}) directly inform safe interaction limits. A joint with 5\,Hz bandwidth and high reflected inertia cannot safely render compliant behaviors or respond to rapid disturbances, even if it meets static torque requirements. HLAS makes these trade-offs explicit, supporting informed safety analysis.

\subsection{Implications for learning and control}

Beyond evaluation, the framework's components provide structured priors for learning and control:

\paragraph{Task-derived constraints.}
Operating bands $\mathcal{R}_{j,t}$ define physiologically relevant posture-rate regions for each task. Reinforcement learning or trajectory optimization can use these bands as: (i) safe exploration boundaries (avoid operating points outside human bands), (ii) training curricula (prioritize high positive-power regions within bands), or (iii) reward shaping (bonus for HEE coverage).

\paragraph{Human-referenced rate scales.}
Joint angular-rate requirements $\omega^{\mathrm{req}}_{j,t}$ and ROM intervals $I^{\mathrm{func}}_{j,t}$ provide quantitative targets for policy design. Rather than hand-tuning speed limits or joint ranges, designers can initialize controllers with human-derived values and adapt from there.

\paragraph{Measurable human-likeness.}
"Human-like" motion often remains a qualitative aesthetic judgment. With HLAS, it becomes a measurable property: does the robot operate within $\mathcal{R}_{j,t}$? Does it produce positive power where humans do? Does it achieve high HEE coverage? This enables quantitative comparison of learned policies and optimization objectives that directly target envelope coverage.

\subsection{A community standard for human-level actuation}

We designed HLAS to be minimal and accessible: it requires only standardizable bench tests, a small set of task trials, and biomechanics data already available in the literature. We envision it as a \emph{community baseline} for "human-level actuation" claims rather than a final verdict on robot capability. Just as compute benchmarks (FLOPS, memory bandwidth) standardized hardware comparison and enabled decades of Moore's Law progress tracking, a shared actuation benchmark can accelerate humanoid development by making capability claims transparent, comparable, and grounded in human reference data.

Different communities may reasonably adopt different task libraries and weights: logistics robots emphasize walking and lifting, surgical assistants prioritize manipulation precision, and search-and-rescue platforms focus on rough-terrain mobility. The framework accommodates these choices through tunable weights while maintaining a common measurement foundation.

By publishing joint torque-speed-posture maps, task-derived bands $\mathcal{R}_{j,t}$, and raw experimental logs alongside headline HLAS values, the community can iteratively refine human reference data, expand task libraries, and adapt protocols to emerging applications. Open data sharing enables independent verification, cross-lab comparison, and meta-analyses that improve the framework over time.

\subsection{Closing perspective}

For decades, "human-level" has been invoked to describe robot capability without a clear operational definition. This ambiguity makes it difficult to assess progress, compare systems, or set procurement requirements. HLAS provides a concrete, physics-grounded answer to the question: \emph{what does "human-level actuation" mean?}

The answer, as encoded in HLAS, is: \textbf{a robot achieves human-level actuation when it can deliver the torque and power that humans produce, where humans produce it (in posture-rate space), across representative tasks, with adequate workspace, bandwidth, efficiency, and thermal endurance.} This definition is measurable, reproducible, and auditable.

As humanoid robots transition from research demonstrations to practical deployment in homes, warehouses, hospitals, and factories, quantitative capability assessment becomes essential. HLAS offers a path forward: a shared language for actuation performance that bridges biomechanics, actuator physics, and system integration. We hope it serves as a foundation for more transparent, comparable, and scientifically grounded claims about what humanoid robots can do.

\section*{Acknowledgement}

I am grateful to Shreyash Shambharkar for providing feedback and corrections to the DOF Atlas.


\appendix
\section{Glossary of Symbols}
\label{app:glossary}

This glossary provides definitions for all symbols used in the Human-Equivalence Envelope (HEE) and Human-Level Actuation Score (HLAS) framework. Symbols are organized by category for easy reference.

\begin{longtable}{@{}l l p{0.58\linewidth}@{}}
\caption{Main symbols used in the Human-Equivalence Envelope (HEE) and Human-Level Actuation Score (HLAS) definitions. Unless noted, dimensionless quantities are normalized to $[0,1]$.}
\label{tab:glossary}\\
\toprule
\textbf{Symbol} & \textbf{Units} & \textbf{Meaning} \\
\midrule
\endfirsthead

\multicolumn{3}{c}{\tablename\ \thetable\ -- \textit{Continued from previous page}} \\
\toprule
\textbf{Symbol} & \textbf{Units} & \textbf{Meaning} \\
\midrule
\endhead

\midrule
\multicolumn{3}{r}{\textit{Continued on next page}} \\
\endfoot

\bottomrule
\endlastfoot

\multicolumn{3}{@{}l}{\textit{\textbf{Indices, sets, and task bands}}}\\[2pt]
$j$ & -- & Joint index (e.g., ankle, knee, hip, shoulder). \\
$t$ & -- & Task index in task set $\mathcal{T}$ (e.g., walking, stairs, reaching). \\
$a$ & -- & Axis index within a joint (e.g., flexion/extension). \\
$\mathcal{T}$ & -- & Set of reference tasks evaluated in HLAS. \\
$\mathcal{J}_t$ & -- & Set of joints contributing to task $t$. \\
$\mathcal{R}_{j,t}$ & -- & Task-derived operating band $\subseteq\{(q,\omega)\}$ for joint $j$ in task $t$. \\
$\mathcal{A}_{j,t}$ & -- & Set of axes at joint $j$ used by task $t$. \\[6pt]

\multicolumn{3}{@{}l}{\textit{\textbf{Kinematics and human reference scaling}}}\\[2pt]
$q$ & rad, deg & Joint angle following ISB conventions (Sec.~\ref{sec:kinematics}). \\
$\omega$ & rad/s & Joint angular velocity. \\
$\dot{\omega}$ & rad/s$^2$ & Joint angular acceleration. \\
$m$ & kg & Reference human body mass (75\,kg). \\
$h$ & m & Reference human height (1.75\,m). \\
$T/m$ & Nm/kg & Mass-normalized human joint moment from literature. \\
$P/m$ & W/kg & Mass-normalized human joint power from literature. \\
$T_{\mathrm{abs}}$ & Nm & Absolute joint torque target, $T_{\mathrm{abs}} = m \cdot (T/m)$. \\
$P_{\mathrm{abs}}$ & W & Absolute joint power target, $P_{\mathrm{abs}} = m \cdot (P/m)$. \\[6pt]

\multicolumn{3}{@{}l}{\textit{\textbf{Robot torque and power fields}}}\\[2pt]
$T^{\mathrm{rob}}_{j}(q,\omega)$ & Nm & Robot continuous-safe torque at $(q,\omega)$. \\
$P^{\mathrm{rob}}_{j}(q,\omega)$ & W & Robot mechanical power, $T^{\mathrm{rob}}_{j} \cdot \omega$. \\
$P_{\mathrm{mech}}$ & W & Measured mechanical power (shaft torque $\times$ velocity). \\
$P_{\mathrm{elec}}$ & W & Measured electrical power ($V_{\mathrm{bus}} \times I_{\mathrm{bus}}$, true RMS). \\
$T^{\mathrm{peak}}_{j}$ & Nm & Peak burst torque ($<500$\,ms). \\
$T^{\mathrm{cont}}_{j}$ & Nm & Continuous torque without thermal derating. \\
$T^{\mathrm{cont}}_{j}\big|_{\text{duty}}$ & Nm & Continuous plateau at task duty (Eq.~\ref{eq:thermal}). \\[6pt]

\multicolumn{3}{@{}l}{\textit{\textbf{Human requirement fields}}}\\[2pt]
$T^{\mathrm{hum}}_{j,t}(q,\omega)$ & Nm & Human torque requirement for $(j,t)$ at $(q,\omega)$. \\
$P^{\mathrm{hum}}_{j,t}(q,\omega)$ & W & Human power requirement for $(j,t)$ at $(q,\omega)$. \\
$P^{\mathrm{hum}+}_{j,t}(q,\omega)$ & W & Positive power: $\max(P^{\mathrm{hum}}_{j,t}, 0)$ (HEE weighting). \\
$T^{\mathrm{req}}_{j,t}\big|_{\text{plateau}}$ & Nm & Human plateau torque for $(j,t)$ (thermal comparison). \\
$\omega^{\mathrm{req}}_{j,t}$ & rad/s & Required peak rate for $(j,t)$ from human data. \\[6pt]

\multicolumn{3}{@{}l}{\textit{\textbf{Workspace and kinematic sufficiency}}}\\[2pt]
$I^{\mathrm{func}}_{j,t}(a)$ & deg & Functional human ROM for axis $a$ in task $t$ (Table~\ref{tab:rom_compact}). \\
$I^{\mathrm{rob}}_{j}(a)$ & deg & Robot safe ROM for axis $a$ at joint $j$. \\
$\rho^{\mathrm{ROM}}_{j,t}$ & [0,1] & ROM coverage factor (Eq.~\ref{eq:rom}). \\
$d^{\mathrm{DoF}}_{j,t}$ & [0,1] & DoF sufficiency (Eq.~\ref{eq:dof}). \\[6pt]

\multicolumn{3}{@{}l}{\textit{\textbf{Human-Equivalence Envelope and delivery}}}\\[2pt]
$h^{(w)}_{j,t}$ & [0,1] & HEE: power-weighted fraction where robot meets human torque \& power simultaneously (Eq.~\ref{eq:hee}). \\
$w(q,\omega)$ & -- & Normalized sample weight $\propto P^{\mathrm{hum}+}_{j,t}$; $\sum w = 1$. \\
$m^{T}_{j,t}$ & [0,1] & Torque margin: $\min \mathrm{clip}(T^{\mathrm{rob}}/T^{\mathrm{hum}})$ (Eq.~\ref{eq:torque}). \\
$m^{P}_{j,t}$ & [0,1] & Power margin: $\min \mathrm{clip}(P^{\mathrm{rob}}/P^{\mathrm{hum}})$ (Eq.~\ref{eq:power}). \\[6pt]

\multicolumn{3}{@{}l}{\textit{\textbf{Bandwidth and rate}}}\\[2pt]
$f^{\tau}_{c,j}$ & Hz & Torque-mode bandwidth: $-3$\,dB crossover (Benchmark 4). \\
$f^{\star}_{t,j}$ & Hz & Target bandwidth for $(j,t)$ (pre-specified). \\
$b^{\tau}_{j,t}$ & [0,1] & Bandwidth margin: $\mathrm{clip}(f^{\tau}_{c,j}/f^{\star}_{t,j})$ (Eq.~\ref{eq:bandwidth}). \\
$\omega^{\max}_{j}$ & rad/s & Maximum safe joint rate under task load. \\
$m^{\omega}_{j,t}$ & [0,1] & Rate margin: $\mathrm{clip}(\omega^{\max}_j / \omega^{\mathrm{req}}_{j,t})$ (Eq.~\ref{eq:rate}). \\[6pt]

\multicolumn{3}{@{}l}{\textit{\textbf{Efficiency and thermal sustainability}}}\\[2pt]
$\eta_j(q,\omega)$ & -- & Electromechanical efficiency: $P_{\mathrm{mech}} / P_{\mathrm{elec}}$. \\
$\bar{\eta}_{j,t}$ & -- & Task-weighted mean efficiency over $\mathcal{R}_{j,t}$ (Eq.~\ref{eq:eff}). \\
$\eta^\star_{t,j}$ & -- & Target efficiency for $(j,t)$ (e.g., 0.80 for walking). \\
$\eta_{j,t}$ & [0,1] & Efficiency margin: $\mathrm{clip}(\bar{\eta}_{j,t} / \eta^\star_{t,j})$ (Eq.~\ref{eq:eff}). \\
$\theta^{\mathrm{therm}}_{j,t}$ & [0,1] & Thermal margin: continuous/required plateau ratio (Eq.~\ref{eq:thermal}). \\
$T_{\mathrm{motor}}$ & °C & Motor winding temperature (logged during tests). \\
$T_{\mathrm{gear}}$ & °C & Gearbox housing temperature (logged during tests). \\[6pt]

\multicolumn{3}{@{}l}{\textit{\textbf{Aggregation: weights and scores}}}\\[2pt]
$w_t$ & -- & Task weight; $\sum_{t\in\mathcal{T}} w_t = 1$. \\
$u_{j,t}$ & -- & Joint weight in task $t$; $\sum_{j\in\mathcal{J}_t} u_{j,t} = 1$. \\
$\boldsymbol{\alpha}$ & -- & Feature-weight vector; $\sum_k \alpha_k = 1$ (Eq.~\ref{eq:hlas-expanded}). \\
$\mathbf{x}_{j,t}$ & -- & Feature vector: $[\rho^{\mathrm{ROM}}, d^{\mathrm{DoF}}, h^{(w)}, b^{\tau}, \eta, \theta^{\mathrm{therm}}]^{\!\top}$. \\
$s_{j,t}$ & [0,1] & Joint-task score: $\boldsymbol{\alpha}^{\!\top} \mathbf{x}_{j,t}$. \\
$s_t$ & [0,1] & Task score: $\sum_{j \in \mathcal{J}_t} u_{j,t} s_{j,t}$. \\
$\mathrm{HLAS}$ & [0,1] & Human-Level Actuation Score: $\sum_{t} w_t s_t$ (Eq.~\ref{eq:hlas-compact}). \\[6pt]

\multicolumn{3}{@{}l}{\textit{\textbf{Transparency and backdrivability}}}\\[2pt]
$J_{\mathrm{ref}}$ & kg·m$^2$ & Reflected inertia at joint (Benchmark 3). \\
$b$ & Nm·s/rad & Viscous friction coefficient. \\
$f_c$ & Nm & Coulomb friction torque. \\
$\tau_{\mathrm{bd}}$ & Nm & Gravity-compensated backdrive torque (Benchmark 3). \\[6pt]

\multicolumn{3}{@{}l}{\textit{\textbf{Utility functions and conventions}}}\\[2pt]
$\mathrm{clip}_{[0,1]}(x)$ & -- & Saturation: $\min(1, \max(0, x))$. \\
$\mathbb{1}[\cdot]$ & -- & Indicator: 1 if condition true, 0 otherwise. \\
$|\cdot|$ & -- & Interval length (ROM) or set cardinality. \\
ISB & -- & Int'l Society of Biomechanics coord. conventions (Sec.~\ref{sec:kinematics}). \\

\end{longtable}

\paragraph{Usage notes:}
\begin{itemize}[leftmargin=12pt,itemsep=2pt]
\item \textbf{Continuous-safe vs. peak:} All $T^{\mathrm{rob}}$ symbols refer to continuous-safe measurements unless marked ``peak.''
\item \textbf{Dimensionless ranges:} Symbols in $[0,1]$ are normalized. 1.0 = full human capability; 0.0 = no capability.
\item \textbf{Power weighting:} Superscript ``+'' (e.g., $P^{\mathrm{hum}+}$) denotes positive power only: $\max(P, 0)$.
\item \textbf{Equation references:} Each symbol's defining equation is cited in the ``Meaning'' column.
\end{itemize}

\paragraph{Usage notes:}
\begin{itemize}[leftmargin=12pt,itemsep=2pt]
\item \textbf{Continuous-safe vs. peak:} All torque/power symbols with superscript ``rob'' (e.g., $T^{\mathrm{rob}}_j$) refer to continuous-safe measurements unless explicitly marked ``peak.''
\item \textbf{Dimensionless ranges:} Symbols in $[0,1]$ are normalized factors or margins. A value of 1.0 indicates full human capability; 0.0 indicates no capability.
\item \textbf{Power weighting:} The superscript ``+'' (e.g., $P^{\mathrm{hum}+}$) denotes positive power only: $\max(P, 0)$. Negative power regions carry zero weight in HEE and efficiency.
\item \textbf{Equation cross-references:} Each symbol's defining equation is cited in the ``Meaning'' column for quick lookup.
\end{itemize}

{\small
\bibliographystyle{plainnat}
\bibliography{references}

@book{Winter2009,
  author    = {Winter, David A.},
  title     = {Biomechanics and Motor Control of Human Movement},
  edition   = {4},
  year      = {2009},
  publisher = {John Wiley \& Sons},
  address   = {Hoboken, NJ}
}

@article{Farris2012,
  author  = {Farris, Dominic J. and Sawicki, Gregory S.},
  title   = {The mechanics and energetics of human walking and running: a joint-level perspective},
  journal = {Journal of The Royal Society Interface},
  year    = {2012},
  volume  = {9},
  number  = {66},
  pages   = {110--118}
}

@article{Zelik2016,
  author  = {Zelik, Karl E. and Adamczyk, Peter G.},
  title   = {A unified perspective on ankle push-off in human walking},
  journal = {Journal of Experimental Biology},
  year    = {2016}
}

@article{DeLeva1996,
  author  = {De Leva, Paolo},
  title   = {Adjustments to {Zatsiorsky--Seluyanov}'s segment inertia parameters},
  journal = {Journal of Biomechanics},
  year    = {1996},
  volume  = {29},
  number  = {9},
  pages   = {1223--1230}
}

@article{Kuo2001,
  author  = {Kuo, Arthur D.},
  title   = {A simple model of bipedal walking predicts the preferred speed--step length relationship},
  journal = {Journal of Biomechanical Engineering},
  year    = {2001},
  volume  = {123},
  number  = {3},
  pages   = {264--269}
}

@article{Kuo2002,
  author  = {Kuo, Arthur D.},
  title   = {Energetics of actively powered locomotion using the simplest walking model},
  journal = {Journal of Biomechanical Engineering},
  year    = {2002},
  volume  = {124},
  number  = {1},
  pages   = {113--120}
}

@article{Kram1990,
  author  = {Kram, Rodger and Taylor, C. Richard},
  title   = {Energetics of running: a new perspective},
  journal = {Nature},
  year    = {1990},
  volume  = {346},
  number  = {6281},
  pages   = {265--267}
}

@article{Gottschall2003,
  author  = {Gottschall, Jinger S. and Kram, Rodger},
  title   = {Energy cost and muscular activity required for leg swing during walking},
  journal = {Journal of Applied Physiology},
  year    = {2005},
  volume  = {99},
  number  = {1},
  pages   = {23--30}
}

@article{Halsey2012StairClimbing,
  author  = {Halsey, Lewis G. and Watkins, David A. R. and Duggan, Brendan M.},
  title   = {The energy expenditure of stair climbing one step and two steps at a time: estimations from measures of heart rate},
  journal = {PLoS One},
  year    = {2012},
  volume  = {7},
  number  = {12},
  pages   = {e51213},
  doi     = {10.1371/journal.pone.0051213},
  issn    = {1932-6203}
}

@book{AAOSROM2020,
  author    = {{American Academy of Orthopaedic Surgeons}},
  title     = {Joint Motion: Method of Measuring and Recording},
  year      = {1965},
  publisher = {American Academy of Orthopaedic Surgeons},
  address   = {Chicago, IL},
  note      = {Source of AAOS range-of-motion reference tables}
}

@article{Morrey1981,
  author  = {Morrey, Bernard F. and Askew, Laurence J. and Chao, Edmund Y. S.},
  title   = {A biomechanical study of normal functional elbow motion},
  journal = {Journal of Bone and Joint Surgery A},
  year    = {1981},
  volume  = {63},
  number  = {6},
  pages   = {872--877}
}

@article{Palmer1985,
  author  = {Palmer, Anne K. and Werner, Francis W.},
  title   = {The triangular fibrocartilage complex of the wrist: anatomy and function},
  journal = {Journal of Hand Surgery},
  year    = {1985},
  volume  = {10},
  number  = {4},
  pages   = {580--588}
}

@article{Wu2002ISB,
  author  = {Wu, Ge and Siegler, Sorin and Allard, Paul and Kirtley, Chris and Leardini, Alberto and Rosenbaum, Dieter and Whittle, Mike and D'Lima, Darryl D. and Cristofolini, Luca and Witte, Hartmut and Schmid, Oskar and Stokes, Ian},
  title   = {ISB recommendation on definitions of joint coordinate system of various joints for the reporting of human joint motion -- part I: ankle, hip, and spine},
  journal = {Journal of Biomechanics},
  year    = {2002},
  volume  = {35},
  number  = {4},
  pages   = {543--548},
  doi     = {10.1016/S0021-9290(01)00222-6}
}

@article{Wu2005ISB,
  author  = {Wu, Ge and van der Helm, Frans C. T. and Veeger, H. E. J. (DirkJan) and Makhsous, Mohsen and van Roy, Peter and Anglin, Carolyn and Nagels, Jochem and Karduna, Andrew R. and McQuade, Kevin and Wang, Xuguang and Werner, Frederick W. and Buchholz, Bryan},
  title   = {ISB recommendation on definitions of joint coordinate systems of various joints for the reporting of human joint motion -- Part II: shoulder, elbow, wrist and hand},
  journal = {Journal of Biomechanics},
  year    = {2005},
  volume  = {38},
  number  = {5},
  pages   = {981--992},
  doi     = {10.1016/j.jbiomech.2004.05.042}
}

@article{Grood1983,
  author  = {Grood, Edward S. and Suntay, Wayne J.},
  title   = {A joint coordinate system for the clinical description of three-dimensional motions: application to the knee},
  journal = {Journal of Biomechanical Engineering},
  year    = {1983},
  volume  = {105},
  number  = {2},
  pages   = {136--144}
}

@article{Cappozzo1995,
  author  = {Cappozzo, Aurelio and Catani, Fabio and Della Croce, Ugo and Leardini, Alberto},
  title   = {Position and orientation in space of bones during movement: experimental artefacts},
  journal = {Clinical Biomechanics},
  year    = {1996},
  volume  = {11},
  number  = {2},
  pages   = {90--100}
}

@article{Delp2007OpenSim,
  author  = {Delp, Scott L. and Anderson, Frank C. and Arnold, Allison S. and others},
  title   = {OpenSim: open-source software to create and analyze dynamic simulations of movement},
  journal = {IEEE Transactions on Biomedical Engineering},
  year    = {2007},
  volume  = {54},
  number  = {11},
  pages   = {1940--1950}
}

@article{Seth2018OpenSim,
  author  = {Seth, Ajay and Hicks, Jennifer L. and Uchida, Thomas K. and Habib, Ayman and Dembia, Christopher L. and Dunne, Joshua J. and Ong, Carmichael F. and DeMers, Matthew S. and Rajagopal, Apoorva and Millard, Matthew and Hamner, Samuel R. and Arnold, Allison S. and Yong, Jeongho and Lakshmikanth, Shrinidhi K. and Sherman, Michael A. and Ku, Johnson P. and Delp, Scott L.},
  title   = {OpenSim: Simulating musculoskeletal dynamics and neuromuscular control to study human and animal movement},
  journal = {PLOS Computational Biology},
  year    = {2018},
  volume  = {14},
  number  = {7},
  pages   = {e1006223},
  doi     = {10.1371/journal.pcbi.1006223},
  url     = {https://pubmed.ncbi.nlm.nih.gov/30048444/}
}

@inproceedings{Pratt1995,
  author    = {Pratt, Gill A. and Williamson, Matthew M.},
  title     = {Series elastic actuators},
  booktitle = {Proceedings of the IEEE/RSJ International Conference on Intelligent Robots and Systems},
  year      = {1995},
  pages     = {399--406}
}

@article{Vanderborght2013,
  author  = {Vanderborght, Bram and Albu-Sch{\"a}ffer, Alin and Bicchi, Antonio and Burdet, Etienne and Caldwell, Darwin G. and Carloni, Raffaella and Catalano, Manuel G. and Eiberger, Oliver and Friedl, Werner and Ganesh, Gowrishankar and Garabini, Manolo and Grebenstein, Markus and Grioli, Giorgio and Haddadin, Sami and Hoppner, Hannes and Jafari, Amir and Laffranchi, Matteo and Lefeber, Dirk and Petit, Florin and Stramigioli, Stefano and Tsagarakis, Nikos G. and Van Damme, Moritz and Van Ham, Ronald and Visser, Lars C. and Wolf, Sebastian},
  title   = {Variable impedance actuators: A review},
  journal = {Robotics and Autonomous Systems},
  year    = {2013},
  volume  = {61},
  number  = {12},
  pages   = {1601--1614},
  doi     = {10.1016/j.robot.2013.06.009}
}

@article{Paine2015,
  author  = {Paine, Nicholas and Oh, Sehoon and Sentis, Luis},
  title   = {Design and control considerations for high-performance series elastic actuators},
  journal = {IEEE/ASME Transactions on Mechatronics},
  year    = {2014},
  volume  = {19},
  number  = {3},
  pages   = {1080--1091},
  doi     = {10.1109/TMECH.2013.2270435}
}

@inproceedings{Hutter2016ANYmal,
  author    = {Hutter, Marco and Gehring, Christian and Jud, Dominic and Lauber, Andreas
               and Bellicoso, C. Dario and Tsounis, Vassilios and Hwangbo, Jemin
               and Bodie, Karen and Fankhauser, Peter and Bloesch, Michael
               and Diethelm, Remo and Bachmann, Samuel and Melzer, Amir
               and Hoepflinger, Mark A.},
  title     = {ANYmal -- A Highly Mobile and Dynamic Quadrupedal Robot},
  booktitle = {2016 IEEE/RSJ International Conference on Intelligent Robots and Systems (IROS)},
  year      = {2016},
  pages     = {38--44},
  publisher = {IEEE},
  doi       = {10.1109/IROS.2016.7758092}
}

@article{Grimminger,
  author  = {Grimminger, Felix and Meduri, Avadesh and Khadiv, Majid and Viereck, Julian and Wuthrich, Manuel and Naveau, Maximilien and Berenz, Vincent and Heim, Steve and Widmaier, Felix and Flayols, Thomas and Fiene, Jonathan and Badri-Sprowitz, Alexander and Righetti, Ludovic},
  title   = {An Open Torque-Controlled Modular Robot Architecture for Legged Locomotion Research},
  journal = {IEEE Robotics and Automation Letters},
  year    = {2020},
  volume  = {5},
  number  = {2},
  pages   = {3650--3657},
  doi     = {10.1109/LRA.2020.2976639}
}

@inproceedings{Katz2019,
  author    = {Katz, Benjamin and Di Carlo, Jared and Kim, Sangbae},
  title     = {Mini Cheetah: A platform for pushing the limits of dynamic quadruped control},
  booktitle = {Proceedings of the IEEE International Conference on Robotics and Automation},
  year      = {2019},
  pages     = {6295--6301}
}

@article{LopezGarcia2020,
  author  = {L{\'o}pez Garc{\'i}a, Pablo and Crispel, Stein and Saerens, Elias and Verstraten, Tom and Lefeber, Dirk},
  title   = {Compact Gearboxes for Modern Robotics: A Review},
  journal = {Frontiers in Robotics and AI},
  year    = {2020},
  volume  = {7},
  pages   = {103},
  doi     = {10.3389/frobt.2020.00103},
  url     = {https://www.frontiersin.org/articles/10.3389/frobt.2020.00103},
  note    = {Review Article}
}

@article{Oberneder2024,
  author  = {Oberneder, Florian and Landler, Stefan and Otto, Michael and Vogel-Heuser, Birgit and Zimmermann, Markus and Stahl, Karsten},
  title   = {Influences of different parameters on selected properties of gears for robot-like systems},
  journal = {Frontiers in Robotics and AI},
  year    = {2024},
  volume  = {11},
  pages   = {1414238},
  doi     = {10.3389/frobt.2024.1414238},
  url     = {https://www.frontiersin.org/articles/10.3389/frobt.2024.1414238}
}

@article{Hogan1985,
  author  = {Hogan, Neville},
  title   = {Impedance control: an approach to manipulation},
  journal = {Journal of Dynamic Systems, Measurement, and Control},
  year    = {1985},
  volume  = {107},
  number  = {1},
  pages   = {1--7}
}

@inproceedings{ColgateBrown1994,
  author    = {Colgate, J. Edward and Brown, J. Michael},
  title     = {Factors affecting the {Z}-width of a haptic display},
  booktitle = {Proceedings of the IEEE International Conference on Robotics and Automation},
  year      = {1994},
  pages     = {3205--3210}
}

@book{Dvir2025,
  editor    = {Dvir, Zeevi},
  title     = {Isokinetics: Muscle Testing, Interpretation and Clinical Applications},
  edition   = {3},
  year      = {2025},
  publisher = {Routledge},
  address   = {New York},
  doi       = {10.4324/9781003380719},
  isbn      = {9781032462394}
}

@article{Ralston1958,
  author  = {Ralston, H. J.},
  title   = {Energy-speed relation and optimal speed during level walking},
  journal = {Arbeitsphysiologie},
  year    = {1958},
  volume  = {17},
  number  = {4},
  pages   = {277--283}
}

@article{Umberger2003,
  author  = {Umberger, Brian R. and Gerritsen, Karin G. M. and Martin, Philip E.},
  title   = {A model of human muscle energy expenditure},
  journal = {Computer Methods in Biomechanics and Biomedical Engineering},
  year    = {2003},
  volume  = {6},
  number  = {2},
  pages   = {99--111},
  doi     = {10.1080/1025584031000091678}
}

@article{Ortega2008,
  author  = {Ortega, Justus D. and Fehlman, Leslie A. and Farley, Claire T.},
  title   = {Effects of aging and arm swing on the metabolic cost of stability in human walking},
  journal = {Journal of Biomechanics},
  year    = {2008},
  volume  = {41},
  number  = {16},
  pages   = {3303--3308},
  doi     = {10.1016/j.jbiomech.2008.06.039}
}

@article{Mentiplay2018,
  author  = {Mentiplay, Benjamin F. and Banky, Megan and Clark, Ross A. and Kahn, Michelle B. and Williams, Gavin},
  title   = {Lower limb angular velocity during walking at various speeds},
  journal = {Gait \& Posture},
  year    = {2018},
  volume  = {65},
  pages   = {190--196},
  doi     = {10.1016/j.gaitpost.2018.06.162}
}

@article{Baltrusch2019,
  author  = {Baltrusch, S. J. and van Die{\'e}n, J. H. and Bruijn, S. M. and Koopman, A. S. and van Bennekom, C. A. M. and Houdijk, H.},
  title   = {The effect of a passive trunk exoskeleton on metabolic costs during lifting and walking},
  journal = {Ergonomics},
  year    = {2019},
  volume  = {62},
  number  = {7},
  pages   = {903--916},
  doi     = {10.1080/00140139.2019.1602288}
}

@article{Sarabon2021Knee,
  author  = {Sarabon, Nejc and Kozinc, Ziga and Perman, Mihael},
  title   = {Establishing reference values for isometric knee extension and flexion strength},
  journal = {Frontiers in Physiology},
  year    = {2021},
  volume  = {12},
  pages   = {767941}
}

@article{Seroyer2010,
  author  = {Seroyer, Shane T. and Nho, Shane J. and Bach, Bernard R. and Bush-Joseph, Charles A. and Nicholson, Gregory P. and Romeo, Anthony A.},
  title   = {The kinetic chain in overhand pitching: its potential role for performance enhancement and injury prevention},
  journal = {Sports Health},
  year    = {2010},
  volume  = {2},
  number  = {2},
  pages   = {135--146},
  doi     = {10.1177/1941738110362656}
}

@article{Shimoyama1990,
  author  = {Shimoyama, Ikuo and Ninchoji, T. and Uemura, K.},
  title   = {The finger-tapping test: a quantitative analysis},
  journal = {Archives of Neurology},
  year    = {1990},
  volume  = {47},
  number  = {6},
  pages   = {681--684}
}

@article{Neptune2001,
  author  = {Neptune, Richard R. and Kautz, Steven A. and Zajac, Felix E.},
  title   = {Contributions of the individual ankle plantar flexors to support, forward progression and swing initiation during walking},
  journal = {Journal of Biomechanics},
  year    = {2001},
  volume  = {34},
  number  = {11},
  pages   = {1387--1398}
}

@article{Anderson2001,
  author  = {Anderson, Frank C. and Pandy, Marcus G.},
  title   = {Dynamic optimization of human walking},
  journal = {Journal of Biomechanical Engineering},
  year    = {2001},
  volume  = {123},
  number  = {5},
  pages   = {381--390}
}

@article{Dodds2014,
  author  = {Dodds, Richard M. and Syddall, Holly E. and Cooper, Rachel and Benzeval, Michaela and Deary, Ian J. and Dennison, Elaine M. and Der, Geoff and Gale, Catharine R. and Inskip, Hazel M. and Jagger, Carol and Kirkwood, Thomas B. and Lawlor, Debbie A. and Robinson, Sian M. and Starr, John M. and Steptoe, Andrew and Tilling, Kate and Kuh, Diana and Cooper, Cyrus and Sayer, Avan Aihie},
  title   = {Grip Strength across the Life Course: Normative Data from Twelve British Studies},
  journal = {PLOS ONE},
  year    = {2014},
  volume  = {9},
  number  = {12},
  pages   = {e113637},
  doi     = {10.1371/journal.pone.0113637},
  url     = {https://journals.plos.org/plosone/article?id=10.1371/journal.pone.0113637}
}

@article{Bohannon2019,
  author  = {Bohannon, Richard W.},
  title   = {Grip strength: an indispensable biomarker for older adults},
  journal = {Clinical Interventions in Aging},
  year    = {2019},
  volume  = {14},
  pages   = {1681--1691}
}

@article{Maffiuletti2010,
  author  = {Maffiuletti, Nicola A.},
  title   = {Assessment of hip and knee muscle function in orthopaedic practice and research},
  journal = {Journal of Bone and Joint Surgery A},
  year    = {2010},
  volume  = {92},
  number  = {1},
  pages   = {220--229}
}

@article{Zajac1989,
  author  = {Zajac, Felix E.},
  title   = {Muscle and tendon: properties, models, scaling, and application to biomechanics and motor control},
  journal = {Critical Reviews in Biomedical Engineering},
  year    = {1989},
  volume  = {17},
  number  = {4},
  pages   = {359--411}
}

@inproceedings{Carpentier2019Pinocchio,
  author    = {Carpentier, Justin and Saurel, Guilhem and Buondonno, Gabriele and
               Mirabel, Joseph and Lamiraux, Florent and Stasse, Olivier and
               Mansard, Nicolas},
  title     = {The {Pinocchio} {C++} library -- A fast and flexible implementation
               of rigid body dynamics algorithms and their analytical derivatives},
  booktitle = {SII 2019 -- International Symposium on System Integrations},
  address   = {Paris, France},
  year      = {2019},
  month     = jan,
  url       = {https://hal.laas.fr/hal-01866228},
  note      = {Rapport LAAS n{\textdegree} 18288}
}

@article{Felis2017RBDL,
  author  = {Felis, Martin L.},
  title   = {{RBDL}: an efficient rigid-body dynamics library using recursive algorithms},
  journal = {Autonomous Robots},
  year    = {2017},
  volume  = {41},
  number  = {2},
  pages   = {495--511}
}

@inproceedings{Todorov2012MuJoCo,
  author    = {Todorov, Emanuel and Erez, Tom and Tassa, Yuval},
  title     = {{MuJoCo}: a physics engine for model-based control},
  booktitle = {Proceedings of the IEEE/RSJ International Conference on Intelligent Robots and Systems},
  year      = {2012},
  pages     = {5026--5033}
}

@inproceedings{Kajita2003,
  author    = {Kajita, Shuuji and Kanehiro, Fumio and Kaneko, Kenji and Fujiwara, Kiyoshi
               and Harada, Kensuke and Yokoi, Kazuhito and Hirukawa, Hirohisa},
  title     = {Biped walking pattern generation by using preview control of zero-moment point},
  booktitle = {Proceedings of the 2003 IEEE International Conference on Robotics and Automation (ICRA)},
  year      = {2003},
  volume    = {2},
  pages     = {1620--1626},
  publisher = {IEEE},
  address   = {Taipei, Taiwan},
  doi       = {10.1109/ROBOT.2003.1241826}
}

@article{Saab2013,
  author  = {Saab, Layale and Ramos, Oscar E. and Keith, Fran\c{c}ois and Mansard, Nicolas and Sou\`eres, Philippe and Fourquet, Jean-Yves},
  title   = {Dynamic Whole-Body Motion Generation Under Rigid Contacts and Other Unilateral Constraints},
  journal = {IEEE Transactions on Robotics},
  year    = {2013},
  volume  = {29},
  number  = {2},
  pages   = {346--362},
  doi     = {10.1109/TRO.2012.2234351}
}

@article{Herzog2014,
  author  = {Herzog, Alexander and Rotella, Nicholas and Mason, Sean and Grimminger, Felix and Schaal, Stefan and Righetti, Ludovic},
  title   = {Momentum Control with Hierarchical Inverse Dynamics on a Torque-Controlled Humanoid},
  journal = {CoRR},
  volume  = {abs/1410.7284},
  year    = {2014},
  url     = {https://arxiv.org/abs/1410.7284},
  eprint  = {1410.7284},
  archivePrefix = {arXiv}
}

@article{Semini2011HyQ,
  author  = {Semini, Claudio and Tsagarakis, Nikos G. and Guglielmino, Emanuele and Focchi, Michele and Cannella, Ferdinando and Caldwell, Darwin G.},
  title   = {Design of HyQ -- a hydraulically and electrically actuated quadruped robot},
  journal = {Proceedings of the Institution of Mechanical Engineers, Part I: Journal of Systems and Control Engineering},
  year    = {2011},
  volume  = {225},
  number  = {6},
  pages   = {831--849},
  doi     = {10.1177/0959651811402275}
}

@inproceedings{Grebenstein2011,
  author    = {Grebenstein, Markus and Albu-Sch{\"a}ffer, Alin and Bahls, Thomas
               and Chalon, Maxime and Eiberger, Oliver and Friedl, Werner
               and Gruber, Robin and Haddadin, Sami and Hagn, Ulrich
               and Haslinger, Robert and H{\"o}ppner, Hannes and J{\"o}rg, Stefan
               and Nickl, Mathias and Nothhelfer, Alexander and Petit, Florian
               and Reill, Josef and Seitz, Nikolaus and Wimb{\"o}ck, Thomas
               and Wolf, Sebastian and W{\"u}sthoff, Tilo and Hirzinger, Gerd},
  title     = {The DLR hand arm system},
  booktitle = {2011 IEEE International Conference on Robotics and Automation (ICRA)},
  year      = {2011},
  pages     = {3175--3182},
  publisher = {IEEE},
  address   = {Shanghai, China},
  doi       = {10.1109/ICRA.2011.5980371}
}

@article{Donelan2008,
  author  = {Donelan, J. Maxwell and Li, Qingguo and Naing, Vincent and others},
  title   = {Biomechanical energy harvesting: generating electricity during walking with minimal user effort},
  journal = {Science},
  year    = {2008},
  volume  = {319},
  number  = {5864},
  pages   = {807--810}
}

@book{Hanselman1994,
  author    = {Hanselman, Duane C.},
  title     = {Brushless Permanent-Magnet Motor Design},
  publisher = {McGraw-Hill},
  year      = {1994},
  address   = {New York, NY}
}

@article{Kuindersma2016Atlas,
  author  = {Kuindersma, Scott and Deits, Robin and Fallon, Maurice and Valenzuela, Andr{\'e}s and Dai, Hongkai and Permenter, Frank and Koolen, Twan and Marion, Pat and Tedrake, Russ},
  title   = {Optimization-based locomotion planning, estimation, and control design for the Atlas humanoid robot},
  journal = {Autonomous Robots},
  year    = {2016},
  volume  = {40},
  number  = {3},
  pages   = {429--455},
  doi     = {10.1007/s10514-015-9479-3}
}

@article{Tong2024,
  author  = {Tong, Yuchuang and Liu, Haotian and Zhang, Zhengtao},
  title   = {Advancements in humanoid robots: A comprehensive review and future prospects},
  journal = {IEEE/CAA Journal of Automatica Sinica},
  year    = {2024},
  volume  = {11},
  number  = {2},
  pages   = {301--328},
  month   = feb,
  doi     = {10.1109/JAS.2023.124140}
}

@inproceedings{Bledt2018,
  author    = {Bledt, Gerardo and Powell, Matthew J. and Katz, Benjamin and {Di Carlo}, Jared and Wensing, Patrick M. and Kim, Sangbae},
  title     = {MIT Cheetah 3: Design and Control of a Robust, Dynamic Quadruped Robot},
  booktitle = {2018 IEEE/RSJ International Conference on Intelligent Robots and Systems (IROS)},
  year      = {2018},
  pages     = {2245--2252},
  publisher = {IEEE},
  doi       = {10.1109/IROS.2018.8593885}
}

@inproceedings{Seok2013,
  author    = {Seok, Sangok and Wang, Albert and Chuah, Meng Yee and Otten, David and Lang, Jeffrey and Kim, Sangbae},
  title     = {Design Principles for Energy-Efficient Legged Locomotion and Implementation on the {MIT} Cheetah Robot},
  booktitle = {2013 IEEE International Conference on Robotics and Automation (ICRA)},
  year      = {2013},
  pages     = {3307--3312},
  publisher = {IEEE},
  doi       = {10.1109/ICRA.2013.6631020}
}

@article{Collins2015,
  author  = {Collins, Steven H. and Wiggin, Matthew B. and Sawicki, Gregory S.},
  title   = {Reducing the energy cost of human walking using an unpowered exoskeleton},
  journal = {Nature},
  year    = {2015},
  volume  = {522},
  number  = {7555},
  pages   = {212--215}
}

@article{Zhang2017,
  author  = {Zhang, Juanjuan and Fiers, Pieter and Witte, Kirby A. and Jackson, Rachel W. and Poggensee, Katherine L. and Atkeson, Christopher G. and Collins, Steven H.},
  title   = {Human-in-the-loop optimization of exoskeleton assistance during walking},
  journal = {Science},
  year    = {2017},
  volume  = {356},
  number  = {6344},
  pages   = {1280--1284},
  doi     = {10.1126/science.aal5054}
}

@article{HannafordRyu2002,
  author  = {Hannaford, Blake and Ryu, Jee-Hwan},
  title   = {Time-domain passivity control of haptic interfaces},
  journal = {IEEE Transactions on Robotics and Automation},
  year    = {2002},
  volume  = {18},
  number  = {1},
  pages   = {1--10}
}

@article{DeVita2000,
  author  = {DeVita, Paul and Hortob{\'a}gyi, Tibor},
  title   = {Age causes a redistribution of joint torques and powers during gait},
  journal = {Journal of Applied Physiology},
  year    = {2000},
  volume  = {88},
  number  = {5},
  pages   = {1804--1811},
  doi     = {10.1152/jappl.2000.88.5.1804}
}

@article{Cupp1999,
  author  = {Cupp, T. and Oeffinger, D. and Tylkowski, C. and Augsburger, S.},
  title   = {Age-related kinetic changes in normal pediatrics},
  journal = {Journal of Pediatric Orthopaedics},
  year    = {1999},
  volume  = {19},
  number  = {4},
  pages   = {475--478},
  doi     = {10.1097/00004694-199907000-00010}
}

@online{pixabay-skeleton-41550,
  author       = {Clker-Free-Vector-Images},
  title        = {Skeleton, Human, Skeletal},
  year         = {n.d.},
  url          = {https://pixabay.com/vectors/skeleton-human-skeletal-anatomy-41550/},
  urldate      = {2025-11-09},
  organization = {Pixabay},
  note         = {Free for use under the Pixabay Content License}
}
}

\end{document}